\documentclass{article}

\PassOptionsToPackage{numbers}{natbib}
\usepackage[eandd, final]{neurips_2026}

\usepackage[utf8]{inputenc} 
\usepackage[T1]{fontenc}    
\usepackage{hyperref}       
\usepackage{url}            
\usepackage{booktabs}       
\usepackage{amsfonts}       
\usepackage{nicefrac}       
\usepackage{microtype}      
\usepackage{xcolor}         
\usepackage{graphicx}
\usepackage{amsmath}
\usepackage{multirow}
\usepackage{makecell}
\usepackage[numbers]{natbib}
\usepackage{subcaption}
\usepackage{tikz}
\usepackage{wrapfig}

\usepackage{titletoc}

\title{Self-Supervised Visual Representation Learning: Pretrain–Finetuning or Joint Training?}

\author{%
  Nusrat Munia \\
  Department of Computer Science\\
  University of Kentucky\\
  Lexington, KY 40506, USA \\
  \texttt{Nusrat.Munia@uky.edu} \\
  \And
    Tyler Ward \\
  Department of Computer Science\\
  University of Kentucky\\
  Lexington, KY 40506, USA \\
  \texttt{tyler.ward@uky.edu} \\
  \And
    Nishat Nayla \\
  Department of Computer Science\\
  University of Kentucky\\
  Lexington, KY 40506, USA \\
  \texttt{Nishat.Nayla@uky.edu} \\
  \And
    Matthew A. Massey \\
    Kentucky Geological Survey\\
    University of Kentucky\\
    Lexington, KY 40506, USA\\
  \texttt{matthew.massey@uky.edu} \\
  \And
    Abdullah-Al-Zubaer Imran\\
    Department of Computer Science\\
    University of Kentucky\\
    Lexington, KY 40506, USA\\
    \texttt{aimran@uky.edu} \\
}

\begin{document}

\maketitle

\begin{abstract}
Self-supervision is a powerful technique for learning visual representations from unlabeled data. Existing techniques primarily adopt a two-stage approach for self-supervised learning (SSL): a pretraining stage on unlabeled data followed by a finetuning stage on labeled data. While this pipeline has demonstrated extreme effectiveness, the interaction between self-supervised and supervised learning objectives remains insufficiently understood. In this work, we systematically investigate whether jointly optimizing the self-supervised and supervised objectives during training provides a better alternative. We compare two training paradigms: (1) the aforementioned pretraining followed by finetuning (PFT) and (2) joint training (JT), where self-supervised and supervised losses are optimized simultaneously in the same network. Across eight representative SSL methods and diverse computer vision tasks on natural, medical, crisis response, and remote sensing data, we evaluate performance under varying percentages of labeled data. Our results reveal that the relative effectiveness of PFT and JT depends strongly on the task at hand, the availability of labeled data, and the complexity of the domain. We find that JT consistently improves data and training efficiency while being robust in low-label settings, while PFT is more reliable in more specialized domains. We further analyze representation quality, robustness, and cross-domain generalization, providing new insights into how self-supervised and supervised objectives interact during optimization. We establish a comprehensive empirical benchmark for hybrid SSL–based semi-supervised learning and offer practical guidance for selecting appropriate training strategies across diverse vision applications. 
\end{abstract}

\section{Introduction}
\label{sec:intro}

Self-supervised learning (SSL) is a paradigm that enables models to learn meaningful visual representations without relying on human-annotated labels. Instead of explicit supervision, SSL generates its own supervisory signals directly from the data itself by formulating surrogate or pretext tasks that drive the network to learn invariant, discriminative, and semantically rich features~\cite{gui2024survey,jing2020self,liu2021self}. The learned representations can then be transferred to downstream tasks such as classification, detection, or segmentation with minimal fine-tuning, offering an efficient and scalable alternative to fully supervised training.
SSL strategies have shown remarkable performance on downstream tasks, even when compared against traditional fully-supervised models \cite{ali2024hyperspectral, azizi2023robust, nielsen2023self, wang2015optimizing, zhai2019s4l}. 

Early SSL approaches explored predicting the spatial arrangement between image patches~\cite{doersch2015unsupervised}, solving jigsaw puzzles~\cite{noroozi2016unsupervised}, reconstructing or colorizing missing regions~\cite{pathak2016context,zhang2016colorful}, and exploiting temporal continuity in unlabeled videos~\cite{wang2015unsupervised}. These methods showed that meaningful visual representations can be learned without manual labels, laying the foundation for modern SSL frameworks.

Contrastive learning emerged as a major milestone in SSL by viewing representation learning as an instance discrimination task~\cite{oord2018representation,chen2020simple,he2020momentum}. Methods like SimCLR~\cite{chen2020simple} and MoCo~\cite{he2020momentum} learned strong features by matching augmented views of the same image and separating different ones, achieving performance close to supervised models. Subsequent research introduced non-contrastive and redundancy reduction techniques such as BYOL~\cite{grill2020bootstrap}, SimSiam~\cite{chen2021exploring}, and Barlow Twins~\cite{zbontar2021barlow}, which removed the need for negative pairs and showed that asymmetry, normalization or feature de-correlation were sufficient to prevent representational collapse. More recently, the field has been transformed by masked image modeling (MIM) approaches inspired by masked language modeling in NLP with the release of BERT~\cite{devlin2019bert}. Methods like MAE~\cite{he2022masked}, SimMIM~\cite{xie2022simmim}, and iBOT~\cite{zhou2021ibot} train Vision Transformers (ViTs)~\cite{dosovitskiy2020image} to reconstruct missing patches or predict token embeddings, yielding representations with strong global semantic understanding.

\begin{figure*}[t]
    \centering
    \includegraphics[width=\linewidth]{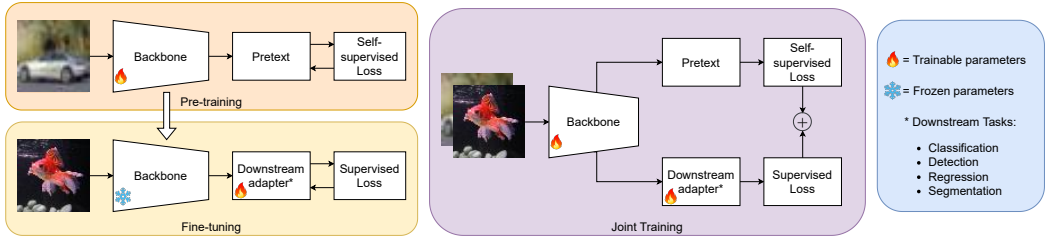}
    \caption{Illustration of the two competing training paradigms in self-supervised learning. The pre-train$\rightarrow$fine-tune framework (left) follows a two-stage process where the backbone is first trained with a self-supervised loss on unlabeled data and then fine-tuned on labeled data using a supervised loss. During fine-tuning, the pretrained backbone is kept frozen while only downstream adapters (e.g., classification, detection, regression, or segmentation heads) are trained. In contrast, the joint training paradigm (right) optimizes both self-supervised and supervised losses simultaneously within a unified framework, allowing the backbone and downstream adapters to learn jointly from both labeled and unlabeled samples. Trainable and frozen parameters are indicated by distinct symbols.}
    \label{fig:arch}
\end{figure*}

Today, SSL stands as a foundation of modern computer vision. It powers large-scale vision-language models (such as CLIP~\cite{radford2021learning}, ALIGN~\cite{jia2021scaling}, BLIP~\cite{li2022blip}, and Flamingo~\cite{alayrac2022flamingo}), serves as a pre-training backbone for advanced vision representation models (including DINO~\cite{caron2021emerging}, iBOT~\cite{zhou2021ibot}, BEiT~\cite{bao2021beit}, MAE~\cite{he2022masked}, and DINOv2~\cite{oquab2023dinov2}), and provides the foundation for modern multimodal and generative frameworks (such as Stable Diffusion~\cite{rombach2022high}, BLIP-2~\cite{li2023blip2}, LLaVA~\cite{liu2023visual}, Kosmos-2~\cite{peng2023kosmos2}, and GPT-4~\cite{achiam2023gpt}), unifying perception, reasoning, and generation across modalities. The trajectory of SSL, from handcrafted pretext tasks to large-scale transformer-based pre-training, marks a paradigm shift from designing narrow, task-specific objectives to learning general-purpose representations aligned with the intrinsic structure of visual and multimodal data. 

Despite these advances, a fundamental question remains insufficiently explored: \textit{how should self-supervised and supervised objectives interact during model training?} The prevalent two-stage paradigm: a self-supervised pre-training stage followed by fully supervised fine-tuning, which has become the \textit{de facto} pipeline for leveraging SSL in practice. However, this sequential process may not fully exploit the potential synergy between the two learning signals. 

In this work, we conduct a systematic investigation into the interplay between self-supervised pre-training and jointly optimized supervised and self-supervised learning (Fig.~\ref{fig:arch}). Specifically, we evaluate the following two competing SSL approaches:
\begin{enumerate}
\item 
 \textbf{Pre-train $\rightarrow$ Fine-tune (PFT):} a conventional two-stage setup in which models undergo self-supervised pre-training followed by supervised fine-tuning on labeled data for evaluation.
 \item 
 \textbf{Joint Training (JT):} a single-stage setup where self-supervised and supervised losses are optimized simultaneously during training.
\end{enumerate}

\begin{figure*}[t]
    \centering
    \includegraphics[width=\linewidth]{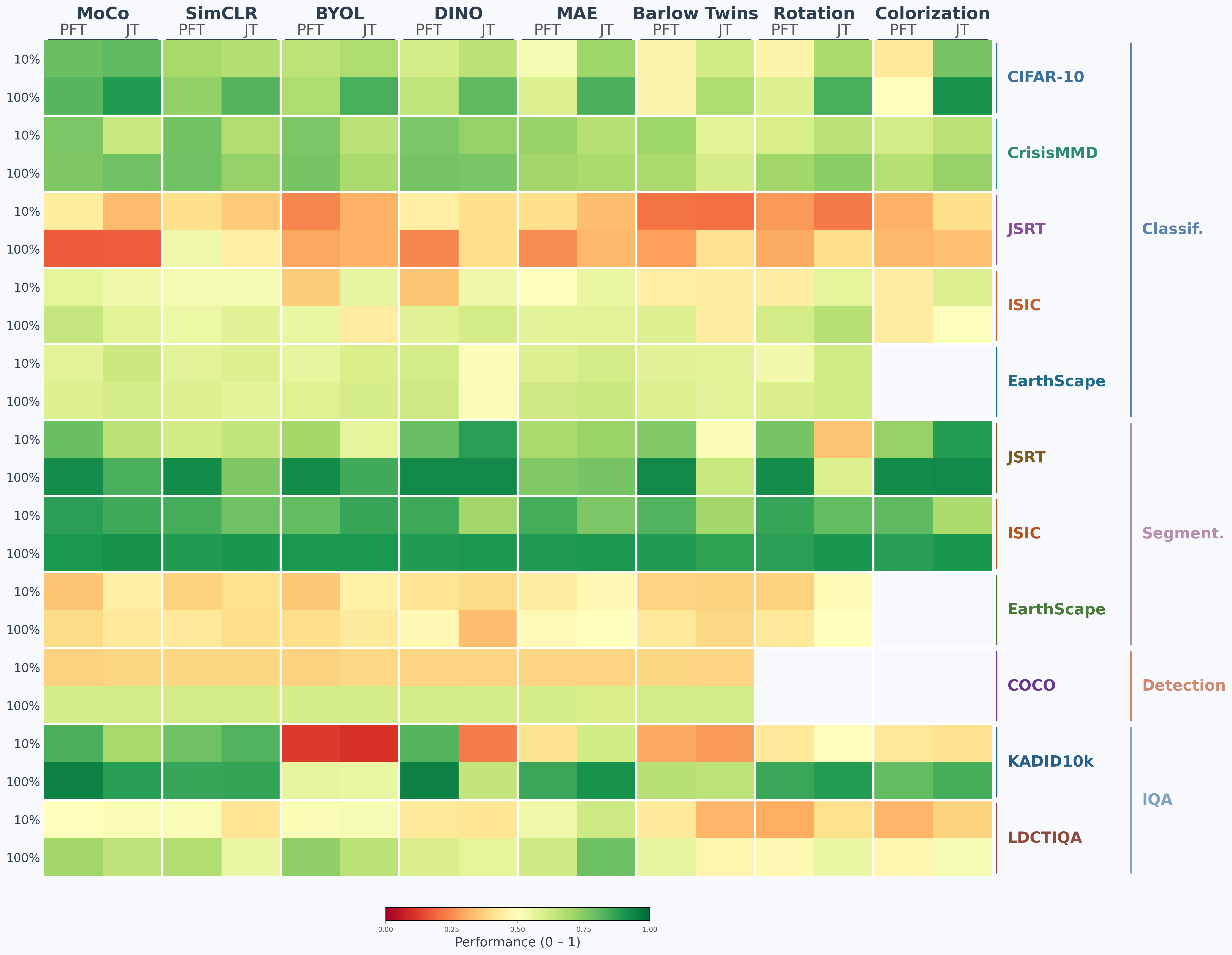}
    \caption{Overall performance comparison of self-supervised learning methods across multiple datasets and tasks. The heatmap reports performance (normalized to the range [0, 1]) for eight SSL approaches under two evaluation protocols.
    Results are shown across diverse domains, including natural images (CIFAR-10), crisis understanding (CrisisMMD), medical imaging (ISIC, JSRT), remote sensing (EarthScape), object detection (COCO), and image quality assessment (KADID10k, LDCTIQA). Rows correspond to different label fractions (10\% and 100\%), while columns represent SSL methods grouped by learning paradigm. Warmer colors indicate lower performance, and cooler colors indicate higher performance.}
    \label{fig:result_heatmap}
\end{figure*}

We perform extensive experiments across eleven datasets and eight representative SSL frameworks: Colorization, Rotation, SimCLR~\cite{chen2020simple}, BYOL~\cite{grill2020bootstrap}, MoCo~\cite{he2020momentum}, DINO~\cite{caron2021emerging}, MAE~\cite{he2022masked}, and Barlow Twins~\cite{zbontar2021barlow} to rigorously evaluate both training regimes under varying data scales, domains, and tasks. Our evaluation 
covers a broad range of vision tasks that include image classification, detection, semantic segmentation, and image quality assessment (IQA) to comprehensively assess the transferability and robustness of learned representations. 

To ensure a broad and realistic analysis, we experiment on both generic vision and domain-specific datasets. For the general domain, we employ standard benchmarks such as CIFAR-10~\cite{krizhevsky2009learning}, COCO~\cite{lin2014microsoft}, PASCAL VOC2012~\cite{everingham2012pascal}, KADID-10k~\cite{lin2019kadid} and KonIQ-10k~\cite{hosu2020koniq}. For specialized domains, we evaluate across (i) the crisis domain using CrisisMMD~\cite{alam2018crisismmd} and DMD~\cite{mouzannar2018damage}; (ii) the remote sensing domain using the EarthScape dataset~\cite{massey2025earthscape}; and (iii) the medical domain using ISIC~\cite{gutman2016skin} for skin lesion segmentation and classification; JSRT~\cite{shiraishi2000development} for chest x-ray segmentation and classification; and LDCTIQA~\cite{lee2023lowdosect} for computed tomography (CT) IQA. Overall, this work provides a unified empirical analysis and benchmark for understanding how self-supervised and supervised objectives can be combined effectively. Our main contributions are as follows:

\begin{itemize}
    \item A comprehensive evaluation of eight SSL frameworks under two training paradigms, PFT and JT, across classification, detection, IQA, segmentation, and geospatial analysis using diverse datasets.
    \item An analysis of performance under full-label, low-label, and cross-dataset settings, revealing when JT offers clear advantages and when PFT remains more reliable.
    \item A comparison of computational cost and training efficiency, showing that JT can significantly reduce training time while maintaining competitive performance gains in several tasks.
\end{itemize}

\section{Preliminary}
\label{sec:background}

\textbf{Self-Supervised Learning:}
SSL enables models to learn invariant and semantically meaningful representations from unlabeled data by constructing surrogate objectives that serve as implicit supervision signals. While many formulations exist, most SSL approaches can be broadly divided into three categories based on their learning objectives: contrastive, non-contrastive, and generative.

\textbf{Contrastive SSL:}
Contrastive learning formulates representation learning as an instance discrimination task, where the model aligns embeddings of augmented views of the same image (positive pairs) and separates them from others (negatives) using the InfoNCE loss~\cite{oord2018representation}. SimCLR~\cite{chen2020simple} demonstrated that strong augmentations, large batches, and a nonlinear projection head can yield representations comparable to supervised pre-training. MoCo~\cite{he2020momentum} introduced a momentum encoder with a dynamic memory queue for stable and scalable contrastive learning. SwAV~\cite{caron2020unsupervised} combined contrastive and clustering principles by assigning samples to learned prototypes, while InfoMin~\cite{tian2020makes} emphasized selecting augmentations that preserve minimal but sufficient mutual information. AdCo~\cite{hu2021adco} further improved discrimination by treating negatives as learnable adversarial parameters. Multimodal extensions such as CLIP~\cite{radford2021learning} and ALIGN~\cite{jia2021scaling} applied contrastive objectives across image–text pairs for joint vision–language pre-training. 

\textbf{Non-Contrastive SSL:}
Non-contrastive methods remove the need for explicit negative pairs by enforcing invariance and diversity through architectural asymmetry~\cite{grill2020bootstrap,chen2021exploring}, statistical constraints~\cite{zbontar2021barlow,bardes2021vicreg}, or teacher–student consistency~\cite{caron2021emerging,zhou2021ibot}. Instead of contrasting samples, these methods prevent representational collapse by ensuring embeddings from augmented views remain informative, de-correlated, and semantically aligned. They provide a stable alternative to contrastive frameworks, enabling efficient training even with smaller batch sizes or limited negatives. Redundancy-reduction approaches such as Barlow Twins~\cite{zbontar2021barlow} and VICReg~\cite{bardes2021vicreg} constrain cross-view feature correlations to maintain informative and diverse representations. Barlow Twins minimizes the cross-correlation matrix between embeddings from two augmented views to enforce invariance while de-correlating feature dimensions. 

Self-distillation methods~\cite{grill2020bootstrap,caron2021emerging,zhou2021ibot} instead use an online–teacher setup, where the teacher encoder is updated as an exponential moving average of the online encoder to provide consistent targets. BYOL~\cite{grill2020bootstrap} showed that meaningful representations can emerge solely from predicting the teacher’s latent features, while DINO~\cite{caron2021emerging} extended this paradigm to ViT using temperature-scaled soft targets to maintain stability and enable semantic clustering. iBOT~\cite{zhou2021ibot} further incorporated masked patch prediction into this framework, unifying image-level alignment with token-level reconstruction. 

\textbf{Generative SSL:}
Generative self-supervised methods learn representations by reconstructing input data or predicting missing content. Early reconstruction-based tasks such as context encoders~\cite{pathak2016context}, colorization~\cite{zhang2016colorful,dong2021self, hou2021self, hu2022self, pandey2022self, tiwari2023real, yang2023self}, and rotation prediction~\cite{devgon2020orienting, feng2019self, jing2018self, yamaguchi2021image} demonstrated that pixel-level prediction can capture semantic and spatial relationships. More recent approaches adopt MIM inspired by masked language modeling in NLP. For example, MAE~\cite{he2022masked} randomly masks a large portion of image patches and uses a Vision Transformer~\cite{dosovitskiy2020image} encoder–decoder to reconstruct the missing pixel content. 

\textbf{Semi-Supervised Learning (Semi-SL):}
Semi-SL bridges supervised and self-supervised paradigms by leveraging both labeled and unlabeled data to enhance generalization when annotations are scarce~\cite{oliver2018realistic,van2020survey}. Unlike SSL, which relies solely on self-generated signals, Semi-SL incorporates limited human-provided labels to guide representation learning while exploiting the structure of large unlabeled datasets, proving effective in domains such as medical imaging and remote sensing~\cite{cheplygina2019not,tuia2021toward}. Early methods, including the $\Pi$-Model~\cite{laine2016temporal} and Mean Teacher~\cite{tarvainen2017mean}, relied on consistency regularization to enforce prediction stability across augmentations, while later approaches such as MixMatch~\cite{berthelot2019mixmatch}, ReMixMatch~\cite{berthelot2019remixmatch}, and FixMatch~\cite{sohn2020fixmatch} combined this principle with pseudo-labeling and strong-weak augmentation strategies. Recent frameworks like UDA~\cite{xie2020unsupervised} further integrate SSL pretraining or co-train supervised and unsupervised objectives to achieve more robust and data-efficient representations. This synergy between labeled and unlabeled learning motivates our analysis of the joint optimization scheme. While JT is itself a form of Semi-SL, it differs from these methods in that the unsupervised signal comes from an SSL pretext task rather than from pseudo-labels or consistency regularization on the supervised objective. 

\textbf{Joint optimization of self-supervised and supervised objectives:}
Several prior works have explored simultaneous optimization of SSL and supervised losses, though typically restricted to a single SSL method or task family. S4L~\cite{zhai2019s4l} introduced rotation prediction and exemplar-based SSL as auxiliary objectives for semi-supervised image classification, demonstrating gains on ImageNet at low label fractions. SimCLRv2~\cite{chen2020big} extended SimCLR with a labeled-data fine-tuning step that retains the contrastive objective, showing strong semi-supervised performance. JGCL~\cite{akkas2022jgcl} jointly optimizes contrastive and supervised losses for graph node classification. Another work \cite{zhang2023combining} studied combinations of SSL and supervised learning under noisy labels, focusing on robustness rather than systematic comparison.  Each of the above studies fixes one SSL objective and one task, leaving open whether the joint-training advantage they observe is general or specific to their setup. 

\section{Methods}
\textbf{Problem Formulation:}
Let $\mathcal{D}_l = \{(x_i, y_i)\}_{i=1}^{N_l}$ denote a labeled dataset containing $N_l$ samples, where $x_i \in \mathcal{X}$ represents an input image and $y_i \in \mathcal{Y}$ is its corresponding ground truth label. Let $\mathcal{D}_u = \{x_j\}_{j=1}^{N_u}$ represent an unlabeled dataset, which may belong to the same or a different domain. The model $f_{\theta}$, parameterized by $\theta$, encodes $x$ into a latent representation $h = f_{\theta}(x)$, and an optional projection head $g_{\phi}$ produces an embedding $z = g_{\phi}(h)$ used for the self-supervised objective. The self-supervised loss $\mathcal{L}_{\text{SSL}}$ encourages representation consistency across transformations or masked views of the same image, depending on the SSL framework. For labeled samples $(x_i, y_i) \in \mathcal{D}_l$, the model is optimized to minimize a task-oriented supervised loss $\mathcal{L}_{\text{sup}}$ as $\mathcal{L}_{\text{Sup}} = \frac{1}{N_l} \sum_{i=1}^{N_l} \ell_{\text{task}}\big(f_{\theta}(x_i), y_i \big),$
where $\ell_{\text{task}}$ depends on the nature of the downstream task.

\textbf{PFT:}
In the sequential setting, the model is first pre-trained on unlabeled data ($\mathcal{D}_u$) using a self-supervised objective ($\mathcal{L}_{\text{SSL}}$) and subsequently fine-tuned on a labeled subset for the downstream evaluation. This approach follows the conventional transfer learning paradigm used in traditional SSLs, where self-supervised representations are repurposed for task-specific optimization.

\textbf{JT:}
In the joint setting, self-supervised and supervised objectives are optimized simultaneously within a unified framework. The model learns from both labeled and unlabeled data in a single training phase, balancing the self-supervised loss $ \mathcal{L}_{\text{SSL}} $  defined on the unlabeled data and the supervised loss $ \mathcal{L}_{\text{sup}} $ defined on the labeled data. The overall model training objective is defined as  
$ \mathcal{L}_{\text{total}} =  \mathcal{L}_{\text{SSL}} + \mathcal{L}_{\text{sup}} $. This formulation allows the network to jointly preserve invariance from unlabeled samples while leveraging discriminative cues from labeled data. Joint optimization shares the encoder $f_{\theta}$ between two loss landscapes. The encoder's update at step $t$ is
$\Delta \theta = -\eta \left( \nabla_{\theta} \mathcal{L}_{\text{SSL}} + \nabla_{\theta} \mathcal{L}_{\text{sup}} \right)$, which is constructive when the two gradients are aligned.
Joint optimization has been shown to improve data efficiency and robustness under low-label or domain-shifted conditions~\cite{zhai2019s4l,chen2020big,imran2020self}. In our experiments, we systematically analyze how different SSL objectives interact with supervised losses and impact feature transferability, convergence, and generalization across datasets.

To further assess the advantages of this semi-supervised formulation, we conduct experiments varying the labeled data proportions (10\%, 20\%, 50\%, and 100\%). This setup closely mimics real-world conditions where labeled data are scarce but unlabeled data are abundant. During JT training, the supervised loss is selectively applied to labeled samples, whereas the self-supervised objective continues to update the shared encoder across both labeled and unlabeled batches. This joint optimization allows the model to maintain rich structural representations from the self-supervised task while progressively aligning them with class-discriminative cues. Such a hybrid learning paradigm encourages mutual reinforcement between the two objectives: self-supervision provides robust feature priors that stabilize the supervised optimization under limited labels, and supervision, in turn, helps refine and ground the self-supervised embeddings toward task-relevant semantics.

\section{Experimental Evaluation}
\label{sec:experiment}

\subsection{Implementation Details}

\textbf{Data:}
We evaluate our methods on eleven publicly available datasets spanning general and domain-specific settings. CIFAR-10~\cite{krizhevsky2009learning}, COCO~\cite{lin2014microsoft}, PASCAL VOC2012~\cite{everingham2010pascal}, KADID-10k~\cite{lin2019kadid}, and KonIQ-10k~\cite{hosu2020koniq} represent small-scale, large-scale, and distortion-focused natural image benchmarks, respectively, covering varying data complexity and visual quality. ISIC 2016~\cite{gutman2016skin}, JSRT~\cite{shiraishi2000development}, LDCTIQA~\cite{lee2023lowdosect}, CrisisMMD~\cite{alam2018crisismmd}, DMD~\cite{mouzannar2018damage}, and EarthScape~\cite{massey2025earthscape} span medical, crisis, and remote sensing domains, providing diverse evaluation settings across application domains.

\textbf{Backbone Architectures:}
For each of our experiments centered on the performance of SSL under the traditional PFT regimen, we integrate a ResNet-18~\cite{he2016deep} backbone within each SSL framework. For downstream evaluation, we employ a linear evaluation protocol where the encoder is frozen and only a linear layer is trained on labeled data. In the JT setting, the encoder is shared and optimized jointly through both supervised and self-supervised losses. For object detection, we employ the recently released YOLOv12~\cite{tian2025yolov12} model. For the PFT configuration, we employ SSL pre-training, inject the pre-trained weights into the YOLOv12 backbone, and fine-tune YOLOv12. During JT training, the SSL pipeline is integrated directly with YOLOv12, and the losses are minimized jointly. For IQA, we use a regression-based evaluation setup. During fine-tuning, the SSL pre-trained backbone is kept frozen, and a small regression head is trained on top using only labeled IQA samples. In JT, we combine contrastive loss on unlabeled pairs with an MSE regression loss on labeled samples. For semantic segmentation, we use a U-Net~\cite{ronneberger2015u} with a ResNet-18 encoder. In the PFT setting, SSL pre-trained weights initialize the encoder, and the decoder network is fine-tuned using labeled masks. In the JT setting, the encoder is shared between the SSL and segmentation objectives, and both losses are optimized together using labeled and unlabeled samples within the same training loop.

\textbf{Evaluation:}
We assess model performance using standard metrics that capture both accuracy and class-level balance. For general-domain datasets such as CIFAR-10, we report Top-1 and Top-5 accuracies to evaluate overall recognition performance. For domain-specific datasets, including CrisisMMD and EarthScape, we additionally report F1 scores to provide a fair comparison across classes. For the object detection task, we evaluate the performance of the SSL training paradigms using four common metrics: precision (P), recall (R), and mean average precision (mAP) at an intersection over union (IoU) threshold of 0.5, and again at a threshold ranging from 0.5 to 0.95 in steps of 0.05. For the IQA task, the goal is to predict differential mean opinion scores (DMOS) on a 1–5 quality scale. To evaluate, we report two correlation coefficients: Spearman Rank Order Correlation Coefficient (SROCC) and Pearson Linear Correlation Coefficient (PLCC). 
For semantic segmentation, we evaluate performance using the Dice coefficient and mean IoU (mIoU). 

\begin{figure}[t]
    \centering
    \subcaptionbox{Standard CIFAR-10}{
   \includegraphics[width=0.45\linewidth, trim={0cm 0cm 4cm 0cm},clip]{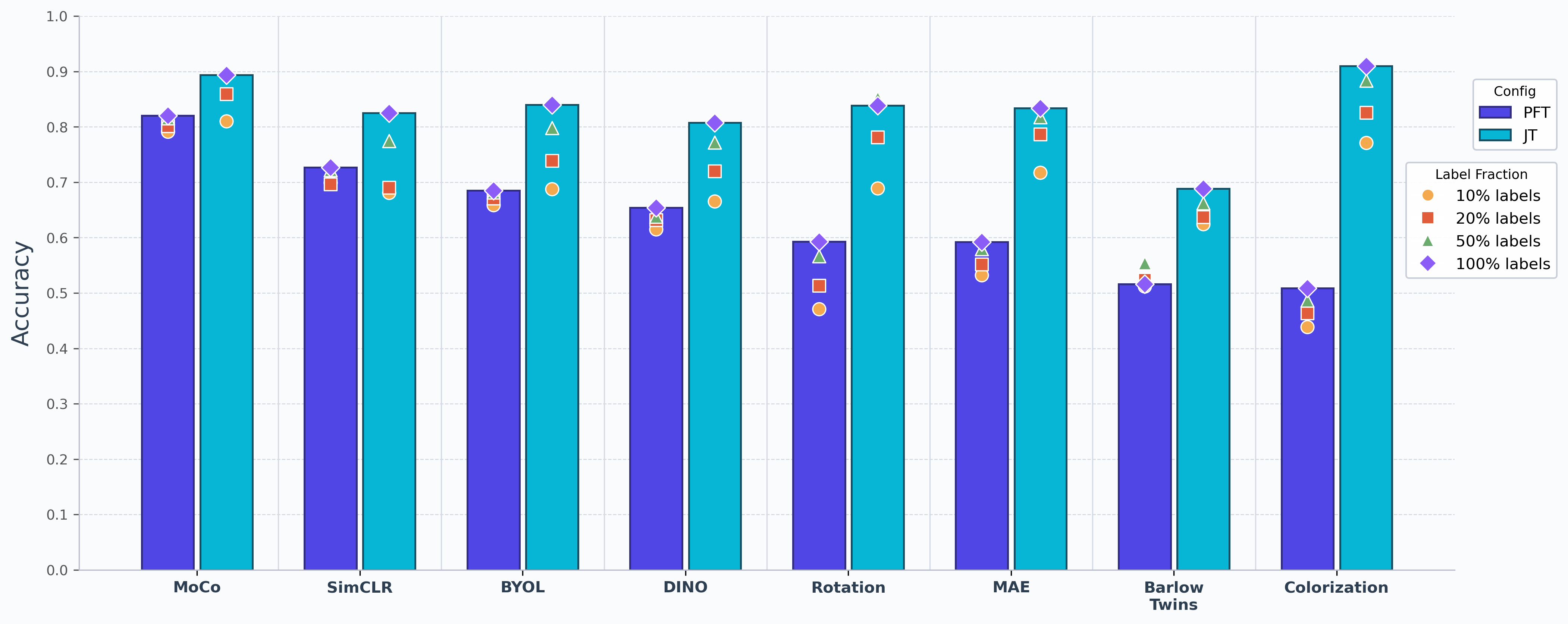}}
    \hfill
    \subcaptionbox{Adversarial CIFAR-10}{
   \includegraphics[width=0.53\linewidth]{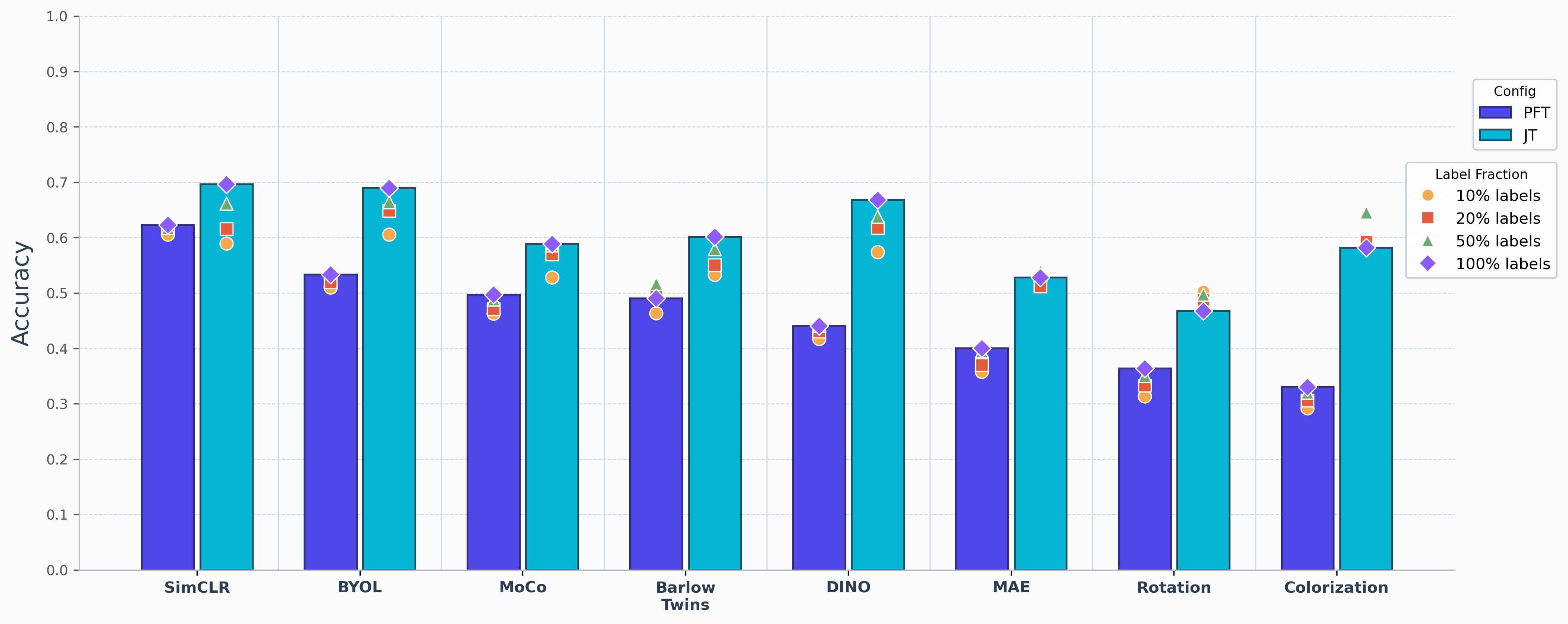}}
    \caption{
    CIFAR-10 classification accuracy under PFT and JT settings. Left: standard training and evaluation. Right: training and evaluation with 10\% adversarial noise. Bars show accuracy at 100\% labels, while markers indicate performance across different label fractions.
    }
    \label{fig:cifar10_combined}
\end{figure}

\subsection{Results and Discussion}

Fig.~\ref{fig:result_heatmap} summarizes the overall performance trends of the explored SSL methods across multiple datasets, downstream tasks, label fractions, and training paradigms. Overall, the results demonstrate that the effectiveness of PFT and JT depends strongly on the downstream application, data availability, and domain complexity. Across several classification, segmentation, and image quality assessment tasks, JT consistently improves training efficiency while achieving competitive or superior performance, particularly under limited-label settings. 

In contrast, PFT remains more reliable for several domain-specific tasks and contrastive learning frameworks, especially when sufficient labeled data are available. We further observe that reconstruction-oriented and auxiliary-task-based SSL methods, including MAE, Rotation, and Colorization, generally benefit more from JT, while contrastive and redundancy-reduction methods such as MoCo, SimCLR, DINO, and Barlow Twins often exhibit stronger transferability under PFT. The experiments additionally reveal that SSL representations learned through JT are frequently more robust under low-data and noisy conditions while maintaining competitive generalization performance. 

\begin{figure}[t]
    \centering
    \subcaptionbox{CrisisMMD}{
   \includegraphics[width=0.45\linewidth, trim={0cm 0cm 4cm 0cm},clip]{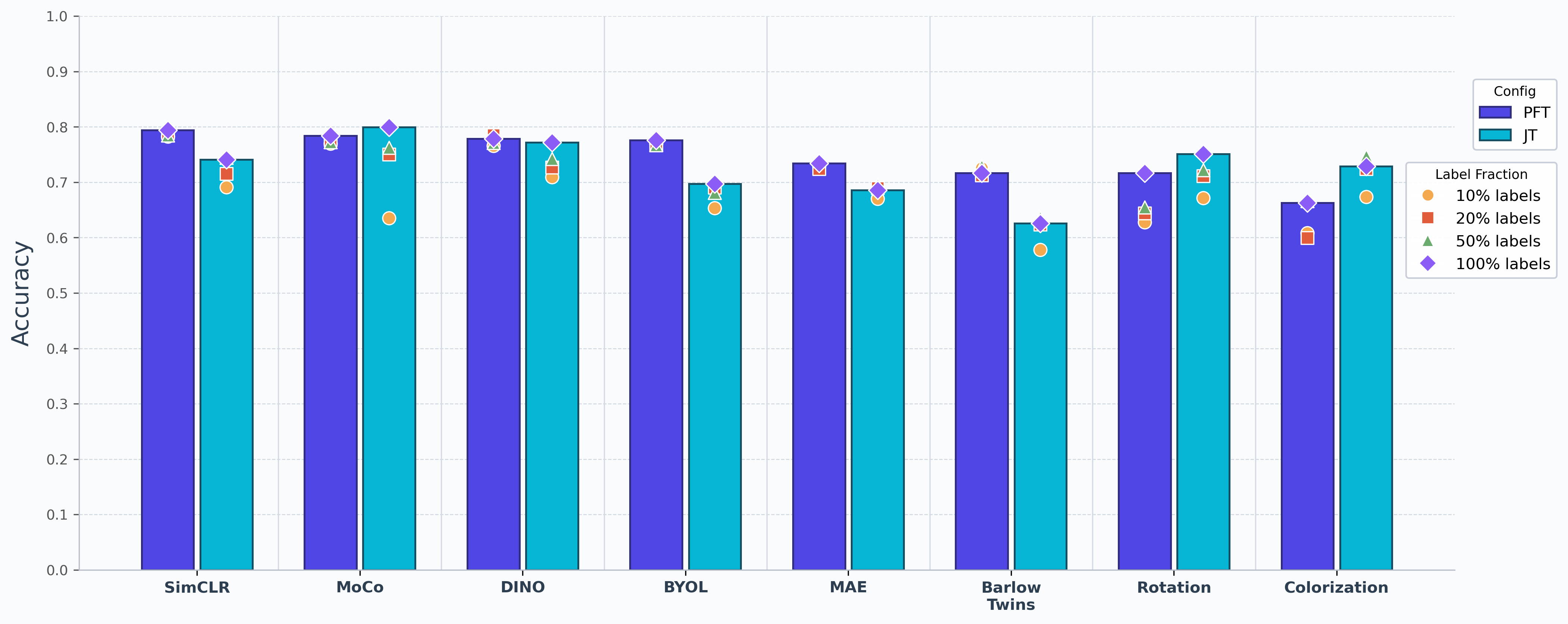}}
    \hfill
    \subcaptionbox{DMD}{
   \includegraphics[width=0.53\linewidth]{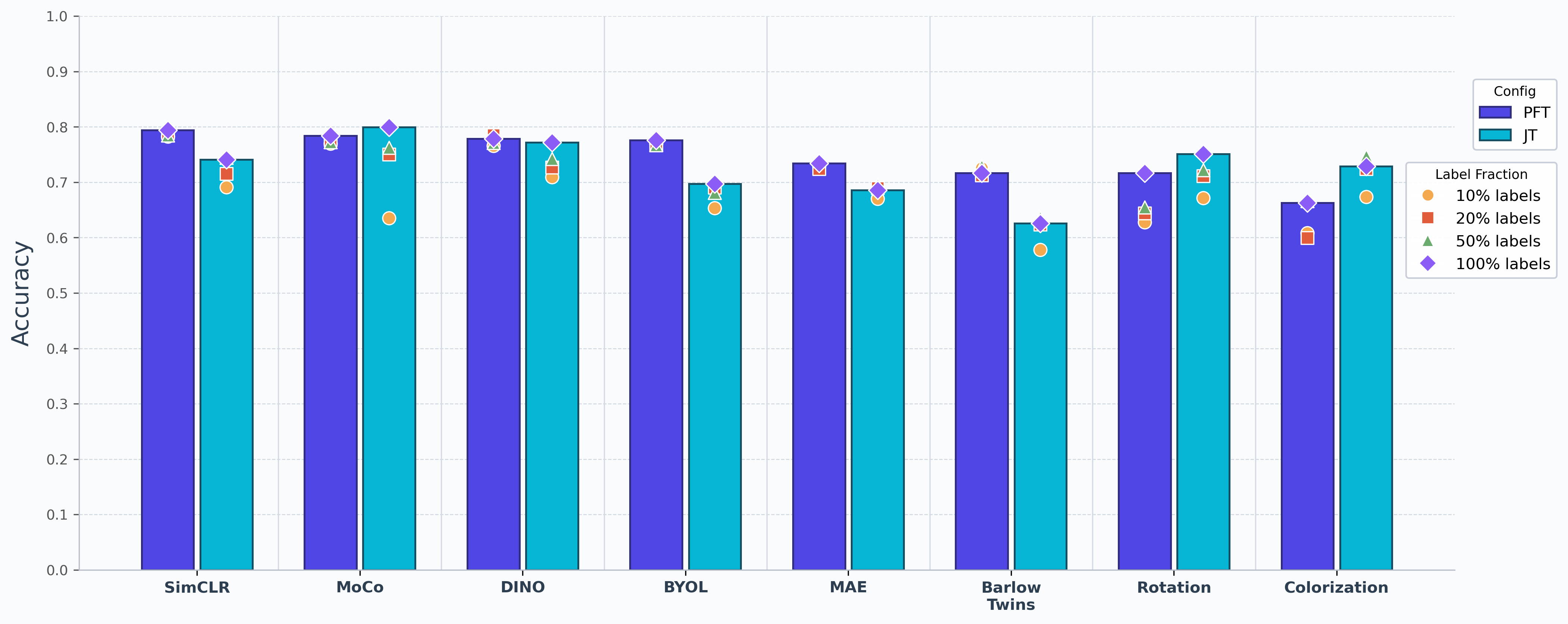}}
    \caption{
    Crisis analysis performance. (a) Standard CrisisMMD~\cite{alam2018crisismmd} classification performance under in-domain evaluation. (b) Cross-dataset generalization where models are trained on CrisisMMD and evaluated on the DMD dataset~\cite{mouzannar2018damage}. Bar plots represent performance at 100\% labeled data, while overlaid markers show results across varying label fractions under both PFT and JT protocols.
    }
    \label{fig:crisismmd-dmd}
\end{figure}

\textbf{Image Classification:} 
The classification results from CIFAR-10 and CrisisMMD are shown in Figs.~\ref{fig:cifar10_combined} and \ref{fig:crisismmd-dmd}. JT produces the largest gains for reconstruction-oriented methods: on CIFAR-10, Colorization JT achieves 0.9099 accuracy vs 0.5090 under PFT at 100\% labels, and MAE JT reaches 0.8341 vs 0.5920. These gains persist across label fractions and under mild adversarial noise, where JT consistently preserves higher accuracy. Contrastive methods such as MoCo and SimCLR benefit more modestly from JT on natural images, while showing stronger PFT performance on domain-shifted datasets. On CrisisMMD, SimCLR PFT achieves 0.7941 accuracy with 0.7742 cross-domain transfer to DMD, whereas JT improves in-domain performance for Colorization and Rotation while reducing training time by up to 7 times. These results confirm that JT improves data and training efficiency for generative pretexts, while contrastive methods retain stronger transferability under PFT when label availability is sufficient. Detailed results on classification are in the Appendix.

\textbf{Image Segmentation:}
We evaluate segmentation performance on ISIC-2016, JSRT, and EarthScape under different SSL training paradigms. The complete ISIC-2016 and JSRT segmentation results are provided in Appendices~\ref{appendixisic}, \ref{appendix:jsrt_seg}, and \ref{appendix:earthscape}. On ISIC-2016, JT improves performance for several methods at full supervision while reducing training time substantially: MoCo JT achieves the highest Dice of 0.9123 and mIoU of 0.8704 vs 0.8946 and 0.8540 under PFT, with a 4$\times$ reduction in training time (Table~\ref{tab:isic_seg}). However, DINO and Barlow Twins degrade under JT at 10\% labels, indicating that self-distillation and redundancy-reduction objectives can be destabilized by supervised segmentation gradients under severe label scarcity. On JSRT, the contrast is sharper: Colorization JT improves Dice from 0.7261 to 0.8862 at 10\% labels, while Rotation JT collapses from 0.9203 to 0.5940 at full supervision, reflecting spatial incompatibility between rotation-equivariant features and dense anatomical localization. Overall, neither paradigm consistently dominates for segmentation, and the optimal strategy depends strongly on the SSL objective and label availability.

\textbf{Object Detection:}
Performance is broadly comparable across paradigms, with most methods differing by less than 0.005 mAP@50 (Appendix~\ref{appendix:crisis}). The primary JT advantage is computational: training time is reduced by up to 80\% at 10\% labels while maintaining competitive detection performance. Cross-dataset generalization to PASCAL VOC2012 is similarly uniform across paradigms, suggesting that the YOLOv12 task-specific optimization dominates the pretraining signal regardless of paradigm.

\begin{figure}[t]
    \centering
    \includegraphics[width=0.97\linewidth, trim={0cm 0cm 0cm 2cm},clip]{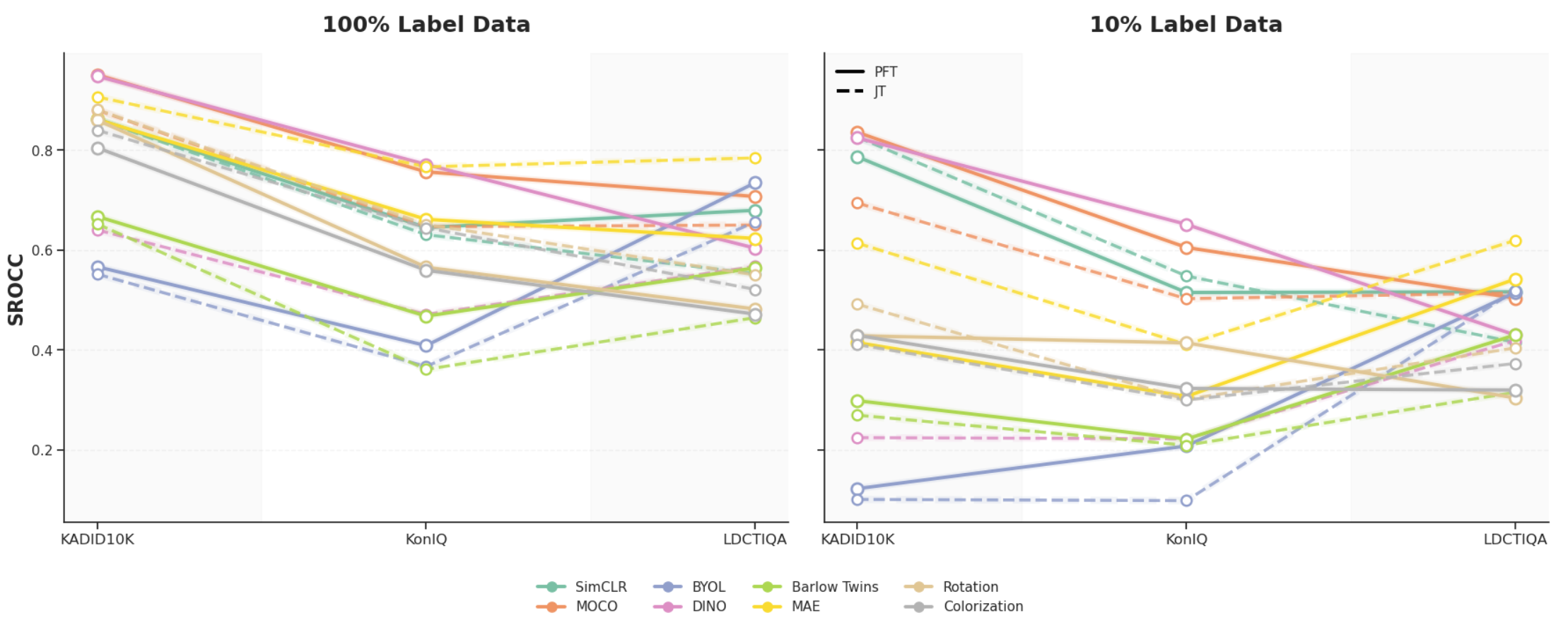}
    \caption{IQA performance (SROCC) for the eight SSL techniques with PFT and JT evaluated on three different datasets (KADID10k, KonIQ, and LDCTIQA). KADID10k-trained models are evaluated on both KADID10k and KonIQ datasets. Full supervision: 100\% labeled data (left), limited supervision: 10\% labeled data (right). Other proportion results are available in Appendix~\ref {appendix:iqa}.}
    \label{fig:iqa_srocc}
\end{figure}

\textbf{Image Quality Assessment:}
Fig.~\ref{fig:iqa_srocc} presents IQA results across KADID-10K, KonIQ-10K, and LDCTIQA. As seen, contrastive methods strongly favor PFT; MoCo with PFT achieves SROCC 0.9487 on KADID-10K vs 0.8820 under JT, and DINO PFT reaches 0.9458 vs 0.6420 under JT at full supervision. Reconstruction-oriented methods benefit from JT; MAE JT achieves SROCC 0.9083 vs 0.8592 under PFT on KADID-10K, with a larger gain on LDCTIQA. Rotation and Colorization show similar JT improvements under low-label and cross-dataset settings. 

\begin{figure}[t]
    \centering
    \includegraphics[width=\linewidth]{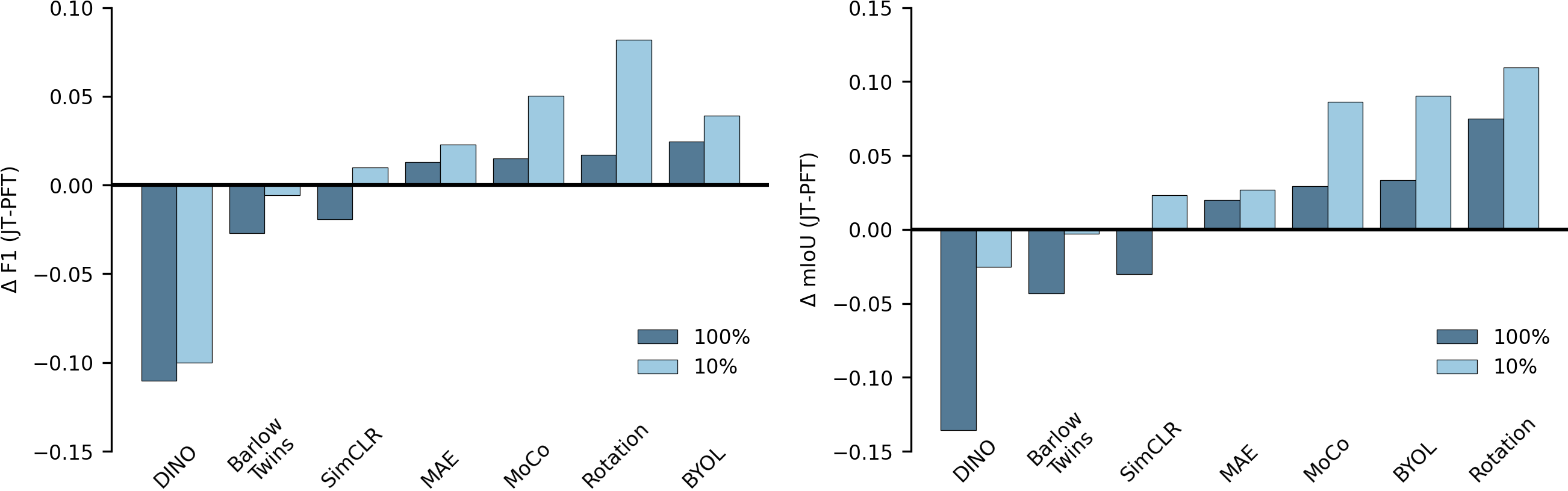}
    \caption{Change in downstream performance under joint training ($\mathrm{JT} - \mathrm{PFT}$) across SSL methods on EarthScape for multilabel classification (left) and semantic segmentation (right) under full (100\%) and limited (10\%) supervision. Positive values indicate improved performance under JT relative to PFT, whereas negative values indicate stronger performance under PFT.}
    \label{fig:earthscape_delta}
\end{figure}

\textbf{Generalizability:}
Figs.~\ref{fig:crisismmd-dmd} and \ref{fig:earthscape_delta} present the main cross-domain generalization results, while additional IQA and detection transfer experiments are provided in Appendices~\ref{appendix:iqa} and \ref{appendix:coco}. On CrisisMMD$\rightarrow$DMD transfer, contrastive methods strongly favor PFT: SimCLR PFT achieves DMD accuracy of 0.8037 at 10\% labels vs 0.6771 under JT. In contrast, auxiliary-task methods, Colorization and Rotation, show modest JT improvements in cross-domain accuracy at full supervision (0.5463 vs. 0.5409). On COCO$\rightarrow$PASCAL VOC2012, cross-dataset detection performance is remarkably uniform across paradigms. SimCLR PFT and JT achieve nearly identical mAP@50 of 0.855 and 0.854, respectively, and MoCo PFT and JT both reach 0.855. On EarthScape, geographic domain-shift retention under JT is more variable. BYOL JT improves in-domain mIoU from 0.3487 to 0.3820 at full supervision, while cross-domain mIoU remains comparable. In IQA datasets, contrastive methods under PFT maintain stronger cross-dataset SROCC consistency on KADID-10K$\rightarrow$KonIQ transfer. MoCo PFT achieves an SROCC of 0.9487 on KADID-10K and remains competitive on KonIQ, while MAE and Rotation benefit more from JT.

\textbf{Efficiency analysis:}
JT eliminates the sequential pretraining stage and consistently reduces wall-clock training time across all tasks and SSL methods. On COCO detection, DINO JT reduces training from 21.3 to 4.3 hours at 10\% labels with a mAP@50 difference of only 0.001. On CrisisMMD, Rotation JT cuts training from 7.88 to 1.03 hours while improving accuracy from 0.7167 to 0.7512. These efficiency gains are particularly significant given that JT achieves them without sacrificing task performance in most settings, establishing it as the preferred paradigm when computational resources are constrained or rapid iteration is required.

\textbf{Adversarial Robustness:}
Fig.~\ref{fig:cifar10_combined} reveals that robustness under adversarial perturbations is strongly method- and paradigm-dependent. On CIFAR-10 under noise strength 0.1, JT substantially improves robustness for reconstruction-oriented and auxiliary-task methods. Colorization JT preserves an accuracy of 0.8782 vs 0.3993 under PFT, and MAE JT achieves 0.7860 vs 0.5723. These gains mirror the clean-accuracy improvements and suggest that joint optimization encourages these methods to learn features that are inherently more invariant to input perturbations. Under stronger noise (0.5), contrastive methods show comparatively stronger resilience. BYOL JT and SimCLR JT achieve better accuracy compared to their PFT's. For IQA under adversarial settings (Fig.~\ref{fig:iqa_adv}), MoCo and DINO maintain strong SROCC with smaller drops than reconstruction-oriented and auxiliary-task methods under both PFT and JT. PFT consistently outperforms JT for both methods, suggesting that the sequential pretraining stage produces more stable perceptual representations that are less sensitive to sparse pixel-level perturbations.

\textbf{Interpretability:}
Grad-CAM visualizations on the CIFAR-10 dataset for different SSL methods under the PFT and JT training paradigms are provided in Fig.~\ref{fig:gradcam_ssl}. Contrastive methods such as SimCLR, BYOL, and Barlow Twins focus on relatively small localized regions near the target object, whereas MoCo, MAE, and Rotation exhibit broader and more distributed activation patterns. The visualizations further indicate that the transition from PFT to JT can noticeably change the spatial focus of the learned representations, particularly for MAE and Rotation, even with moderate classification performance.

\begin{figure}[t]
\centering

\resizebox{\linewidth}{!}{
\begin{tabular}{c c c c c c c c c}
& Colorization & Rotation & SimCLR & BYOL & MoCo & MAE & DINO & Barlow Twins\\
PFT
&
 \includegraphics[width=0.15\linewidth]{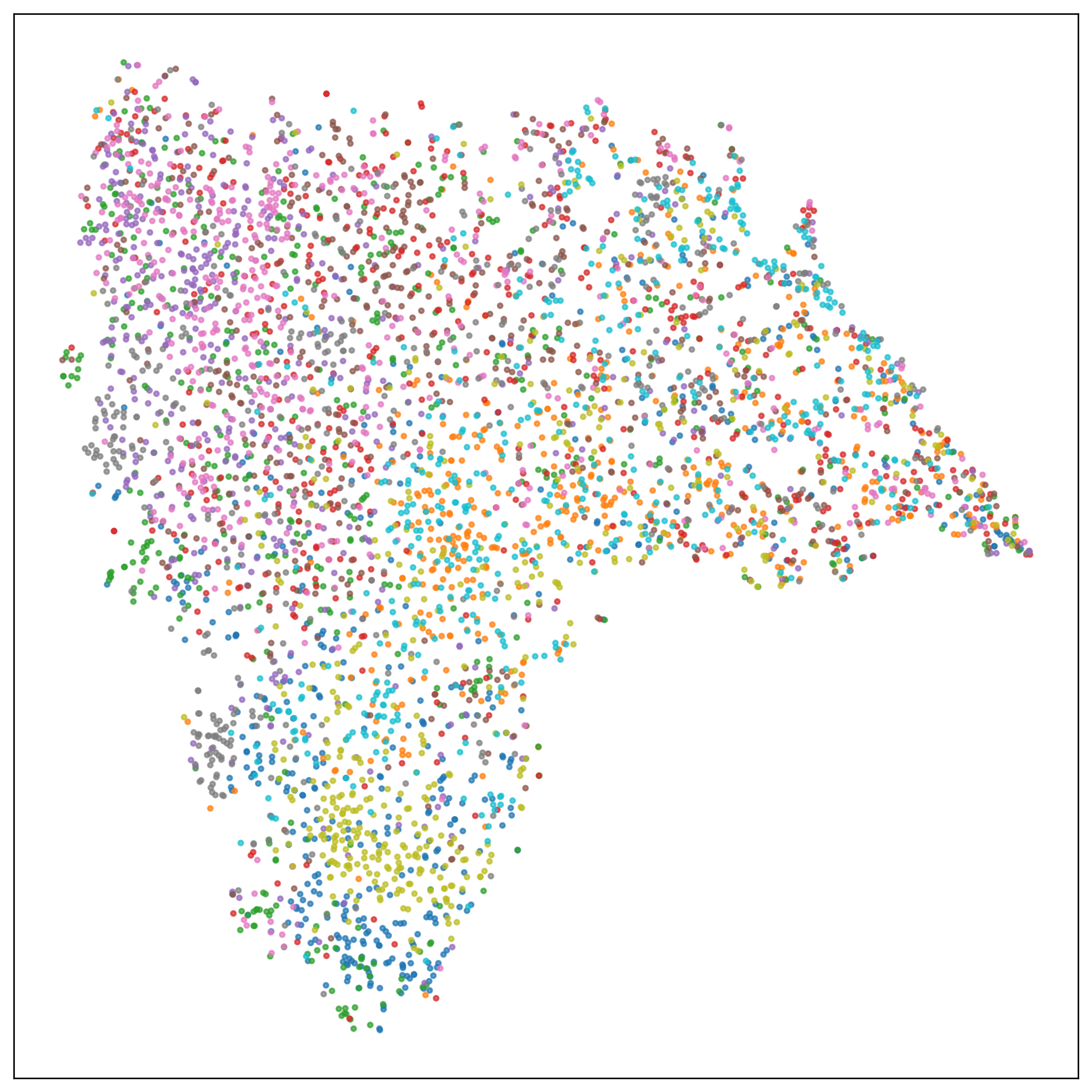}
&
 \includegraphics[width=0.15\linewidth]{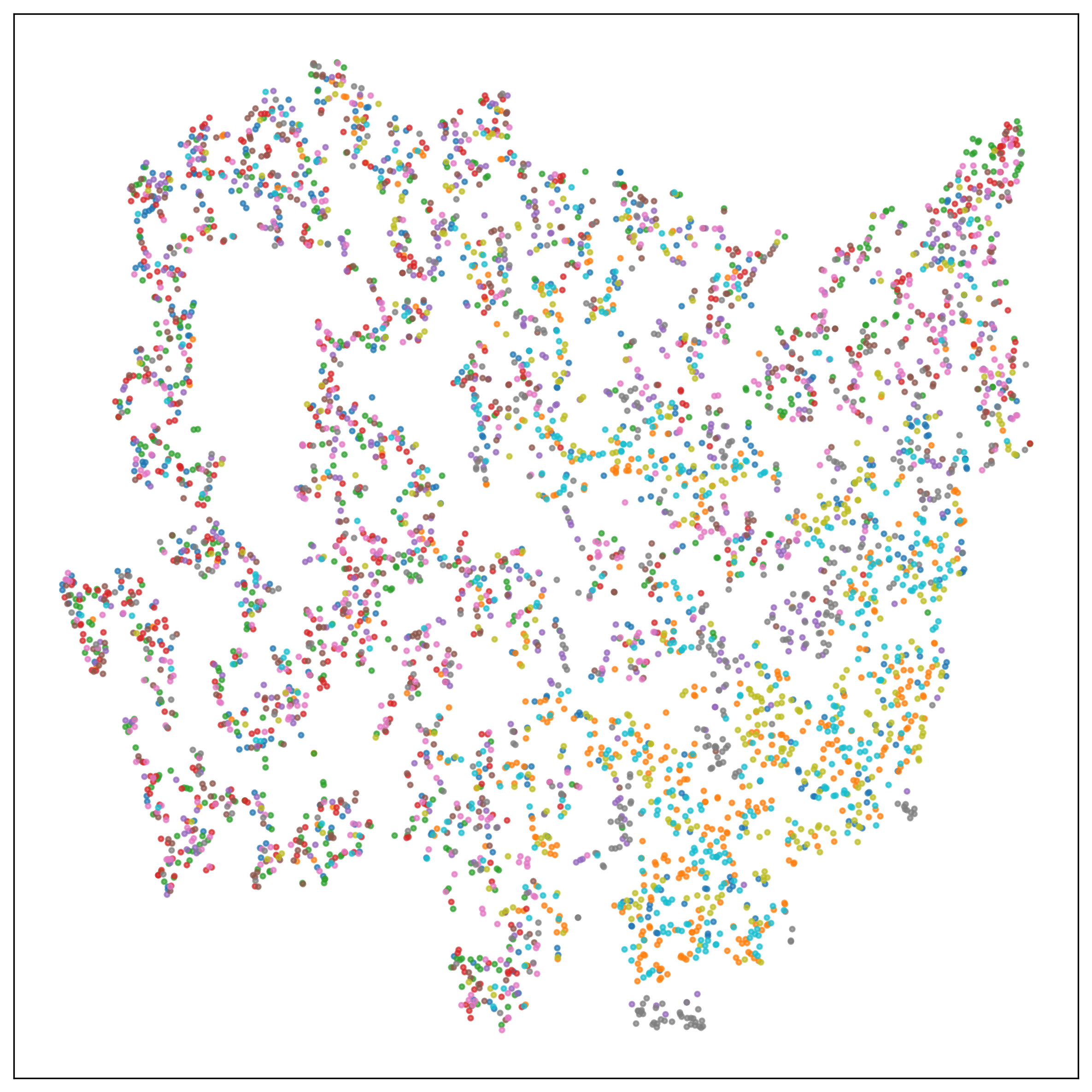}
&
 \includegraphics[width=0.15\linewidth]{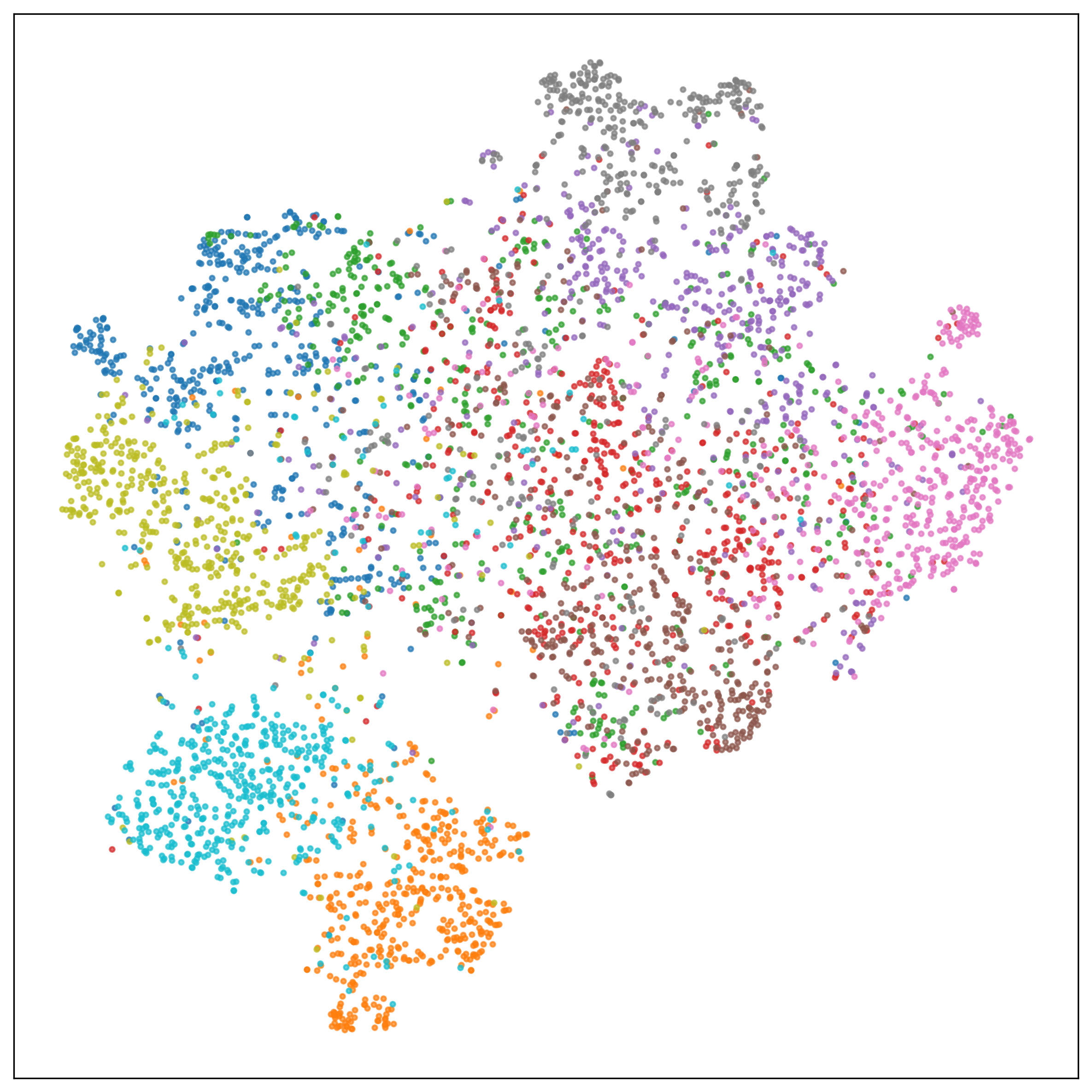}
&
\includegraphics[width=0.15\linewidth]{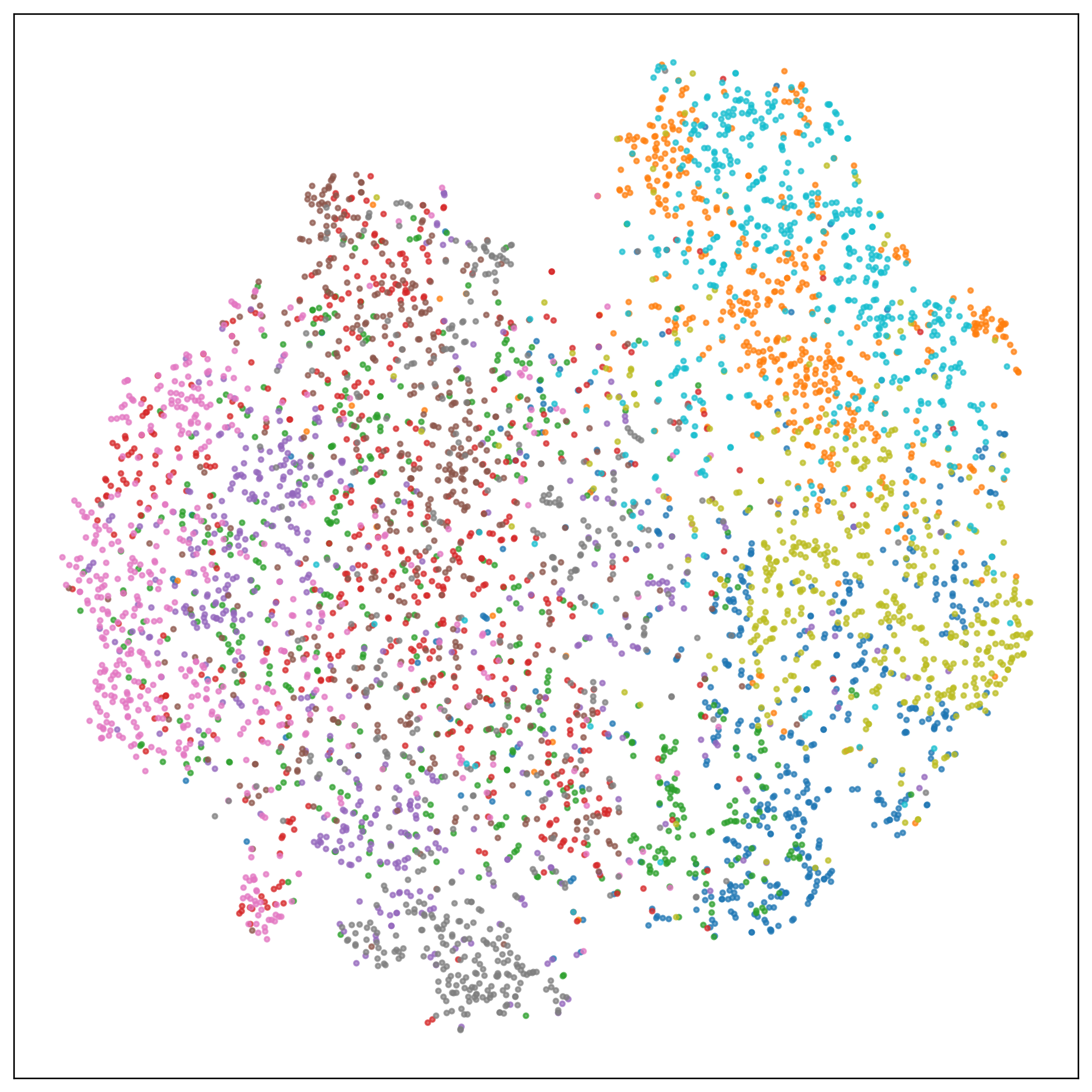}
&
\includegraphics[width=0.15\linewidth]{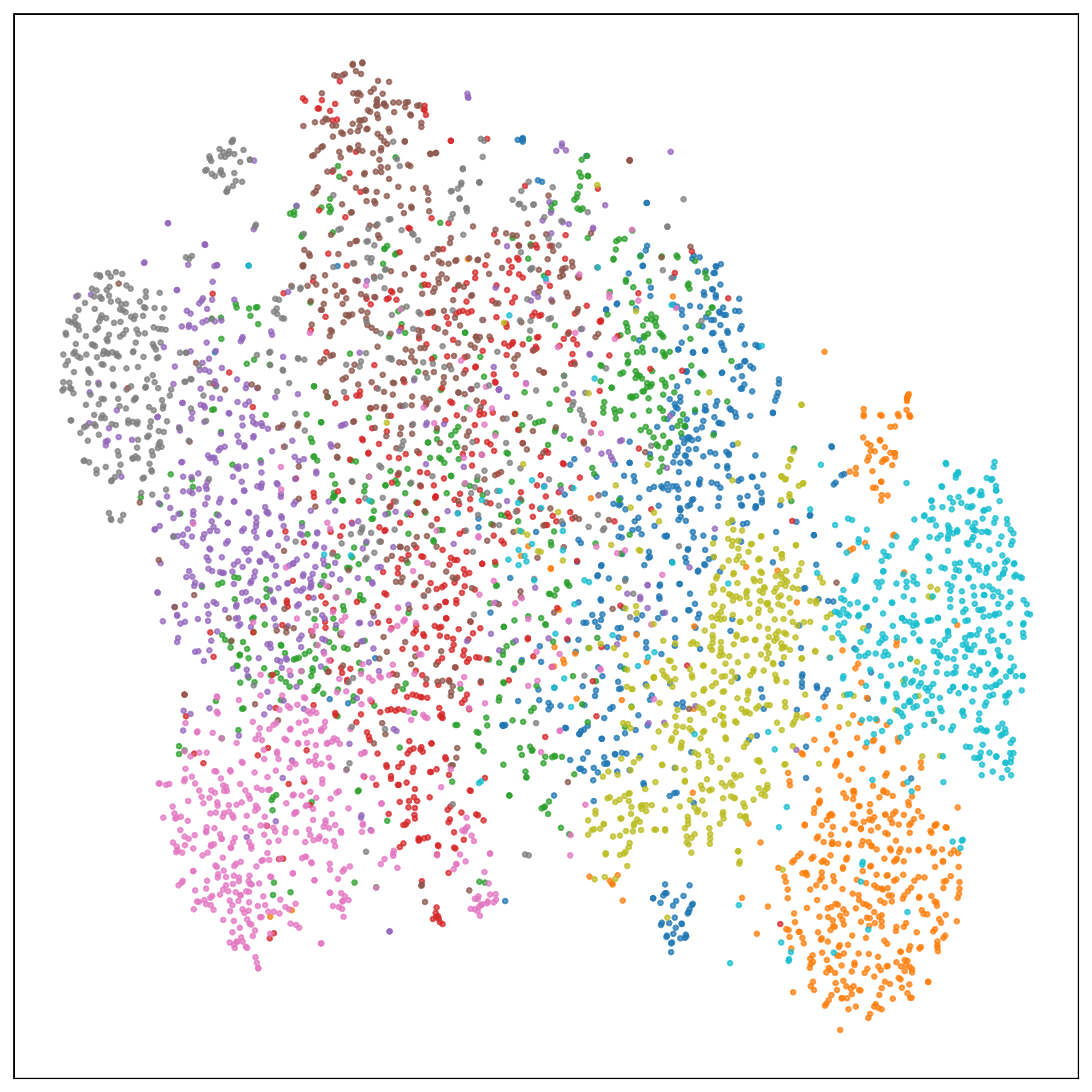}
&
\includegraphics[width=0.15\linewidth]{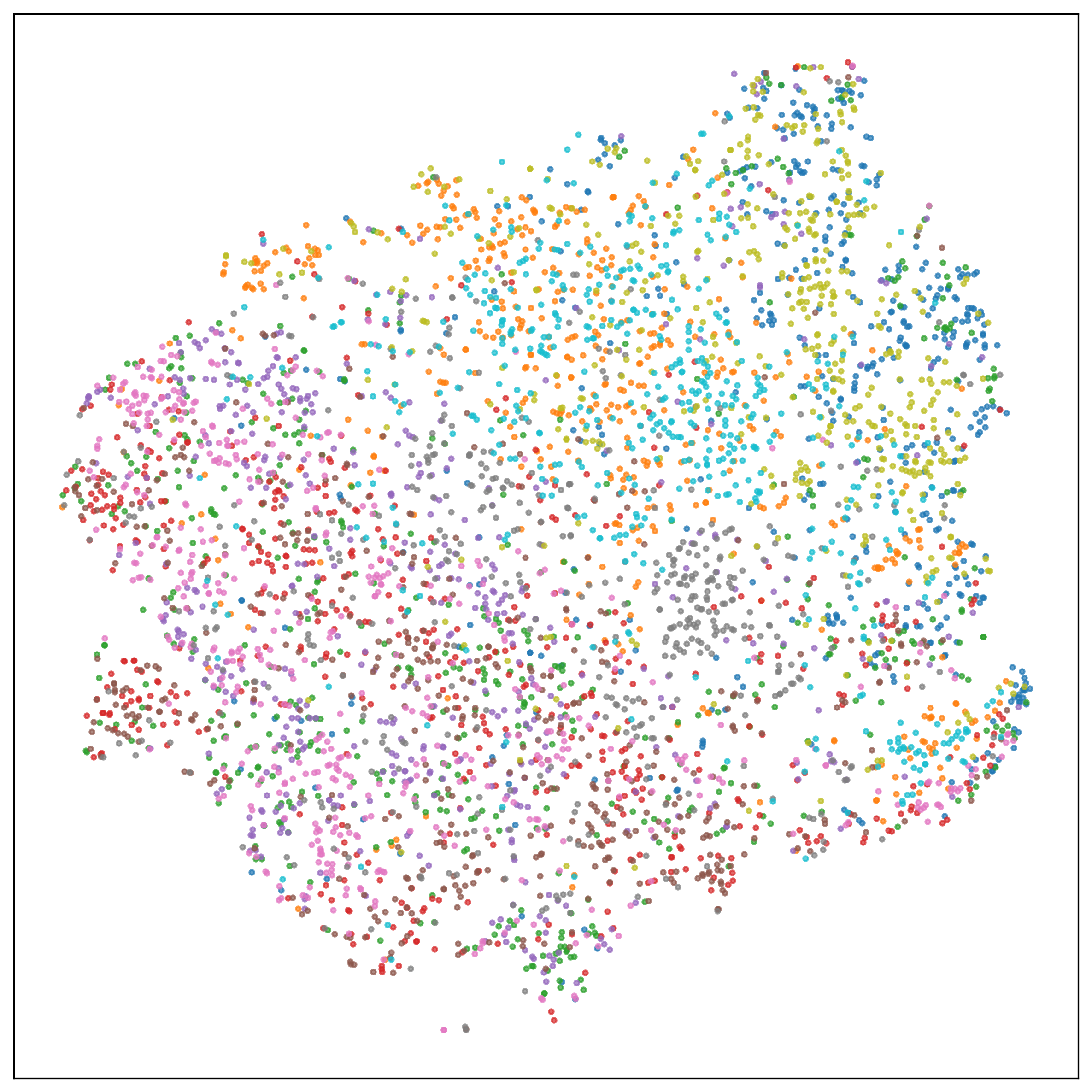}
&
\includegraphics[width=0.15\linewidth]{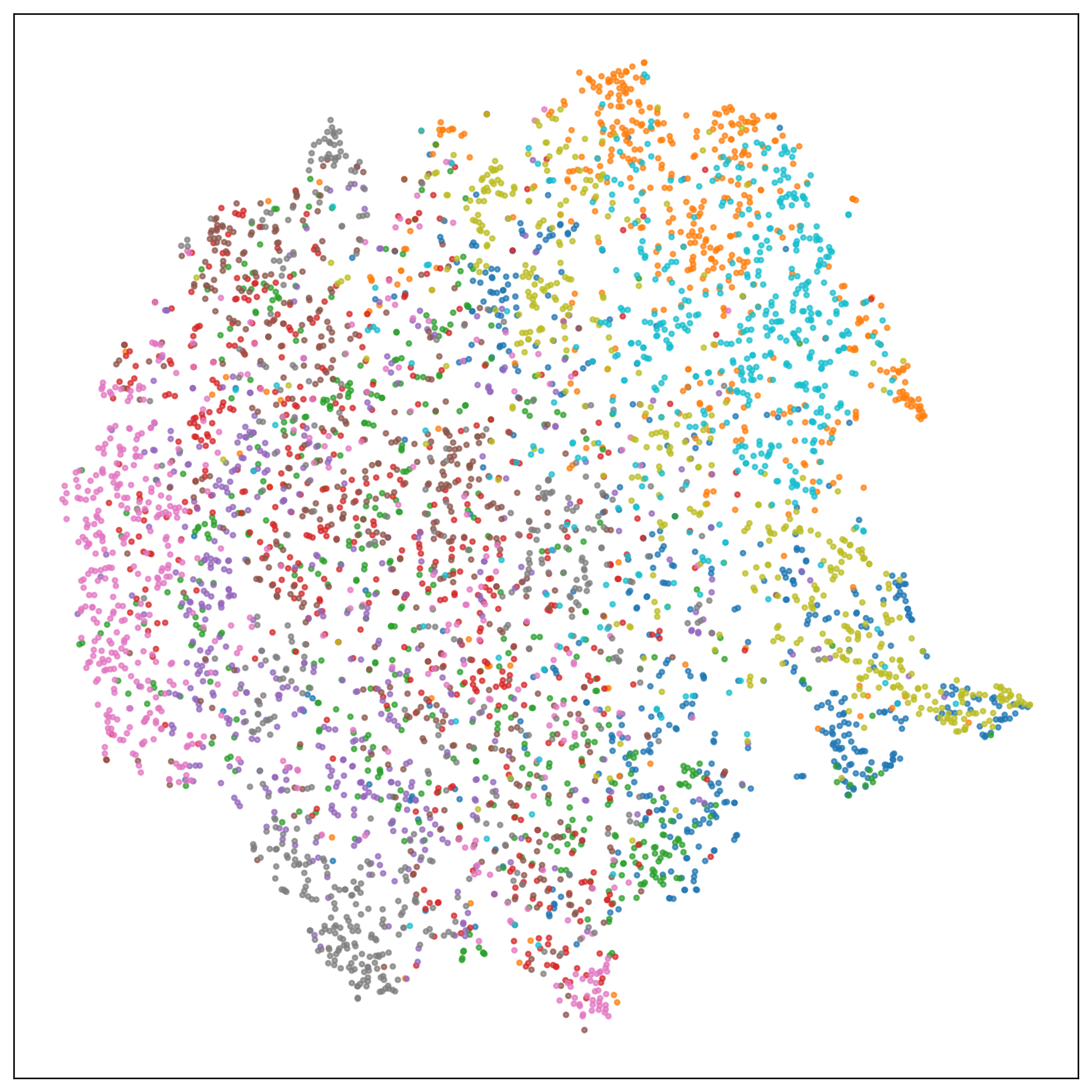}
&
\includegraphics[width=0.15\linewidth]{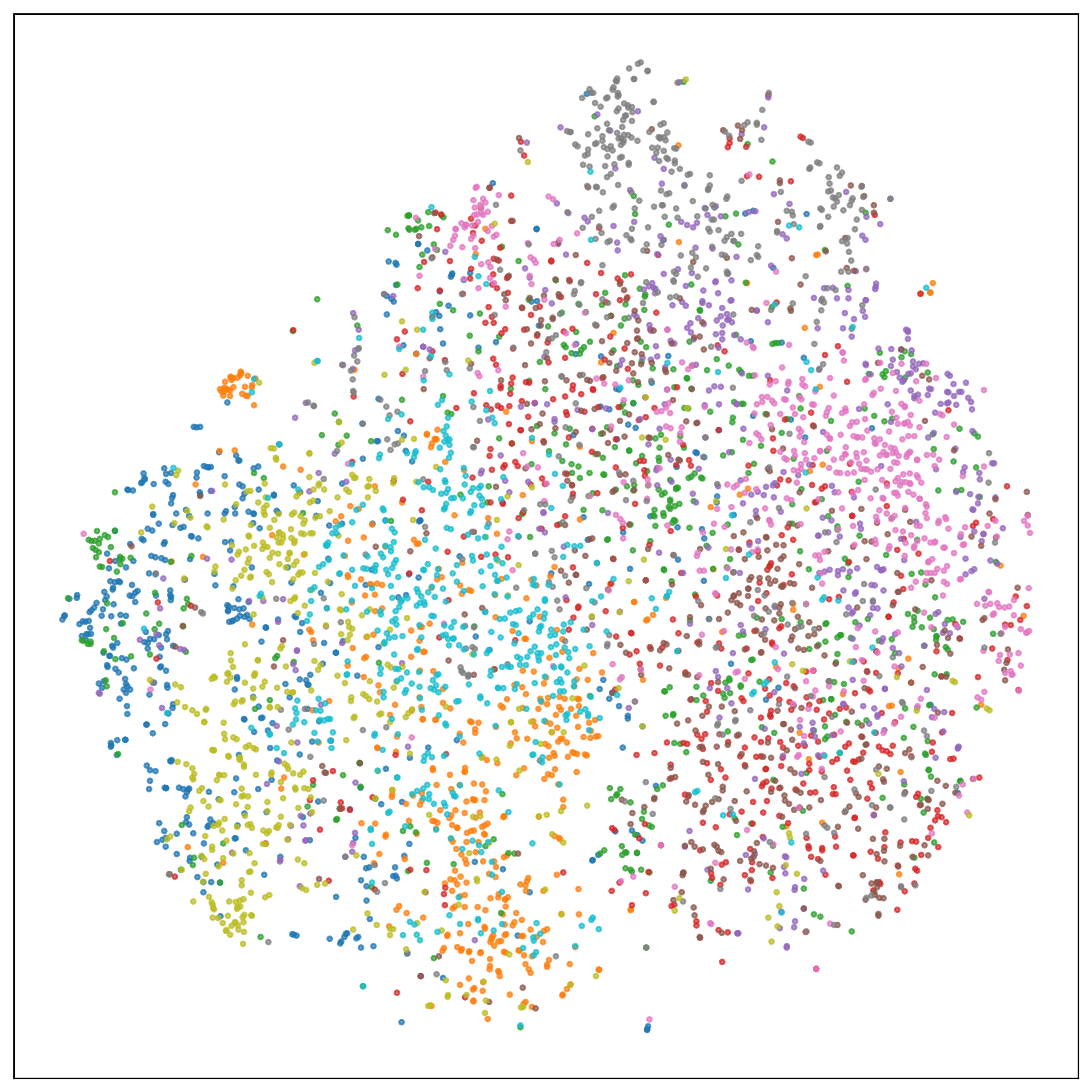}
\medskip\\
JT
&
\includegraphics[width=0.15\linewidth]{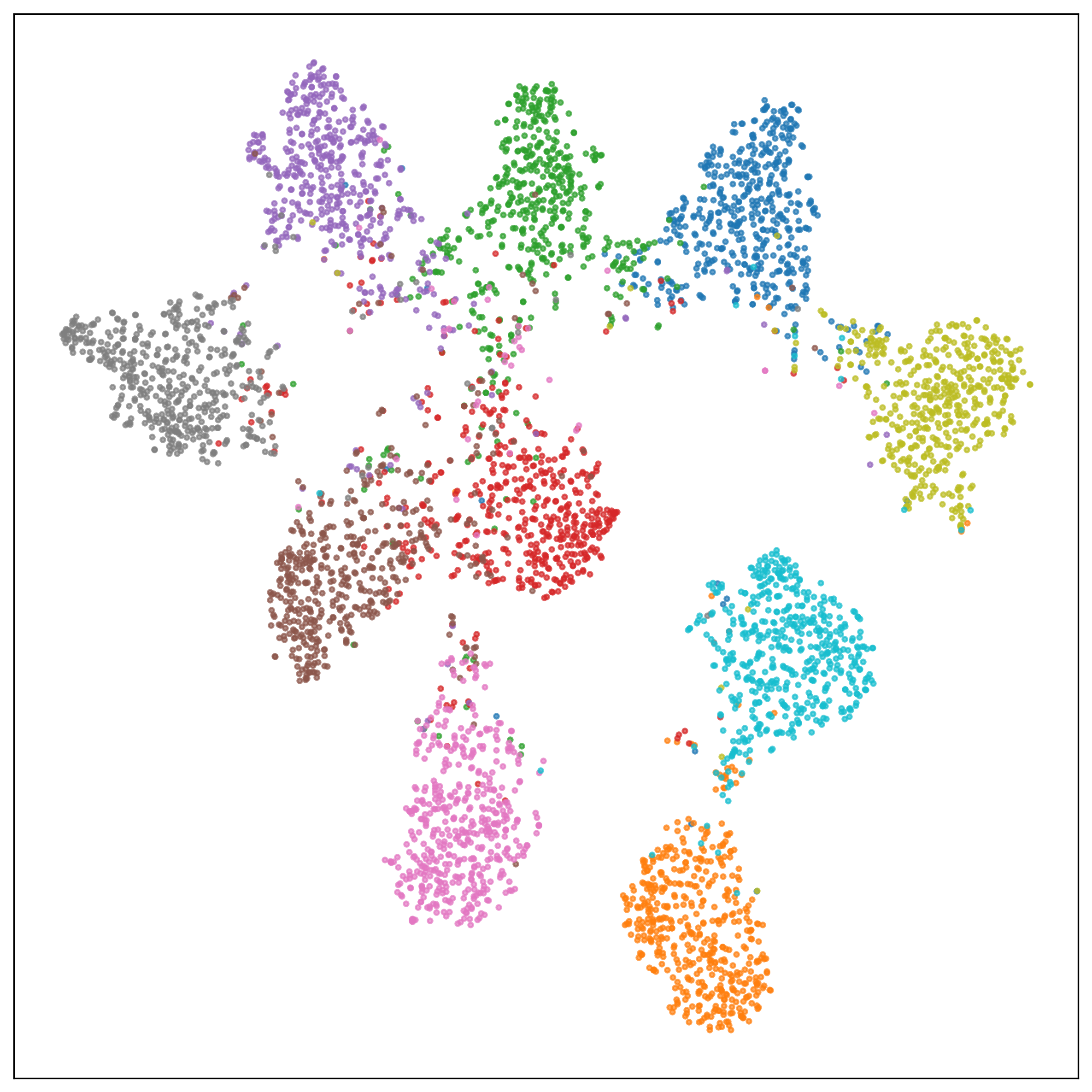}
&
\includegraphics[width=0.15\linewidth]{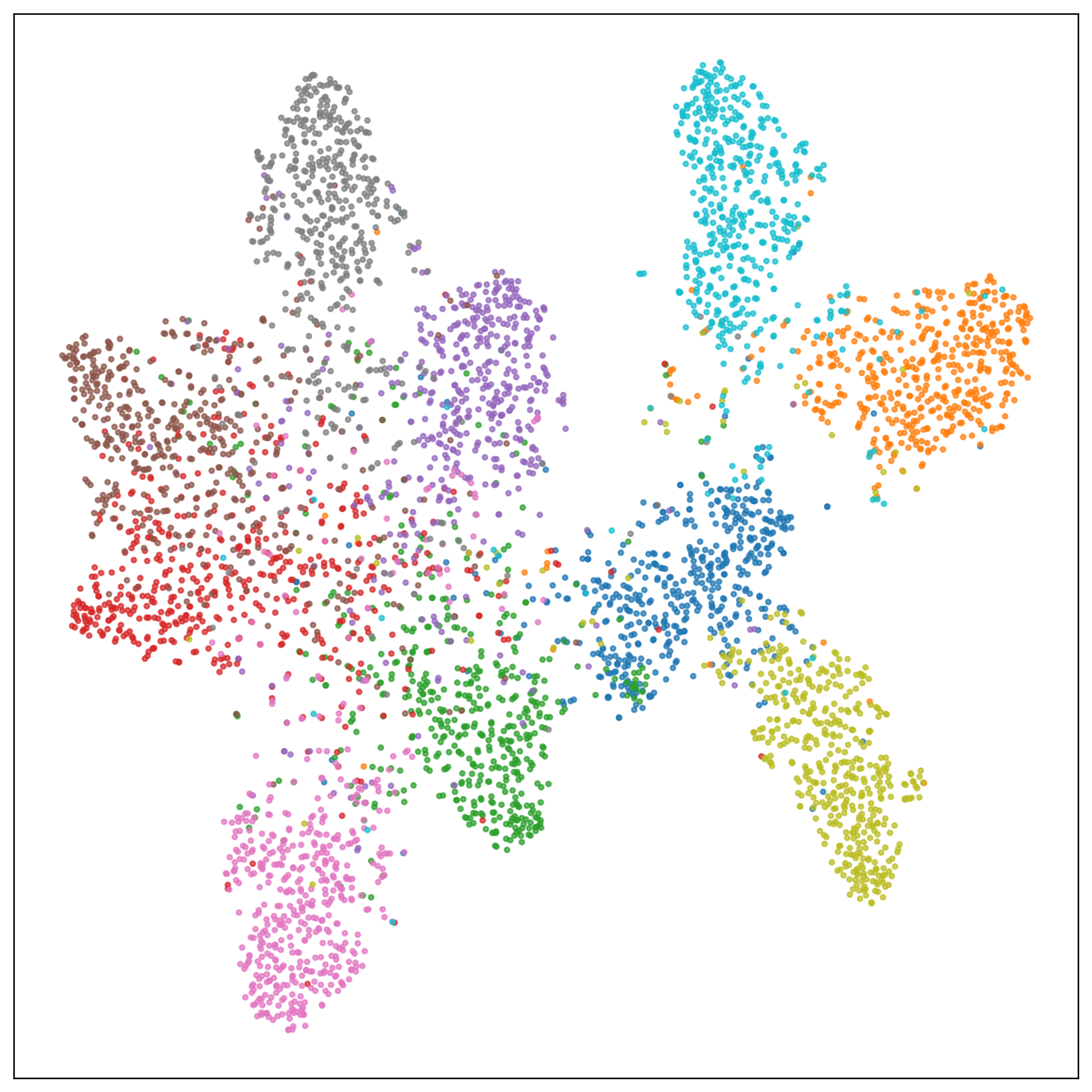}
&
\includegraphics[width=0.15\linewidth]{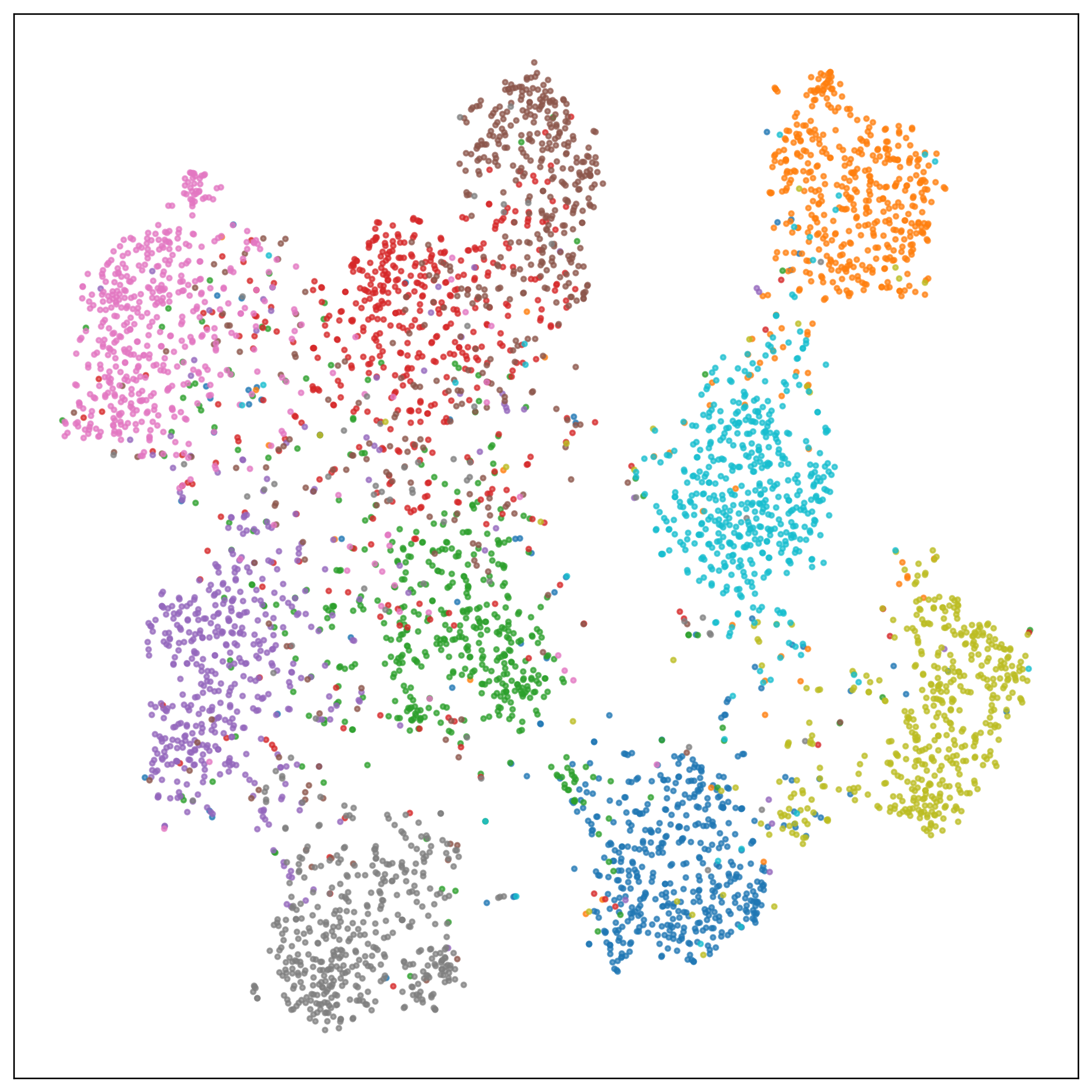}
&
\includegraphics[width=0.15\linewidth]{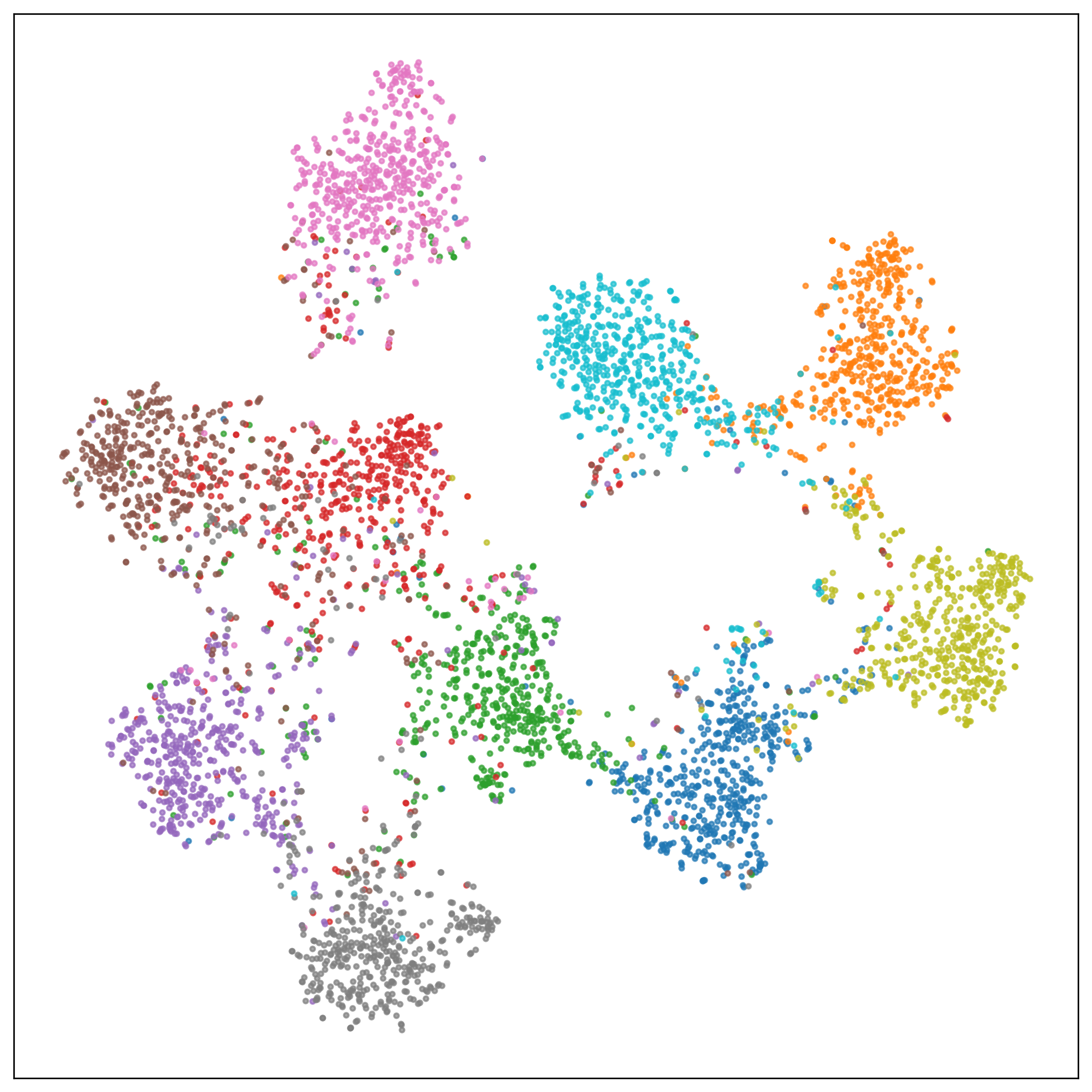}
&
\includegraphics[width=0.15\linewidth]{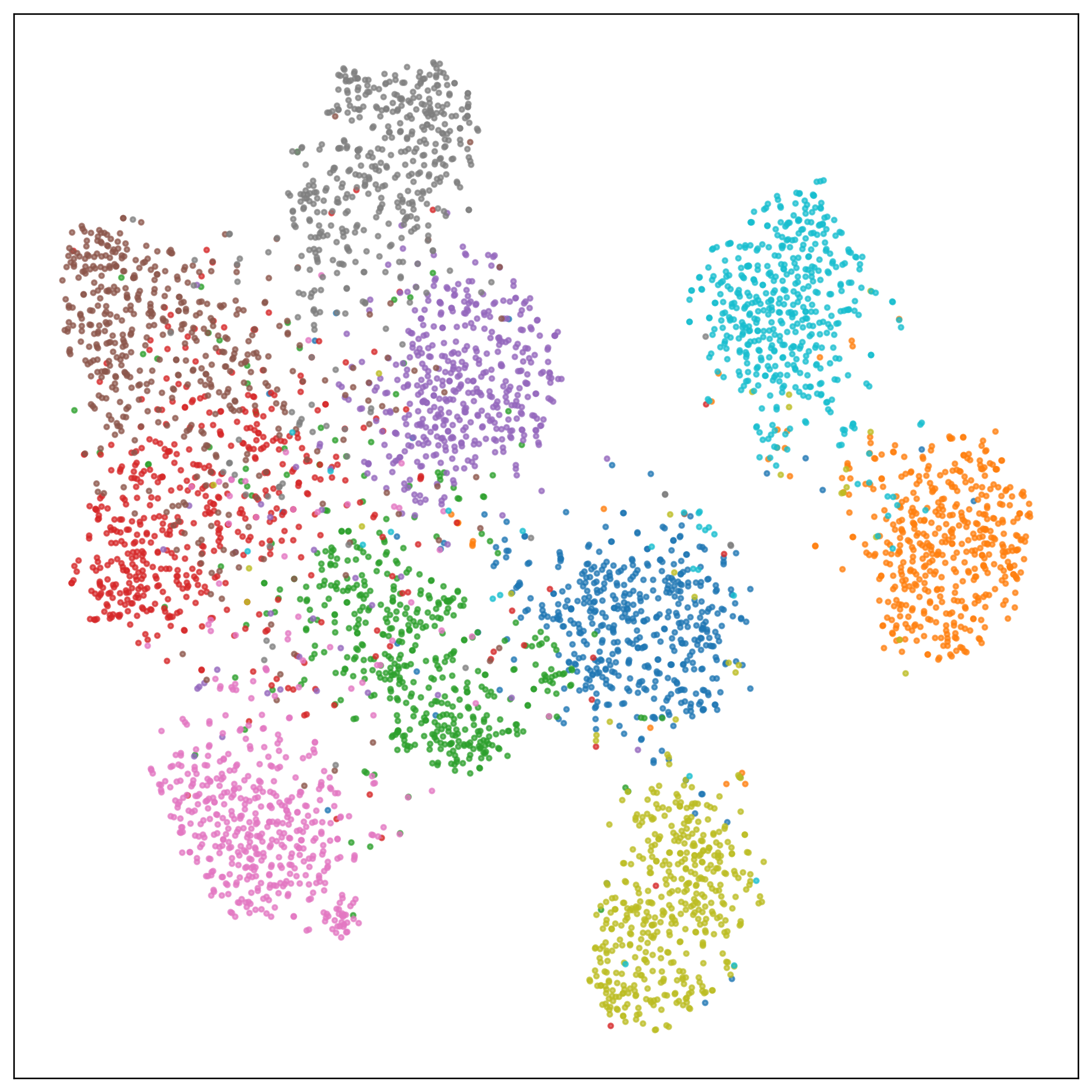}
&
\includegraphics[width=0.15\linewidth]{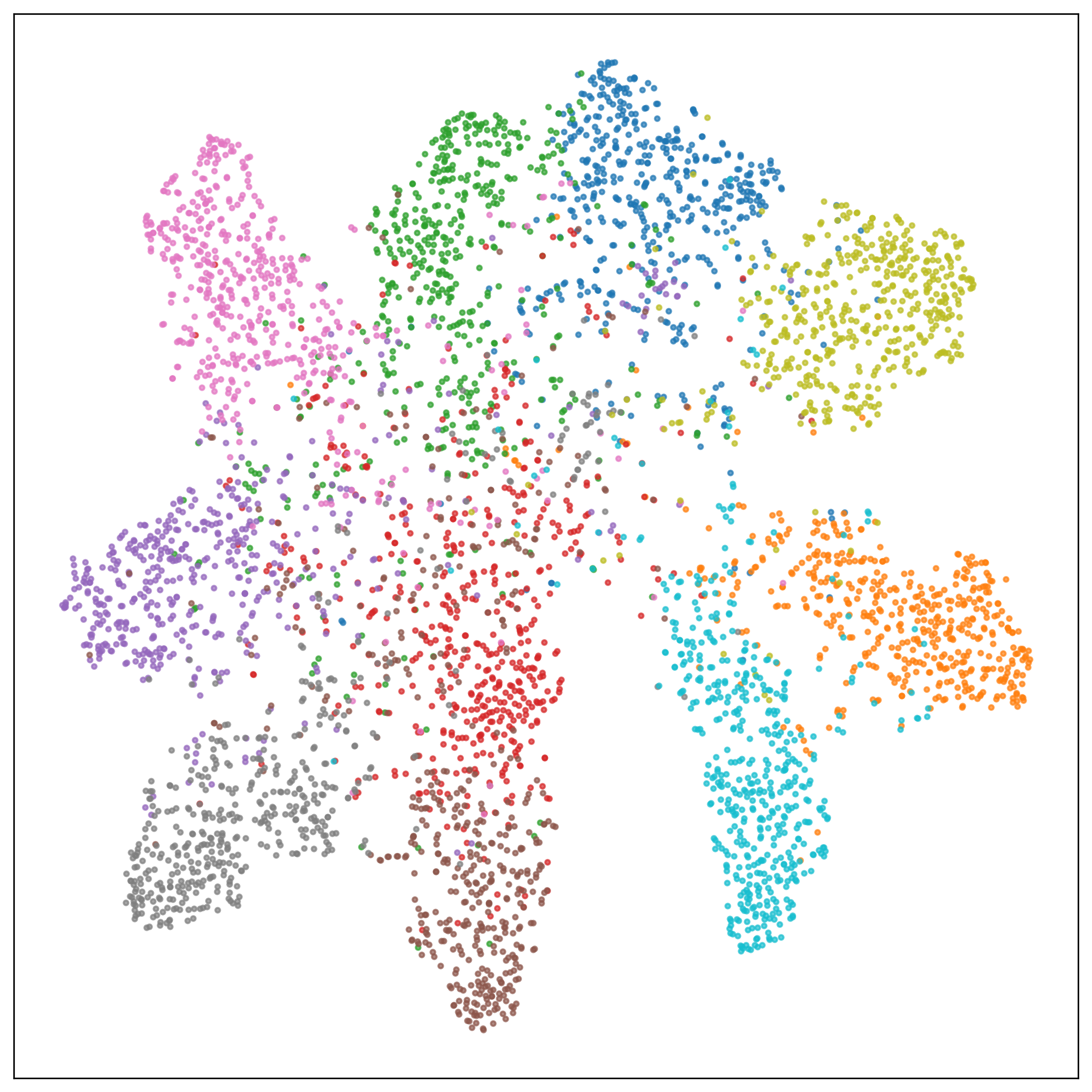}
&
\includegraphics[width=0.15\linewidth]{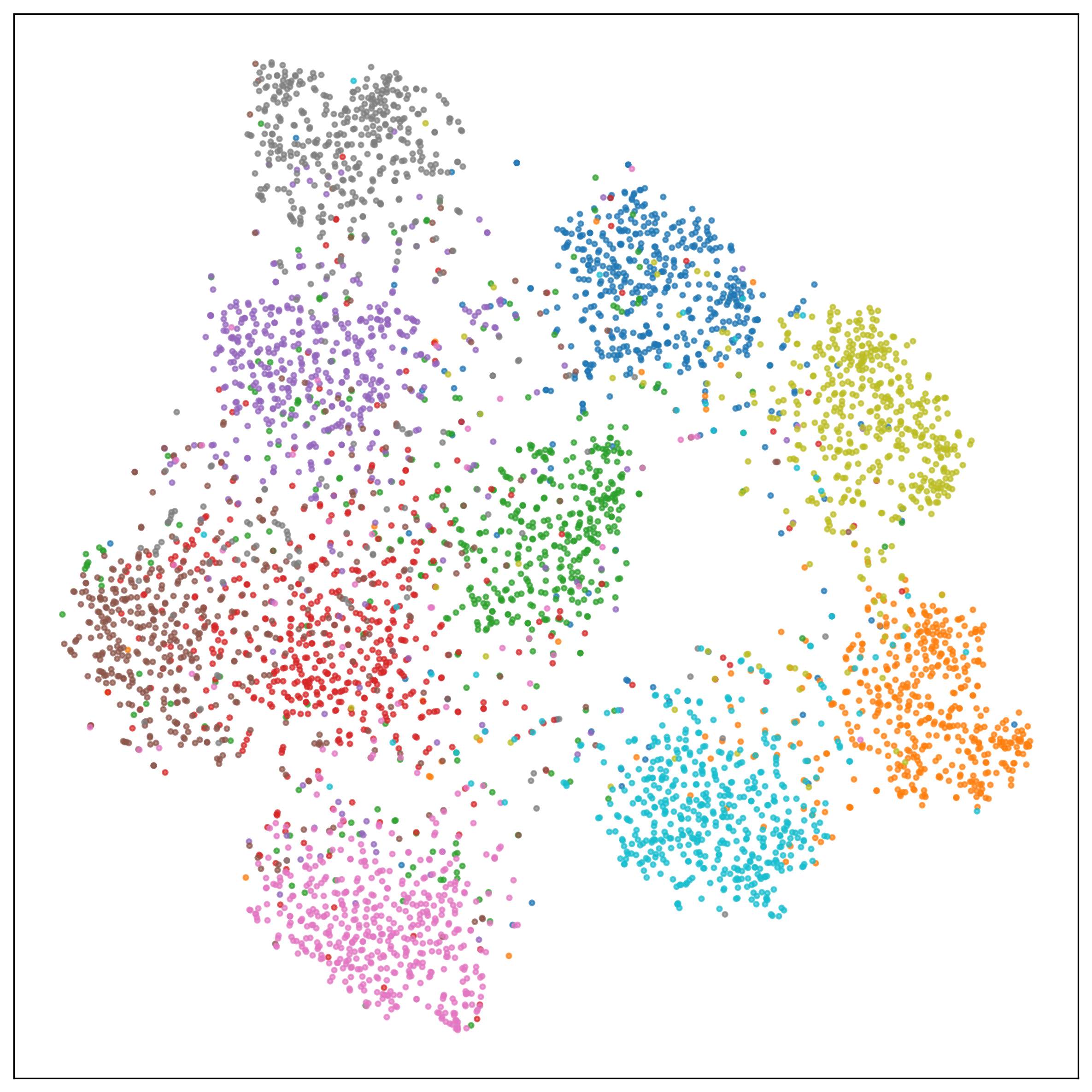}
&
\includegraphics[width=0.15\linewidth]{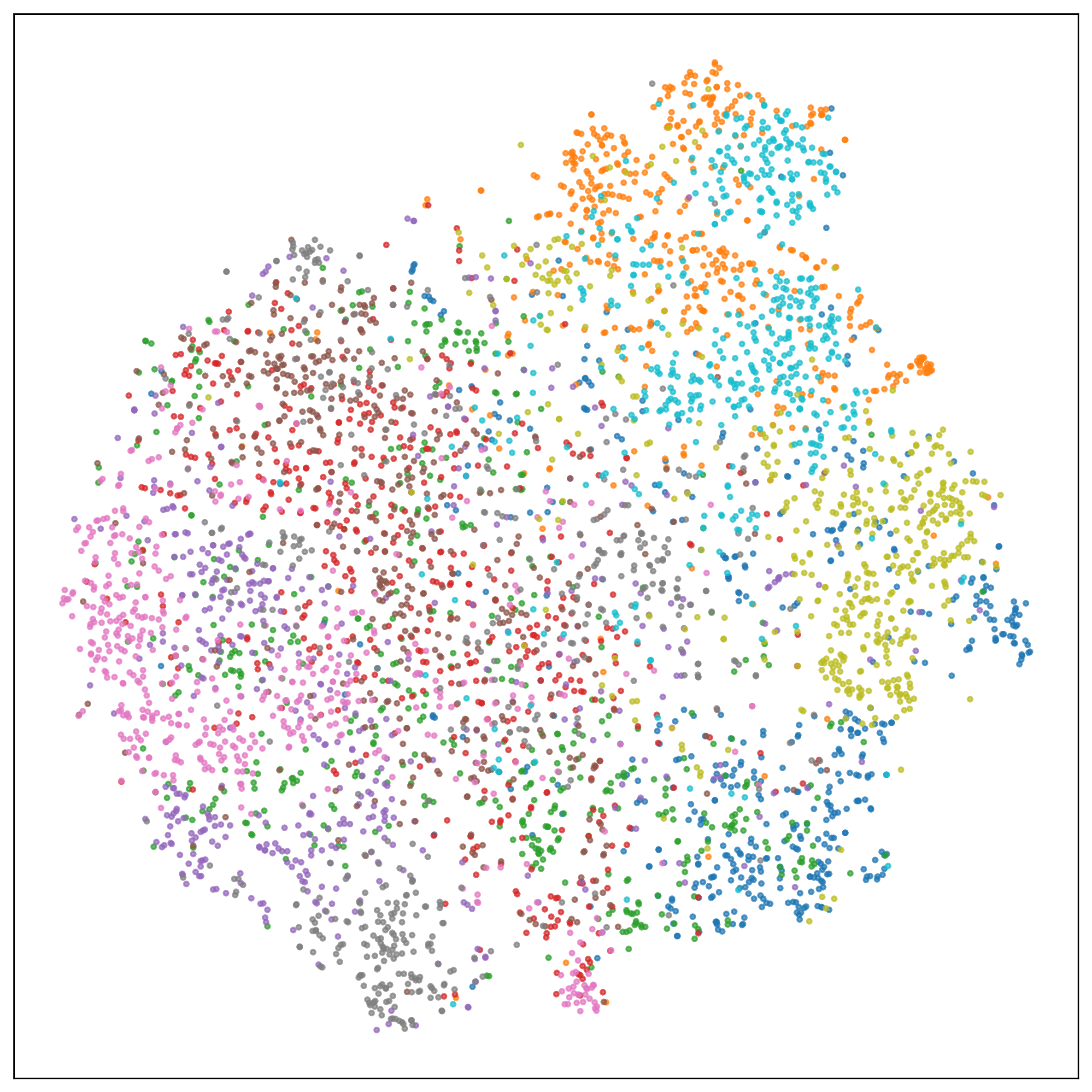}

\end{tabular}
}
\caption{t-SNE embeddings on CIFAR-10 for various SSL methods (PFT vs JT). Embedding color code --- Blue (airplane), orange (automobile), green (bird), red (cat), purple (deer), brown (dog), pink (frog), gray (horse), light green (ship), teal (truck).}
\label{fig:tsne_cifar10}
\end{figure}

\textbf{Feature Analysis:}
We also interpret the feature embeddings learned under different SSL methods and training paradigms (Fig.~\ref{fig:tsne_cifar10}). Across most methods, JT produces more compact and semantically separated class clusters compared with PFT, particularly for reconstruction-oriented approaches such as Colorization and Rotation. In contrast, several contrastive methods, including DINO and Barlow Twins, already exhibit relatively strong class separation under PFT, resulting in smaller visual differences between the two paradigms. These results further demonstrate that JT can improve representation organization and class-level feature discrimination in the low-level embedding space.

\section{Conclusion}
We have presented a systematic empirical study of two competing paradigms for integrating self-supervised and supervised learning: the conventional pretrain-finetune (PFT) and the relatively less explored joint training (JT). Through a comprehensive evaluation across a diverse collection of SSL objectives, datasets, and downstream tasks, including classification, detection, segmentation, and image quality assessment. We demonstrate that the relative effectiveness of these approaches is heavily influenced by the availability of labeled data, the complexity of the domain, and the specific characteristics of the task. Our findings suggest that JT consistently improves training efficiency and performs particularly well in low-label settings, where jointly optimizing supervised and self-supervised objectives can encourage more transferable and robust representations. At the same time, PFT remains a reliable strategy in more specialized or noisy domains. We believe that this work can serve as a practical reference for selecting effective SSL training strategies tailored to specific computer vision tasks, while also inspiring future research along multimodal self-supervision or deeper investigations into the theoretical foundations of joint representation learning.

\bibliographystyle{plainnat}
\bibliography{main}

\clearpage

\appendix

\section*{Appendices}
\startcontents[appendices]
\printcontents[appendices]{}{1}{\setcounter{tocdepth}{2}}

\clearpage

\section{Datasets}
\label{s_dataset}

\subsection{General Domain Datasets}
We evaluate our methods on several widely used benchmarks to ensure a comprehensive assessment across different domains.

\paragraph{CIFAR-10.} A small-scale natural image dataset containing 60,000 color images of size 32×32 pixels, evenly distributed across ten object categories~\cite{krizhevsky2009learning}. It is commonly used to benchmark representation learning performance and generalization under limited data.

\paragraph{COCO.} A large-scale natural image dataset containing annotations for object detection, image segmentation, and captioning tasks across 330,000 images ($>$200,000 labeled)~\cite{lin2014microsoft}. Each image is resized to 640$\times$640 pixels before being passed to the YOLOv12 model. COCO is one of the most popular benchmarks for object detection models.

\paragraph{PASCAL VOC2012.} This dataset contains approximately 20,000 bounding box-annotated images, split into 20 different classes. Along with COCO, it is one of the seminal benchmarks in the area of object detection ~\cite{everingham2012pascal}. Like COCO, before being passed to YOLOv12, the images are universally scaled to 640$\times$640 pixels.

\paragraph{KADID-10k.} An artificial-distortion IQA dataset containing 10,125 images, generated by applying 25 distortion types at 5 degradation levels to 81 reference images~\cite{lin2019kadid}. Distortions include blur, noise, color artifacts, compression, and various spatial degradations, which allow systematic evaluation of model robustness under known distortion categories. Image quality was rated through a subjective degradation category rating (DCR) study, where human observers assigned a 1–5 quality score depending on distortion severity. 

\paragraph{KonIQ-10k.} A large-scale authentic image quality assessment dataset containing 10,073 real-world images sourced from Flickr, selected to maximize diversity in content, aesthetics, and natural distortions~\cite{hosu2020koniq}. Each image is annotated using a crowdsourced subjective study where participants rated image quality on a five-level scale (1--5). Instead of providing a single score, KonIQ offers distributional annotations $(c_1\text{--}c_5)$, where each $c_i$ represents the fraction of raters who selected score $i$. To obtain a continuous quality value consistent with other IQA datasets, we compute the expected MOS score as: $\text{MOS}_{1\text{--}5}
= 1c_1 + 2c_2 + 3c_3 + 4c_4 + 5c_5$.
This produces a smooth 1--5 quality score aligned with KADID-10k’s rating range, which enables fair cross-dataset comparisons.

\subsection{Domain-Specific Datasets}
To examine cross-domain and task-specific generalization, we evaluate on datasets from three specialized domains: crisis event understanding, aerial geospatial imagery, and medical image segmentation.

\paragraph{CrisisMMD.} A multimodal social media dataset containing tweets and associated images collected during multiple natural disasters ~\cite{alam2018crisismmd}. We use the image-based informativeness classification task, which consists of two classes (informative vs.\ not-informative). All images are resized to $224\times224$ prior to training. We adopt the official train/validation/test split introduced by Gupta et al.~\cite{gupta2024crisiskan}, which contains 9,599 training samples, 1,573 validation samples, and 1,534 test samples.

\paragraph{DMD.} Contains 4,882 image–text pairs collected from multiple disaster events~\cite{mouzannar2018damage}. We use only the image modality and adopt the informativeness labels provided in the dataset to evaluate transfer performance from models trained on CrisisMMD.

\paragraph{EarthScape.} Is a multimodal geospatial dataset designed for Earth surface and surficial geologic mapping tasks~\cite{massey2025earthscape}. The dataset includes DEMs, aerial RGB+NIR imagery, multi-resolution terrain shape features, and GIS vector modalities associated with Earth surface geomorphic processes. The dataset exhibits substantial class imbalance and a long-tailed distribution across seven surficial geologic material classes. In addition, EarthScape contains two geographically separated study areas sharing the same label space, enabling evaluation of geographic domain shift and model robustness under covariate shift. In this study, a single-channel DEM input was used for downstream multi-label classification and semantic segmentation. We used 8,416 spatially independent training patches, 768 validation patches, and 1,536 test patches, with no geographic overlap between splits.

\paragraph{ISIC.} Contains dermoscopic images paired with binary lesion segmentation masks. We follow the official split, which includes 900 training images and 379 test images~\cite{gutman2016skin}. Segmentation performance is evaluated by training a U-Net decoder on top of SSL-pretrained encoders.

\paragraph{JSRT.} Includes 154 conventional chest radiographs with a lung nodule (100 malignant and 54 benign nodules) and 93 radiographs without a nodule~\cite{shiraishi2000development}. We utilize an image size of $224\times224$ when passing this data to our various models.

\paragraph{LDCTIQA.}
Includes low-dose abdominal CT images acquired under different radiation dose settings and annotated by expert radiologists for diagnostic image quality assessment~\cite{lee2023lowdosect}. The dataset is specifically designed for evaluating perceptual and diagnostic quality in low-dose CT reconstruction and enhancement tasks, containing paired low-quality and reference-quality CT scans from multiple patients. We resize all images to $224\times224$ before passing them to our various models.

\section{Evaluation Metrics}

In this appendix, we provide formulae and definitions for the metrics used in this paper to assess the performance of two competing training paradigms for SSL across a variety of computer vision tasks. In each applicable equation, true positives/negatives and false positives/negatives are denoted by TP/TN and FP/FN, respectively.

\subsection{Classification}

For classification, we evaluated the various SSL techniques and their PFT and JT performance using the following metrics:

\begin{enumerate}
    \item \textbf{Accuracy:} The proportion of correct predictions made by the model out of all predictions.
    \begin{equation}
        Accuracy = \frac{TP + TN}{TP + TN + FP + FN}.
    \end{equation}
    \item \textbf{F1 Score:} The harmonic mean of precision and recall.
    \begin{equation}
        F1 = \frac{2TP}{2TP + FP +FN}.
    \end{equation}
\end{enumerate}

\subsection{Segmentation} 

For segmentation, the following metrics were used:

\begin{enumerate}
    \item \textbf{Dice-Sørensen Coefficient:} Also known as the Dice similarity coefficient or Dice score, this metric measures the similarity or overlap between two sets of data (denoted by X and Y in the equation) and is often used in image segmentation to compare predicted results with the ground truth.
    \begin{equation}
        \frac{2|X \cap Y|}{|X| + |Y|}.
    \end{equation}
    \item \textbf{Mean Intersection-over-Union (mIoU):} The average of IoU scores calculated for each class in a dataset. IoU, also called the Jaccard index, measures the average overlap between predicted masks and the ground truth. It is calculated via the following equation:
    \begin{equation}
        \frac{|X \cap Y|}{|X \cup Y|}.
    \end{equation}
\end{enumerate}

\subsection{Object Detection}

For object detection, we used:

\begin{enumerate}
    \item \textbf{Precision:} Measures how many of the positive predictions made by the model are actually correct.
    \begin{equation}
        P = \frac{TP}{TP + FP}.
    \end{equation}
    \item \textbf{Recall:} Also called sensitivity, this metric measures how many of the actual positive cases were correctly identified by the model. 
    \begin{equation}
       R = \frac{TP}{TP + FN}.
    \end{equation}
    \item \textbf{Mean Average Precision (mAP):} A standard metric for evaluating object detection models, mAP summarizes the precision–recall trade-off across all object classes. It is computed as the mean of the Average Precision (AP) over $N$ classes:
    \begin{equation}
        mAP = \frac{1}{N} \sum_{i=1}^{N} AP_i,
    \end{equation}
    where $AP_i$ denotes the AP for class $i$. AP measures the area under the precision–recall curve for a given class. It is computed by integrating precision as a function of recall:
    \begin{equation}
        AP = \int_{0}^{1} p(r)\, dr
    \end{equation}
    where $p(r)$ denotes precision as a function of recall $r$. In practice, this is approximated using a finite set of recall levels. In this work, we report both mAP@50 and mAP@50--95. mAP@50 computes the mean AP at a single IoU threshold of 0.5, while mAP@50--95 averages AP over multiple IoU thresholds ranging from 0.5 to 0.95 in increments of 0.05, providing a more comprehensive evaluation of detection performance.
\end{enumerate}

\subsection{Image Quality Assessment}

For assessing image quality, we used the following metrics:

\begin{enumerate}
    \item \textbf{Spearman Rank Correlation Coefficient (SROCC):} Measures the monotonic relationship between predicted and ground-truth rankings.
    \begin{equation}
    SROCC = 1 - \frac{6 \sum_{i = 1}^{n} d_i^2}{n (n^2 - 1)},
    \end{equation}
    where $d_i$ is is the difference between the ranks of $y_i$ and $\hat{y}$.
    \item \textbf{Pearson Linear Correlation Coefficient (PLCC):} Measures the linear correlation between predicted and ground-truth scores.
    \begin{equation}
    PLCC = \frac{
    \sum_{i = 1}^{n} (y_i - \bar{y})(\hat{y}_i - \bar{\hat{y}})
    }{
    \sum_{i = 1}^{n} (y_i - \bar{y})^2 \;
    \sum_{i = 1}^{n} (\hat{y}_i - \bar{\hat{y}})^2,
    }
    \end{equation}
    where $y_i$ and $\hat{y}_i$ denote the ground truth and predicted scores for sample $i$, and $\bar{y}$ and $\bar{\hat{y}}$ are their respective means.
\end{enumerate}

\clearpage

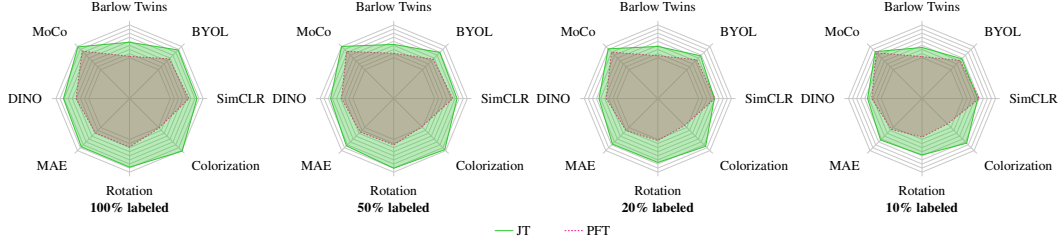
\begin{figure}
\resizebox{\textwidth}{!}{
\begin{tikzpicture}
 
\definecolor{pftpink}{RGB}{255,20,147}
\definecolor{jtlime}{RGB}{50,200,50}
\definecolor{gridc}{RGB}{200,200,200}
 
\newcommand{\drawpanel}[1]{
  \foreach \v/\lab in {0.50/0.50, 0.55/0.55, 0.60/0.60, 0.65/0.65, 0.70/0.70, 0.75/0.75, 0.80/0.80, 0.85/0.85, 0.90/0.90}{
    \pgfmathsetmacro{\rr}{\v/0.95*5}
    \draw[gridc,thin]
      (90:\rr cm)--(45:\rr cm)--(0:\rr cm)--(-45:\rr cm)--
      (-90:\rr cm)--(-135:\rr cm)--(180:\rr cm)--(135:\rr cm)--cycle;
  }
  \foreach \i in {0,...,7}{
    \pgfmathsetmacro{\ang}{90-\i*45}
    \draw[gridc,thin] (0,0)--(\ang:5cm);
  }
  \node[font=\Huge,above]       at (90:5.5cm)   {Barlow Twins};
  \node[font=\Huge,above right] at (45:5.5cm)   {BYOL};
  \node[font=\Huge,right]       at (0:5.5cm)    {SimCLR};
  \node[font=\Huge,below right] at (-45:5.5cm)  {Colorization};
  \node[font=\Huge,below]       at (-90:5.5cm)  {Rotation};
  \node[font=\Huge,below left]  at (-135:5.5cm) {MAE};
  \node[font=\Huge,left]        at (180:5.5cm)  {DINO};
  \node[font=\Huge,above left]  at (135:5.5cm)  {MoCo};
}

\begin{scope}[xshift=0cm]
  \drawpanel{}
  \pgfmathsetmacro{\Pax}{0.5163/0.95*5*cos(90)}  \pgfmathsetmacro{\Pay}{0.5163/0.95*5*sin(90)}
  \pgfmathsetmacro{\Pbx}{0.6855/0.95*5*cos(45)}  \pgfmathsetmacro{\Pby}{0.6855/0.95*5*sin(45)}
  \pgfmathsetmacro{\Pcx}{0.7269/0.95*5*cos(0)}   \pgfmathsetmacro{\Pcy}{0.7269/0.95*5*sin(0)}
  \pgfmathsetmacro{\Pdx}{0.509/0.95*5*cos(-45)}  \pgfmathsetmacro{\Pdy}{0.509/0.95*5*sin(-45)}
  \pgfmathsetmacro{\Pex}{0.5926/0.95*5*cos(-90)} \pgfmathsetmacro{\Pey}{0.5926/0.95*5*sin(-90)}
  \pgfmathsetmacro{\Pfx}{0.592/0.95*5*cos(-135)} \pgfmathsetmacro{\Pfy}{0.592/0.95*5*sin(-135)}
  \pgfmathsetmacro{\Pgx}{0.6541/0.95*5*cos(180)} \pgfmathsetmacro{\Pgy}{0.6541/0.95*5*sin(180)}
  \pgfmathsetmacro{\Phx}{0.8205/0.95*5*cos(135)} \pgfmathsetmacro{\Phy}{0.8205/0.95*5*sin(135)}
  \fill[pftpink,opacity=0.35]
    (\Pax cm,\Pay cm)--(\Pbx cm,\Pby cm)--(\Pcx cm,\Pcy cm)--(\Pdx cm,\Pdy cm)--
    (\Pex cm,\Pey cm)--(\Pfx cm,\Pfy cm)--(\Pgx cm,\Pgy cm)--(\Phx cm,\Phy cm)--cycle;
  \draw[pftpink,thick,dashed]
    (\Pax cm,\Pay cm)--(\Pbx cm,\Pby cm)--(\Pcx cm,\Pcy cm)--(\Pdx cm,\Pdy cm)--
    (\Pex cm,\Pey cm)--(\Pfx cm,\Pfy cm)--(\Pgx cm,\Pgy cm)--(\Phx cm,\Phy cm)--cycle;
  \pgfmathsetmacro{\Jax}{0.6883/0.95*5*cos(90)}  \pgfmathsetmacro{\Jay}{0.6883/0.95*5*sin(90)}
  \pgfmathsetmacro{\Jbx}{0.84/0.95*5*cos(45)}    \pgfmathsetmacro{\Jby}{0.84/0.95*5*sin(45)}
  \pgfmathsetmacro{\Jcx}{0.825/0.95*5*cos(0)}    \pgfmathsetmacro{\Jcy}{0.825/0.95*5*sin(0)}
  \pgfmathsetmacro{\Jdx}{0.9099/0.95*5*cos(-45)} \pgfmathsetmacro{\Jdy}{0.9099/0.95*5*sin(-45)}
  \pgfmathsetmacro{\Jex}{0.8387/0.95*5*cos(-90)} \pgfmathsetmacro{\Jey}{0.8387/0.95*5*sin(-90)}
  \pgfmathsetmacro{\Jfx}{0.8341/0.95*5*cos(-135)}\pgfmathsetmacro{\Jfy}{0.8341/0.95*5*sin(-135)}
  \pgfmathsetmacro{\Jgx}{0.8079/0.95*5*cos(180)} \pgfmathsetmacro{\Jgy}{0.8079/0.95*5*sin(180)}
  \pgfmathsetmacro{\Jhx}{0.8938/0.95*5*cos(135)} \pgfmathsetmacro{\Jhy}{0.8938/0.95*5*sin(135)}
  \fill[jtlime,opacity=0.35]
    (\Jax cm,\Jay cm)--(\Jbx cm,\Jby cm)--(\Jcx cm,\Jcy cm)--(\Jdx cm,\Jdy cm)--
    (\Jex cm,\Jey cm)--(\Jfx cm,\Jfy cm)--(\Jgx cm,\Jgy cm)--(\Jhx cm,\Jhy cm)--cycle;
  \draw[jtlime,thick]
    (\Jax cm,\Jay cm)--(\Jbx cm,\Jby cm)--(\Jcx cm,\Jcy cm)--(\Jdx cm,\Jdy cm)--
    (\Jex cm,\Jey cm)--(\Jfx cm,\Jfy cm)--(\Jgx cm,\Jgy cm)--(\Jhx cm,\Jhy cm)--cycle;
  \node[font=\Huge\bfseries] at (0,-7cm) {100\% labeled};
\end{scope}

\begin{scope}[xshift=17cm]
  \drawpanel{}
  \pgfmathsetmacro{\Pax}{0.5537/0.95*5*cos(90)}  \pgfmathsetmacro{\Pay}{0.5537/0.95*5*sin(90)}
  \pgfmathsetmacro{\Pbx}{0.6817/0.95*5*cos(45)}  \pgfmathsetmacro{\Pby}{0.6817/0.95*5*sin(45)}
  \pgfmathsetmacro{\Pcx}{0.7225/0.95*5*cos(0)}   \pgfmathsetmacro{\Pcy}{0.7225/0.95*5*sin(0)}
  \pgfmathsetmacro{\Pdx}{0.4864/0.95*5*cos(-45)} \pgfmathsetmacro{\Pdy}{0.4864/0.95*5*sin(-45)}
  \pgfmathsetmacro{\Pex}{0.5669/0.95*5*cos(-90)} \pgfmathsetmacro{\Pey}{0.5669/0.95*5*sin(-90)}
  \pgfmathsetmacro{\Pfx}{0.5797/0.95*5*cos(-135)}\pgfmathsetmacro{\Pfy}{0.5797/0.95*5*sin(-135)}
  \pgfmathsetmacro{\Pgx}{0.6382/0.95*5*cos(180)} \pgfmathsetmacro{\Pgy}{0.6382/0.95*5*sin(180)}
  \pgfmathsetmacro{\Phx}{0.8156/0.95*5*cos(135)} \pgfmathsetmacro{\Phy}{0.8156/0.95*5*sin(135)}
  \fill[pftpink,opacity=0.35]
    (\Pax cm,\Pay cm)--(\Pbx cm,\Pby cm)--(\Pcx cm,\Pcy cm)--(\Pdx cm,\Pdy cm)--
    (\Pex cm,\Pey cm)--(\Pfx cm,\Pfy cm)--(\Pgx cm,\Pgy cm)--(\Phx cm,\Phy cm)--cycle;
  \draw[pftpink,thick,dashed]
    (\Pax cm,\Pay cm)--(\Pbx cm,\Pby cm)--(\Pcx cm,\Pcy cm)--(\Pdx cm,\Pdy cm)--
    (\Pex cm,\Pey cm)--(\Pfx cm,\Pfy cm)--(\Pgx cm,\Pgy cm)--(\Phx cm,\Phy cm)--cycle;
  \pgfmathsetmacro{\Jax}{0.662/0.95*5*cos(90)}   \pgfmathsetmacro{\Jay}{0.662/0.95*5*sin(90)}
  \pgfmathsetmacro{\Jbx}{0.798/0.95*5*cos(45)}   \pgfmathsetmacro{\Jby}{0.798/0.95*5*sin(45)}
  \pgfmathsetmacro{\Jcx}{0.7745/0.95*5*cos(0)}   \pgfmathsetmacro{\Jcy}{0.7745/0.95*5*sin(0)}
  \pgfmathsetmacro{\Jdx}{0.8835/0.95*5*cos(-45)} \pgfmathsetmacro{\Jdy}{0.8835/0.95*5*sin(-45)}
  \pgfmathsetmacro{\Jex}{0.8523/0.95*5*cos(-90)} \pgfmathsetmacro{\Jey}{0.8523/0.95*5*sin(-90)}
  \pgfmathsetmacro{\Jfx}{0.8179/0.95*5*cos(-135)}\pgfmathsetmacro{\Jfy}{0.8179/0.95*5*sin(-135)}
  \pgfmathsetmacro{\Jgx}{0.772/0.95*5*cos(180)}  \pgfmathsetmacro{\Jgy}{0.772/0.95*5*sin(180)}
  \pgfmathsetmacro{\Jhx}{0.9009/0.95*5*cos(135)} \pgfmathsetmacro{\Jhy}{0.9009/0.95*5*sin(135)}
  \fill[jtlime,opacity=0.35]
    (\Jax cm,\Jay cm)--(\Jbx cm,\Jby cm)--(\Jcx cm,\Jcy cm)--(\Jdx cm,\Jdy cm)--
    (\Jex cm,\Jey cm)--(\Jfx cm,\Jfy cm)--(\Jgx cm,\Jgy cm)--(\Jhx cm,\Jhy cm)--cycle;
  \draw[jtlime,thick]
    (\Jax cm,\Jay cm)--(\Jbx cm,\Jby cm)--(\Jcx cm,\Jcy cm)--(\Jdx cm,\Jdy cm)--
    (\Jex cm,\Jey cm)--(\Jfx cm,\Jfy cm)--(\Jgx cm,\Jgy cm)--(\Jhx cm,\Jhy cm)--cycle;
  \node[font=\Huge\bfseries] at (0,-7cm) {50\% labeled};
\end{scope}
 
\begin{scope}[xshift=34cm]
  \drawpanel{}
  \pgfmathsetmacro{\Pax}{0.5242/0.95*5*cos(90)}  \pgfmathsetmacro{\Pay}{0.5242/0.95*5*sin(90)}
  \pgfmathsetmacro{\Pbx}{0.6713/0.95*5*cos(45)}  \pgfmathsetmacro{\Pby}{0.6713/0.95*5*sin(45)}
  \pgfmathsetmacro{\Pcx}{0.6968/0.95*5*cos(0)}   \pgfmathsetmacro{\Pcy}{0.6968/0.95*5*sin(0)}
  \pgfmathsetmacro{\Pdx}{0.4637/0.95*5*cos(-45)} \pgfmathsetmacro{\Pdy}{0.4637/0.95*5*sin(-45)}
  \pgfmathsetmacro{\Pex}{0.5136/0.95*5*cos(-90)} \pgfmathsetmacro{\Pey}{0.5136/0.95*5*sin(-90)}
  \pgfmathsetmacro{\Pfx}{0.552/0.95*5*cos(-135)} \pgfmathsetmacro{\Pfy}{0.552/0.95*5*sin(-135)}
  \pgfmathsetmacro{\Pgx}{0.632/0.95*5*cos(180)}  \pgfmathsetmacro{\Pgy}{0.632/0.95*5*sin(180)}
  \pgfmathsetmacro{\Phx}{0.8019/0.95*5*cos(135)} \pgfmathsetmacro{\Phy}{0.8019/0.95*5*sin(135)}
  \fill[pftpink,opacity=0.35]
    (\Pax cm,\Pay cm)--(\Pbx cm,\Pby cm)--(\Pcx cm,\Pcy cm)--(\Pdx cm,\Pdy cm)--
    (\Pex cm,\Pey cm)--(\Pfx cm,\Pfy cm)--(\Pgx cm,\Pgy cm)--(\Phx cm,\Phy cm)--cycle;
  \draw[pftpink,thick,dashed]
    (\Pax cm,\Pay cm)--(\Pbx cm,\Pby cm)--(\Pcx cm,\Pcy cm)--(\Pdx cm,\Pdy cm)--
    (\Pex cm,\Pey cm)--(\Pfx cm,\Pfy cm)--(\Pgx cm,\Pgy cm)--(\Phx cm,\Phy cm)--cycle;
  \pgfmathsetmacro{\Jax}{0.6373/0.95*5*cos(90)}  \pgfmathsetmacro{\Jay}{0.6373/0.95*5*sin(90)}
  \pgfmathsetmacro{\Jbx}{0.739/0.95*5*cos(45)}   \pgfmathsetmacro{\Jby}{0.739/0.95*5*sin(45)}
  \pgfmathsetmacro{\Jcx}{0.6905/0.95*5*cos(0)}   \pgfmathsetmacro{\Jcy}{0.6905/0.95*5*sin(0)}
  \pgfmathsetmacro{\Jdx}{0.826/0.95*5*cos(-45)}  \pgfmathsetmacro{\Jdy}{0.826/0.95*5*sin(-45)}
  \pgfmathsetmacro{\Jex}{0.7816/0.95*5*cos(-90)} \pgfmathsetmacro{\Jey}{0.7816/0.95*5*sin(-90)}
  \pgfmathsetmacro{\Jfx}{0.787/0.95*5*cos(-135)} \pgfmathsetmacro{\Jfy}{0.787/0.95*5*sin(-135)}
  \pgfmathsetmacro{\Jgx}{0.7201/0.95*5*cos(180)} \pgfmathsetmacro{\Jgy}{0.7201/0.95*5*sin(180)}
  \pgfmathsetmacro{\Jhx}{0.8594/0.95*5*cos(135)} \pgfmathsetmacro{\Jhy}{0.8594/0.95*5*sin(135)}
  \fill[jtlime,opacity=0.35]
    (\Jax cm,\Jay cm)--(\Jbx cm,\Jby cm)--(\Jcx cm,\Jcy cm)--(\Jdx cm,\Jdy cm)--
    (\Jex cm,\Jey cm)--(\Jfx cm,\Jfy cm)--(\Jgx cm,\Jgy cm)--(\Jhx cm,\Jhy cm)--cycle;
  \draw[jtlime,thick]
    (\Jax cm,\Jay cm)--(\Jbx cm,\Jby cm)--(\Jcx cm,\Jcy cm)--(\Jdx cm,\Jdy cm)--
    (\Jex cm,\Jey cm)--(\Jfx cm,\Jfy cm)--(\Jgx cm,\Jgy cm)--(\Jhx cm,\Jhy cm)--cycle;
  \node[font=\Huge\bfseries] at (0,-7cm) {20\% labeled};
\end{scope}

\begin{scope}[xshift=51cm]
  \drawpanel{}
  \pgfmathsetmacro{\Pax}{0.5124/0.95*5*cos(90)}  \pgfmathsetmacro{\Pay}{0.5124/0.95*5*sin(90)}
  \pgfmathsetmacro{\Pbx}{0.6586/0.95*5*cos(45)}  \pgfmathsetmacro{\Pby}{0.6586/0.95*5*sin(45)}
  \pgfmathsetmacro{\Pcx}{0.6993/0.95*5*cos(0)}   \pgfmathsetmacro{\Pcy}{0.6993/0.95*5*sin(0)}
  \pgfmathsetmacro{\Pdx}{0.4388/0.95*5*cos(-45)} \pgfmathsetmacro{\Pdy}{0.4388/0.95*5*sin(-45)}
  \pgfmathsetmacro{\Pex}{0.4713/0.95*5*cos(-90)} \pgfmathsetmacro{\Pey}{0.4713/0.95*5*sin(-90)}
  \pgfmathsetmacro{\Pfx}{0.5322/0.95*5*cos(-135)}\pgfmathsetmacro{\Pfy}{0.5322/0.95*5*sin(-135)}
  \pgfmathsetmacro{\Pgx}{0.6153/0.95*5*cos(180)} \pgfmathsetmacro{\Pgy}{0.6153/0.95*5*sin(180)}
  \pgfmathsetmacro{\Phx}{0.7916/0.95*5*cos(135)} \pgfmathsetmacro{\Phy}{0.7916/0.95*5*sin(135)}
  \fill[pftpink,opacity=0.35]
    (\Pax cm,\Pay cm)--(\Pbx cm,\Pby cm)--(\Pcx cm,\Pcy cm)--(\Pdx cm,\Pdy cm)--
    (\Pex cm,\Pey cm)--(\Pfx cm,\Pfy cm)--(\Pgx cm,\Pgy cm)--(\Phx cm,\Phy cm)--cycle;
  \draw[pftpink,thick,dashed]
    (\Pax cm,\Pay cm)--(\Pbx cm,\Pby cm)--(\Pcx cm,\Pcy cm)--(\Pdx cm,\Pdy cm)--
    (\Pex cm,\Pey cm)--(\Pfx cm,\Pfy cm)--(\Pgx cm,\Pgy cm)--(\Phx cm,\Phy cm)--cycle;
  \pgfmathsetmacro{\Jax}{0.6247/0.95*5*cos(90)}  \pgfmathsetmacro{\Jay}{0.6247/0.95*5*sin(90)}
  \pgfmathsetmacro{\Jbx}{0.6879/0.95*5*cos(45)}  \pgfmathsetmacro{\Jby}{0.6879/0.95*5*sin(45)}
  \pgfmathsetmacro{\Jcx}{0.6814/0.95*5*cos(0)}   \pgfmathsetmacro{\Jcy}{0.6814/0.95*5*sin(0)}
  \pgfmathsetmacro{\Jdx}{0.7715/0.95*5*cos(-45)} \pgfmathsetmacro{\Jdy}{0.7715/0.95*5*sin(-45)}
  \pgfmathsetmacro{\Jex}{0.689/0.95*5*cos(-90)}  \pgfmathsetmacro{\Jey}{0.689/0.95*5*sin(-90)}
  \pgfmathsetmacro{\Jfx}{0.7175/0.95*5*cos(-135)}\pgfmathsetmacro{\Jfy}{0.7175/0.95*5*sin(-135)}
  \pgfmathsetmacro{\Jgx}{0.6659/0.95*5*cos(180)} \pgfmathsetmacro{\Jgy}{0.6659/0.95*5*sin(180)}
  \pgfmathsetmacro{\Jhx}{0.8104/0.95*5*cos(135)} \pgfmathsetmacro{\Jhy}{0.8104/0.95*5*sin(135)}
  \fill[jtlime,opacity=0.35]
    (\Jax cm,\Jay cm)--(\Jbx cm,\Jby cm)--(\Jcx cm,\Jcy cm)--(\Jdx cm,\Jdy cm)--
    (\Jex cm,\Jey cm)--(\Jfx cm,\Jfy cm)--(\Jgx cm,\Jgy cm)--(\Jhx cm,\Jhy cm)--cycle;
  \draw[jtlime,thick]
    (\Jax cm,\Jay cm)--(\Jbx cm,\Jby cm)--(\Jcx cm,\Jcy cm)--(\Jdx cm,\Jdy cm)--
    (\Jex cm,\Jey cm)--(\Jfx cm,\Jfy cm)--(\Jgx cm,\Jgy cm)--(\Jhx cm,\Jhy cm)--cycle;
  \node[font=\Huge\bfseries] at (0,-7cm) {10\% labeled};
\end{scope}
 
\begin{scope}[xshift=25.5cm, yshift=-8.5cm]
  \draw[jtlime,thick]         (-2cm,0)--(-.8cm,0);
  \node[font=\Huge,anchor=west] at (-.7cm,0) {JT};
  \draw[pftpink,thick,dashed] (2.5cm,0)--(3.7cm,0);
  \node[font=\Huge,anchor=west] at (3.8cm,0) {PFT};
\end{scope}
 
\end{tikzpicture}
}
\caption{Radar plots depicting image classification accuracy of the eight SSL techniques on the CIFAR-10 dataset when trained using both PFT and JT paradigms.}
\label{fig:radar_acc_cifar10}
\end{figure}

\section{Detailed Classification Analysis}

\subsection{Unimodal CIFAR$-$10}
\label{appedix:cifar10}
\begin{figure}[t]
    \centering

    \begin{subfigure}[t]{0.18\linewidth}
        \includegraphics[width=\linewidth]{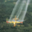}
        \caption{Original}
    \end{subfigure}
    \hfill
    \begin{subfigure}[t]{0.18\linewidth}
        \includegraphics[width=\linewidth]{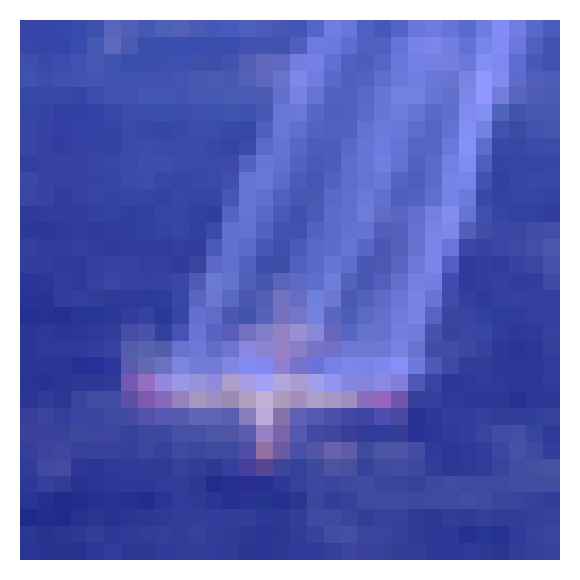}
        \caption{SimCLR PFT}
    \end{subfigure}
    \hfill
    \begin{subfigure}[t]{0.18\linewidth}
        \includegraphics[width=\linewidth]{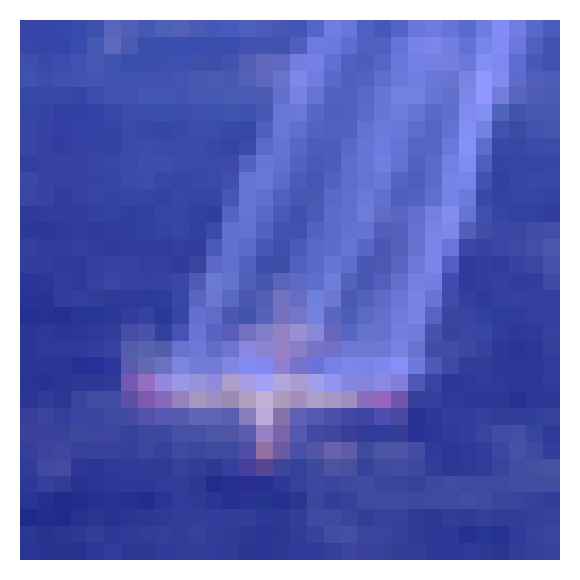}
        \caption{SimCLR JT}
    \end{subfigure}
    \hfill
    \begin{subfigure}[t]{0.18\linewidth}
        \includegraphics[width=\linewidth]{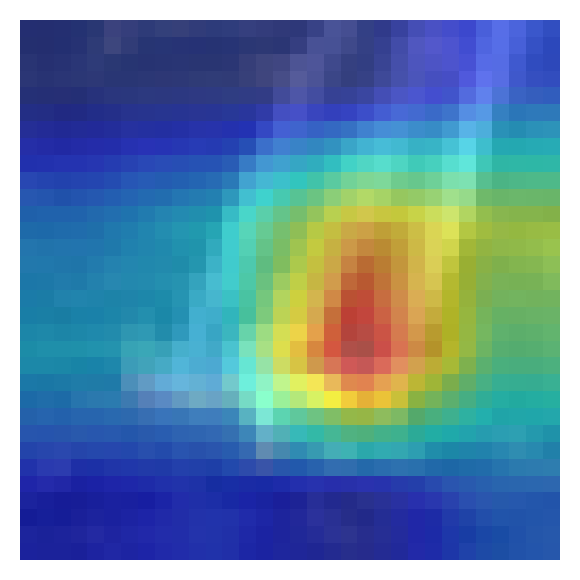}
        \caption{MoCo PFT}
    \end{subfigure}
    \hfill
    \begin{subfigure}[t]{0.18\linewidth}
        \includegraphics[width=\linewidth]{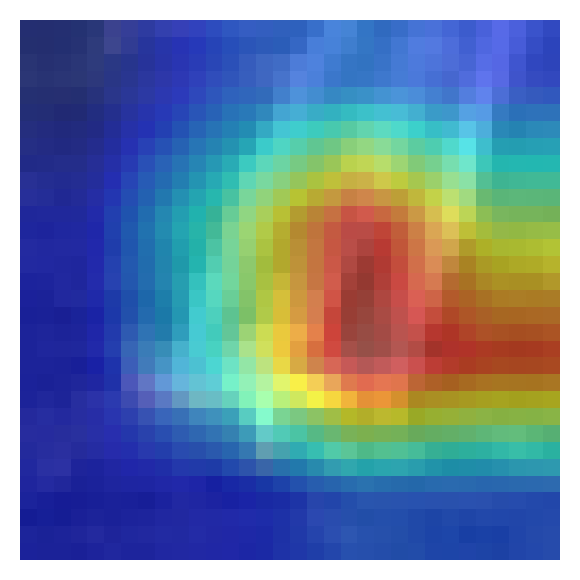}
        \caption{MoCo JT}
    \end{subfigure}

    \vspace{0.5em}

    \begin{subfigure}[t]{0.18\linewidth}
        \includegraphics[width=\linewidth]{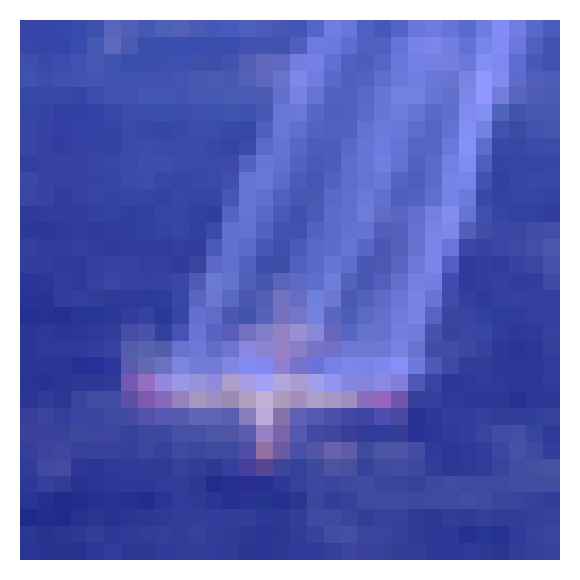}
        \caption{BYOL PFT}
    \end{subfigure}
    \hfill
    \begin{subfigure}[t]{0.18\linewidth}
        \includegraphics[width=\linewidth]{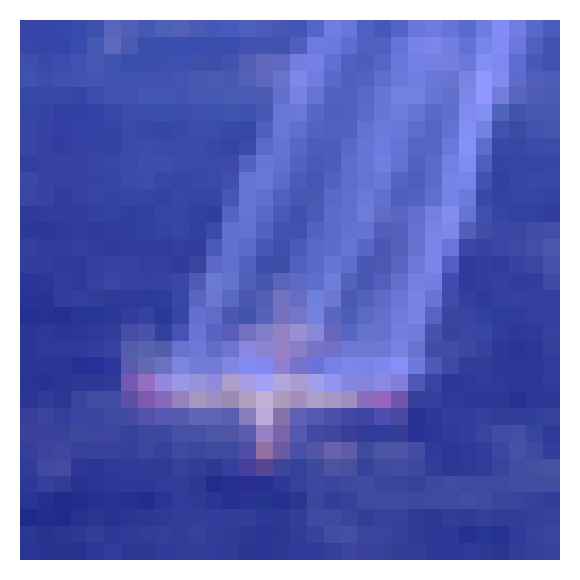}
        \caption{BYOL JT}
    \end{subfigure}
    \hfill
    \begin{subfigure}[t]{0.18\linewidth}
        \includegraphics[width=\linewidth]{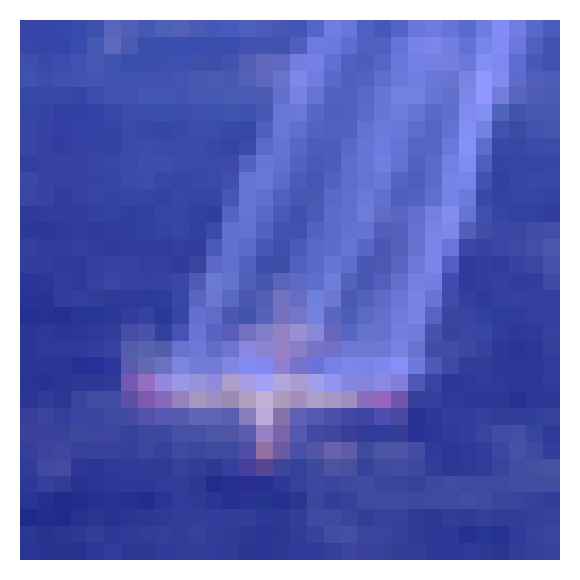}
        \caption{Barlow Twins PFT}
    \end{subfigure}
    \hfill
    \begin{subfigure}[t]{0.18\linewidth}
        \includegraphics[width=\linewidth]{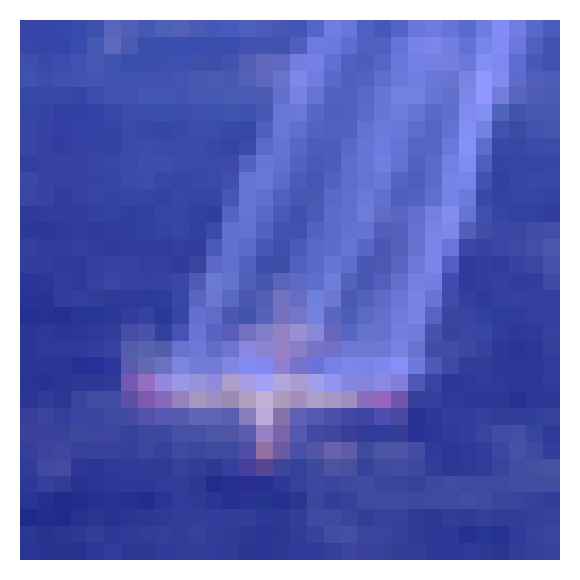}
        \caption{Barlow Twins JT}
    \end{subfigure}
    \hfill
    \begin{subfigure}[t]{0.18\linewidth}
        \includegraphics[width=\linewidth]{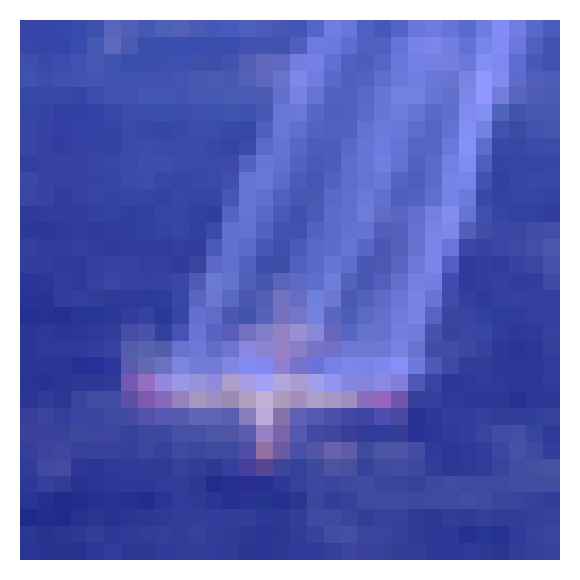}
        \caption{DINO PFT}
    \end{subfigure}

    \vspace{0.5em}

    \begin{subfigure}[t]{0.18\linewidth}
        \includegraphics[width=\linewidth]{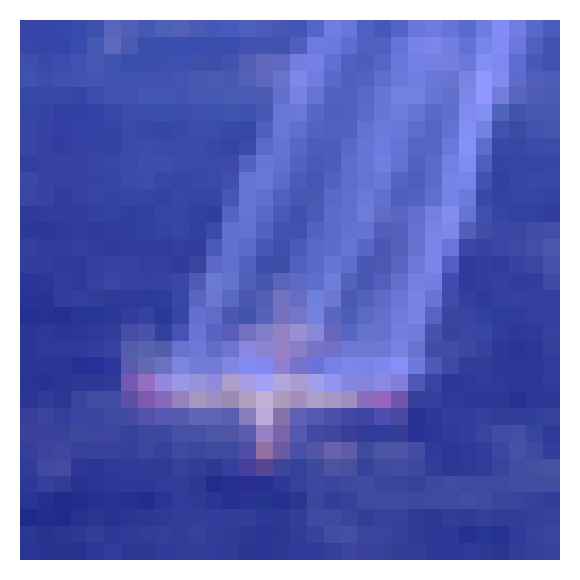}
        \caption{DINO JT}
    \end{subfigure}
    \hfill
    \begin{subfigure}[t]{0.18\linewidth}
        \includegraphics[width=\linewidth]{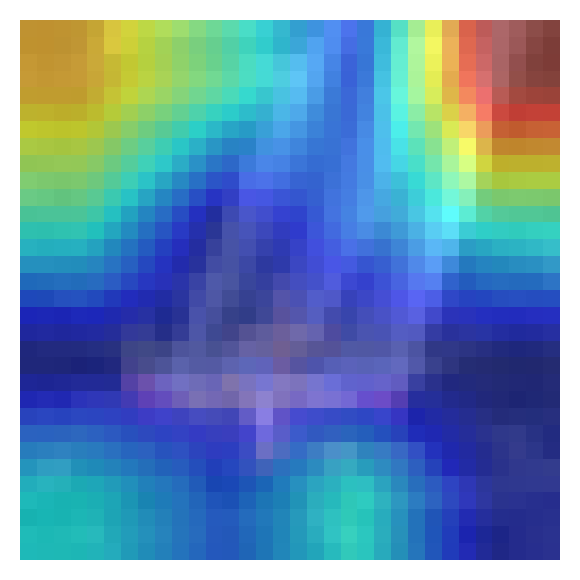}
        \caption{MAE PFT}
    \end{subfigure}
    \hfill
    \begin{subfigure}[t]{0.18\linewidth}
        \includegraphics[width=\linewidth]{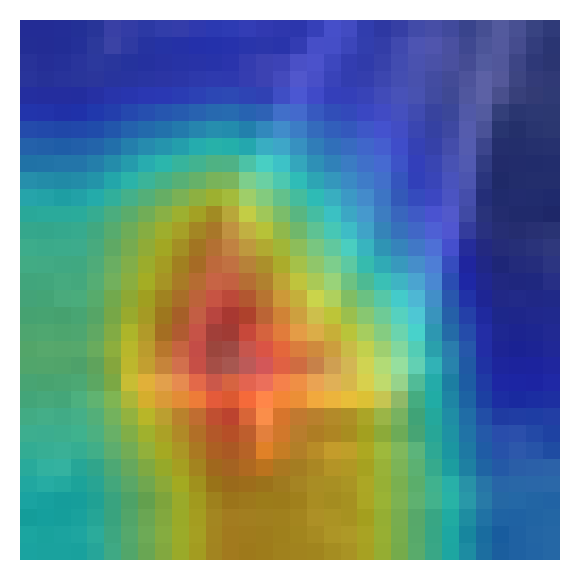}
        \caption{MAE JT}
    \end{subfigure}
    \hfill
    \begin{subfigure}[t]{0.18\linewidth}
        \includegraphics[width=\linewidth]{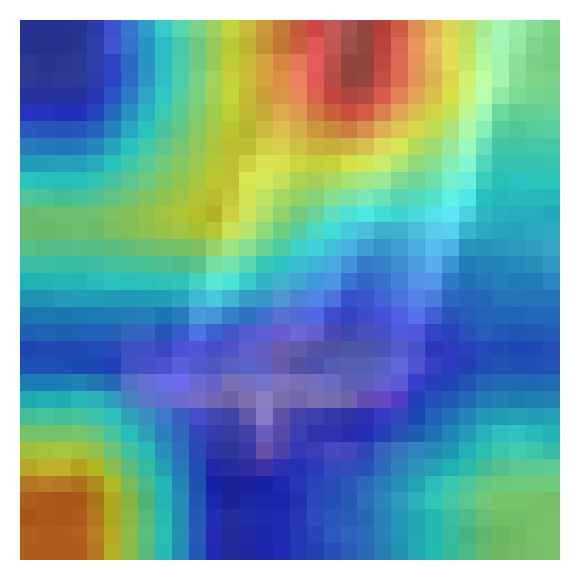}
        \caption{Rotation PFT}
    \end{subfigure}
    \hfill
    \begin{subfigure}[t]{0.18\linewidth}
        \includegraphics[width=\linewidth]{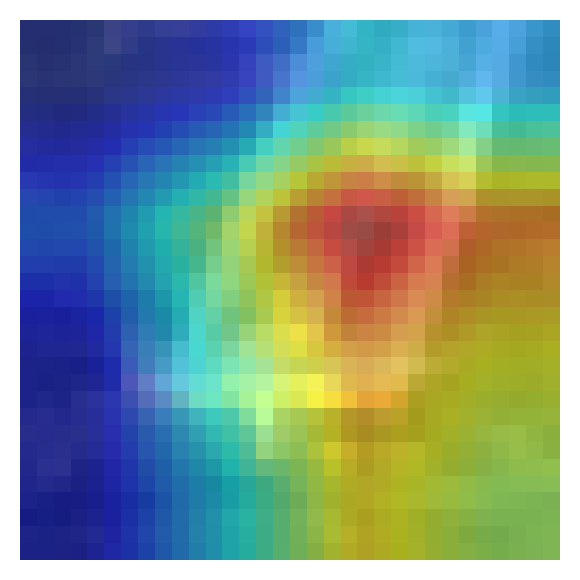}
        \caption{Rotation JT}
    \end{subfigure}

    \caption{
    Grad-CAM visualizations for different SSL methods under PFT and JT settings on CIFAR-10. Warmer regions indicate areas receiving higher attention from the model during classification.
    }
    
    \label{fig:gradcam_ssl}
\end{figure}

We present the complete CIFAR-10 experimental results under the PFT and JT training paradigms, including detailed performance across multiple label fractions and adversarial robustness evaluation. These results provide a more comprehensive comparison of the explored SSL objectives under varying supervision and noisy evaluation settings. Fig.~\ref{fig:radar_acc_cifar10} summarizes the relative classification performance trends of different SSL methods across varying label fractions under both PFT and JT settings. Table~\ref{tab:cifar10_results} shows that JT substantially improves CIFAR-10 classification performance for most SSL methods on the original test set, especially for Colorization, Rotation, MAE, and DINO. The gains are also consistent under mild adversarial noise (0.1), where JT usually preserves higher accuracy and F1 than PFT. Under stronger noise (0.5), performance drops for all methods, but JT still remains better for most frameworks, with SimCLR, BYOL, DINO, and Barlow Twins showing comparatively stronger robustness. Training time also varies across methods: JT is often faster for several methods, such as Colorization, Rotation, MoCo, and MAE, but not uniformly across all settings. Fig~\ref{fig:gradcam_ssl} further illustrates that different SSL objectives learn substantially different spatial attention behaviors under PFT and JT. Contrastive methods such as SimCLR, BYOL, and Barlow Twins focus on relatively localized object regions, whereas MoCo, MAE, and Rotation exhibit broader and more distributed activation patterns. In several cases, the transition from PFT to JT noticeably changes the spatial focus of the learned representations, particularly for MAE and Rotation. Overall, the full CIFAR$-$10 results indicate that JT improves both clean accuracy and mild-noise robustness for many SSL objectives, while the benefit under stronger perturbations is more method-dependent.

\begin{wraptable}{r}{0.51\textwidth}
\vspace{-10pt}
\centering
\caption{Performance comparison of PFT and JT using CLIP on the CIFAR10 dataset under 100\% and 10\% labeled data settings.}
\label{tab:clip_cifar10}

\scriptsize
\setlength{\tabcolsep}{4pt}

\begin{tabular}{cccc}
\toprule
Config & Labeled Data & Accuracy (\%) & F1-score (\%) \\
\midrule
PFT & 100\% & 0.9165 & 0.9165 \\
JT  & 100\% & 0.9275 & 0.9185 \\
PFT & 10\%  & 0.9145 & 0.9089 \\
JT  & 10\%  & 0.8903 & 0.8887 \\
\bottomrule
\end{tabular}

\vspace{-10pt}
\end{wraptable}

\subsection{Multimodal CIFAR$-$10}
Table~\ref{tab:clip_cifar10} presents PFT and JT results using CLIP on CIFAR$-$10. In the JT setting, the image-text contrastive loss is applied to all samples (labeled and unlabeled), while the supervised cross-entropy loss is applied exclusively to labeled samples, with the total loss defined as $\mathcal{L} = \mathcal{L}_{\text{CLIP}} + \mathcal{L}_{\text{sup}}$. In the PFT setting, the model is first pretrained with the standard CLIP contrastive objective and subsequently fine-tuned using only the supervised loss on labeled data. JT outperforms PFT at full supervision (92.75\% vs. 91.65\% accuracy), suggesting that simultaneously optimizing both objectives benefits from the full labeled set. However, at 10\% labels, JT under-performs PFT (89.03\% vs. 91.45\%), suggesting that the supervised signal becomes too sparse to complement the contrastive objective effectively, while PFT benefits from a fully converged contrastive representation before fine-tuning.

\begin{table}
    \centering
    \caption{CIFAR-10 classification performance under PFT and JT across different label fractions and adversarial noise levels. Results are reported on the original test set and under adversarial perturbations with noise strengths 0.1 and 0.5. Metrics include accuracy, macro F1-score, and training time.}
    \resizebox{\linewidth}{!}{
    \small
    \begin{tabular}{llc|ccc|cc|cc}
        \toprule
        & & & \multicolumn{3}{c|}{\textbf{Original Test}} & \multicolumn{2}{c}{\textbf{Adversarial Noise (0.1)}} & \multicolumn{2}{c}{\textbf{Adversarial Noise (0.5)}} \\
        \cmidrule(lr){4-6} \cmidrule(lr){7-8} \cmidrule(lr){9-10}
        \textbf{Framework} & \textbf{Config} & \textbf{Label (\%)} & \textbf{Accuracy} & \textbf{F1} & \textbf{Training Time (s)} & \textbf{Accuracy} & \textbf{F1} & \textbf{Accuracy} & \textbf{F1}\\
        \midrule
        \multirow{8}{*}{\textbf{Colorization}} 
        & PFT & \multirow{2}{*}{100\%} &  0.5090  & 0.5049 & 11898.0 & 0.3993 & 0.3842 & 0.2610  & 0.2132       \\
        & JT & & 0.9099 & 0.9094 & 9093.0  & 0.8782 & 0.8786 & 0.2860  & 0.2577 \\
        & PFT & \multirow{2}{*}{50\%} & 0.4864 & 0.4833 & 11547.0 & 0.3871 & 0.3689 & 0.2530  & 0.2051 \\
        & JT & & 0.8835 & 0.8831 & 6805.0  & 0.8532 & 0.8535 & 0.4373 & 0.4363 \\
        & PFT & \multirow{2}{*}{20\%} &  0.4637 & 0.4611 & 11434.0 & 0.3657 & 0.3512 & 0.2459 & 0.2019 \\
        & JT & & 0.8260  & 0.8246 & 7123.0  & 0.7961 & 0.7956 & 0.3895 & 0.389 \\
        & PFT & \multirow{2}{*}{10\%} & 0.4388 & 0.4330  & 11417.0 & 0.3452 & 0.3266 & 0.2387 & 0.1954  \\
        & JT & & 0.7715 & 0.7699 & 7790.0  & 0.7485 & 0.7479 & 0.4152 & 0.4184\\
        \midrule
        \multirow{8}{*}{\textbf{Rotation}} 
        & PFT & \multirow{2}{*}{100\%} & 0.5926 & 0.5923 & 4451.0  & 0.5031 & 0.4986 & 0.2249 & 0.1874 \\
        & JT & & 0.8387 & 0.8393 & 2241.0  & 0.7578 & 0.7617 & 0.1775 & 0.1259 \\
        & PFT & \multirow{2}{*}{50\%} & 0.5669 & 0.5651 & 4156.0  & 0.4786 & 0.4746 & 0.2222 & 0.1815 \\
        & JT & & 0.8523 & 0.8519 & 4906.0  & 0.7702 & 0.7709 & 0.2245 & 0.1912 \\
        & PFT & \multirow{2}{*}{20\%} & 0.5136 & 0.5113 & 3975.0  & 0.4454 & 0.4417 & 0.2205 & 0.1805 \\
        & JT & & 0.7816 & 0.7806 & 4085.0  & 0.7153 & 0.7138 & 0.2604 & 0.2555 \\
        & PFT & \multirow{2}{*}{10\%} & 0.4713 & 0.4645 & 3903.0  & 0.4084 & 0.3998 & 0.2190  & 0.1818 \\
        & JT & & 0.6890  & 0.6869 & 3670.0  & 0.6656 & 0.6610  & 0.3402 & 0.3067  \\
        \midrule
        \multirow{8}{*}{\textbf{SimCLR}} 
        & PFT & \multirow{2}{*}{100\%} & 0.7269 & 0.7270  & 1340.0  & 0.7200   & 0.7203 & 0.5269 & 0.5272 \\
        & JT & & 0.8250  & 0.8229 & 3096.0  & 0.8168 & 0.8140  & 0.5767 & 0.5837 \\
        & PFT & \multirow{2}{*}{50\%} & 0.7225 & 0.7210  & 1345.0  & 0.7185 & 0.7172 & 0.5176 & 0.5187 \\
        & JT & & 0.7745 & 0.7710  & 1956.0  & 0.7681 & 0.7642 & 0.5546 & 0.5446 \\
        & PFT & \multirow{2}{*}{20\%} & 0.6968 & 0.7028 & 1285.0  & 0.6939 & 0.7009 & 0.5461 & 0.5435 \\
        & JT & & 0.6905 & 0.6835 & 1502.0  & 0.6833 & 0.6751 & 0.5490  & 0.5279 \\
        & PFT & \multirow{2}{*}{10\%} & 0.6993 & 0.6991 & 1298.0  & 0.6925 & 0.6933 & 0.5189 & 0.5226 \\
        & JT & & 0.6814 & 0.673  & 3170.0  & 0.6729 & 0.6637 & 0.5060  & 0.4887 \\
        \midrule
        \multirow{8}{*}{\textbf{BYOL}} 
        & PFT & \multirow{2}{*}{100\%} & 0.6855 & 0.6812 & 5610.0  & 0.6705 & 0.6652 & 0.3964 & 0.391  \\
        & JT & & 0.8400   & 0.8370  & 5349.0  & 0.8322 & 0.8286 & 0.5479 & 0.5526 \\
        & PFT & \multirow{2}{*}{50\%} &  0.6817 & 0.6767 & 5567.0  & 0.6691 & 0.6633 & 0.4073 & 0.4014 \\
        & JT & & 0.7980  & 0.7939 & 5614.0  & 0.7918 & 0.7873 & 0.5363 & 0.5273 \\
        & PFT & \multirow{2}{*}{20\%} & 0.6713 & 0.6677 & 5219.0  & 0.6571 & 0.6526 & 0.3818 & 0.3711 \\
        & JT & & 0.7390  & 0.7323 & 6622.0  & 0.7344 & 0.7272 & 0.5638 & 0.544  \\
        & PFT & \multirow{2}{*}{10\%}  & 0.6586 & 0.6545 & 5239.0  & 0.6438 & 0.6398 & 0.3769 & 0.3605 \\
        & JT & &  0.6879 & 0.6801 & 8470.0  & 0.6844 & 0.6762 & 0.5272 & 0.5016 \\
        \midrule
        \multirow{8}{*}{\textbf{MoCo}} 
        & PFT & \multirow{2}{*}{100\%} & 0.8205 & 0.8179 & 14328.0 & 0.7791 & 0.7770  & 0.2158 & 0.1870  \\
        & JT & & 0.8938 & 0.8938 & 3386.0  & 0.8709 & 0.8711 & 0.3070  & 0.2818 \\
        & PFT & \multirow{2}{*}{50\%} & 0.8156 & 0.8132 & 14199.0 & 0.7653 & 0.764  & 0.2122 & 0.1776 \\
        & JT & & 0.9009 & 0.9000    & 5835.0  & 0.8777 & 0.877  & 0.2941 & 0.2516 \\
        & PFT & \multirow{2}{*}{20\%} & 0.8019 & 0.7998 & 14215.0 & 0.7554 & 0.7536 & 0.1873 & 0.1513 \\
        & JT & & 0.8594 & 0.8586 & 5515.0  & 0.8347 & 0.8335 & 0.3047 & 0.2953 \\
        & PFT & \multirow{2}{*}{10\%} & 0.7916 & 0.7893 & 14303.0 & 0.7384 & 0.7365 & 0.1869 & 0.1473 \\
        & JT & & 0.8104 & 0.8078 & 7931.0  & 0.7842 & 0.7818 & 0.2724 & 0.2577 \\
        \midrule
        \multirow{8}{*}{\textbf{MAE}} 
        & PFT & \multirow{2}{*}{100\%} & 0.5920  & 0.5929 & 8871.0  & 0.5723 & 0.5706 & 0.2285 & 0.1556 \\
        & JT & & 0.8341 & 0.8327 & 1457.0  & 0.7860  & 0.7849 & 0.2711 & 0.2187 \\
        & PFT & \multirow{2}{*}{50\%} & 0.5797 & 0.5809 & 8221.0  & 0.5594 & 0.5574 & 0.2306 & 0.1562 \\
        & JT & & 0.8179 & 0.8159 & 1492.0  & 0.7766 & 0.7752 & 0.3034 & 0.2534 \\
        & PFT & \multirow{2}{*}{20\%} & 0.5520  & 0.5495 & 7707.0  & 0.5228 & 0.5148 & 0.2172 & 0.1225 \\
        & JT & & 0.7870  & 0.7876 & 3804.0  & 0.7398 & 0.7409 & 0.2838 & 0.2422 \\
        & PFT & \multirow{2}{*}{10\%} & 0.5322 & 0.5268 & 7520.0  & 0.4991 & 0.4866 & 0.2164 & 0.1157 \\
        & JT & & 0.7175 & 0.7098 & 7174.0  & 0.6880  & 0.6789 & 0.3583 & 0.3399 \\
        \midrule
        \multirow{8}{*}{\textbf{DINO}} 
        & PFT & \multirow{2}{*}{100\%} & 0.6541 & 0.6467 & 8283.0  & 0.6316 & 0.6248 & 0.2502 & 0.2391 \\
        & JT & & 0.8079 & 0.8050  & 15287.0 & 0.7985 & 0.7952 & 0.5378 & 0.5313 \\
        & PFT & \multirow{2}{*}{50\%} & 0.6382 & 0.6284 & 8027.0  & 0.6232 & 0.6141 & 0.2584 & 0.2532 \\
        & JT & & 0.7720  & 0.7679 & 7666.0  & 0.7634 & 0.7586 & 0.5132 & 0.5063 \\
        & PFT & \multirow{2}{*}{20\%} & 0.6320  & 0.6240  & 8043.0  & 0.6130  & 0.6063 & 0.2486 & 0.2357 \\
        & JT & & 0.7201 & 0.7138 & 8533.0  & 0.7127 & 0.7058 & 0.5224 & 0.5028 \\
        & PFT & \multirow{2}{*}{10\%} & 0.6153 & 0.6081 & 8146.0  & 0.5967 & 0.5907 & 0.2375 & 0.2209 \\
        & JT & & 0.6659 & 0.6585 & 14531.0 & 0.6590  & 0.6511 & 0.4895 & 0.4646 \\
        \midrule
        \multirow{8}{*}{\textbf{Barlow Twins}} 
        & PFT & \multirow{2}{*}{100\%} & 0.5163 & 0.4642 & 8630.0  & 0.5144 & 0.462  & 0.4664 & 0.4062 \\
        & JT & & 0.6883 & 0.6813 & 5572.0  & 0.6781 & 0.671  & 0.5249 & 0.5103 \\
        & PFT & \multirow{2}{*}{50\%} & 0.5537 & 0.5498 & 9050.0  & 0.549  & 0.5453 & 0.4843 & 0.4712 \\
        & JT & & 0.6620  & 0.6532 & 5411.0  & 0.6558 & 0.6465 & 0.5037 & 0.4955 \\
        & PFT & \multirow{2}{*}{20\%} & 0.5242 & 0.5271 & 8737.0  & 0.522  & 0.5261 & 0.4807 & 0.4791 \\
        & JT & & 0.6373 & 0.6259 & 6381.0  & 0.6256 & 0.6126 & 0.4754 & 0.4530  \\
        & PFT & \multirow{2}{*}{10\%} & 0.5124 & 0.4621 & 8555.0  & 0.5015 & 0.4525 & 0.4260  & 0.3728 \\
        & JT & & 0.6247 & 0.6133 & 8712.0  & 0.6156 & 0.6033 & 0.4510  & 0.4297 \\
        \bottomrule
    \end{tabular}
    }
    \label{tab:cifar10_results}
\end{table}

\clearpage

\begin{table}
\vspace{-10pt}
\centering
\small
\caption{Classification performance of different SSL frameworks on CrisisMMD using 128$\times$128 image inputs under the PFT and JT training paradigms with 100\% labeled data.}
\label{tab:crisismmd_128}

\setlength{\tabcolsep}{5pt}

\begin{tabular}{llcc}
\toprule
Framework & Config & Acc & F1 \\
\midrule

SimCLR & PFT & 0.7616 & 0.7536 \\
        & JT  & 0.7342 & 0.7402 \\
\midrule

BYOL   & PFT & 0.7783 & 0.7699 \\
        & JT  & 0.7320 & 0.7408 \\

\midrule

MoCo   & PFT & 0.7807 & 0.7512 \\
        & JT  & 0.7618 & 0.7384 \\
\midrule

MAE    & PFT & 0.6817 & 0.6695 \\
        & JT  & 0.7407 & 0.7181 \\

\midrule

DINO    & PFT & 0.7729 & 0.7726 \\
        & JT  & 0.7726 & 0.7561 \\

\midrule

Barlow Twins    & PFT & 0.7306 & 0.7127 \\
        & JT  & 0.6062 & 0.5986 \\
\bottomrule
\end{tabular}
\vspace{-10pt}
\end{table}

\subsection{CrisisMMD Performance}
\label{appendix:crisis}
Table~\ref{tab:CrisisMMD_DMD_res} presents the cross-dataset generalization results where models are trained on CrisisMMD and evaluated on both the in-domain CrisisMMD test set and the out-of-domain DMD dataset. Overall, contrastive SSL methods such as SimCLR, BYOL, MoCo, and DINO achieve the strongest transfer performance across both datasets, with several methods maintaining competitive accuracy and F1 scores under severe label limitations. SimCLR exhibits particularly strong cross-domain robustness, achieving the best DMD performance at lower label fractions, while DINO and MoCo remain consistently stable across both in-domain and out-of-domain evaluation settings. In contrast, reconstruction-oriented methods such as Colorization and MAE generally show larger performance drops under domain shift, particularly at lower label fractions.

The comparison between PFT and JT further reveals method-dependent behavior. For Colorization, Rotation, and MAE, JT frequently improves in-domain CrisisMMD performance and substantially reduces training time, especially under limited-label settings. However, for several contrastive methods, including SimCLR, BYOL, MoCo, and DINO, PFT often provides stronger cross-dataset transfer performance on DMD, particularly at 10\% and 20\% label fractions. These trends suggest that while JT can improve data efficiency and optimization stability, PFT may preserve more transferable representations for certain contrastive SSL objectives under distribution shift. Across nearly all methods, JT consistently reduces training time compared with PFT while maintaining comparable or competitive downstream performance.

We additionally evaluate CrisisMMD classification using lower-resolution 128$\times$128 image inputs in Table~\ref{tab:crisismmd_128}. The results show that contrastive methods, particularly SimCLR, BYOL, MoCo, and DINO, remain relatively robust under reduced image resolution, while MAE benefits more noticeably from JT. In contrast, Barlow Twins experiences a substantial performance drop under JT, indicating increased sensitivity to reduced-resolution representations.

\begin{table}
    \centering
    \small
    \caption{Cross-dataset generalization results from CrisisMMD to DMD under different SSL training paradigms and label fractions. Models are trained on CrisisMMD and evaluated on both the in-domain CrisisMMD test set and the out-of-domain DMD dataset. Results report accuracy, macro F1-score, and training time.}
    \resizebox{0.5\textheight}{!}{
    \small
    \begin{tabular}{llc|ccc|cc}
        \toprule
        & & & \multicolumn{3}{c|}{\textbf{CrisisMMD}} & \multicolumn{2}{c}{\textbf{DMD}}\\
        \cmidrule(lr){4-6} \cmidrule(lr){7-8} 
        \textbf{Framework} & \textbf{Config} & \textbf{Label (\%)} & \textbf{Accuracy} & \textbf{F1} & \textbf{Training Time (h)} & \textbf{Accuracy} & \textbf{F1} \\
        \midrule
        \multirow{8}{*}{Colorization} 
        & PFT    & \multirow{2}{*}{100} & 0.663  & 0.6699 & 2.15    & 0.5409 & 0.5062 \\
        & JT     &                      & 0.7287 & 0.7263 & 1.03  & 0.5463 & 0.5104 \\
        & PFT    & \multirow{2}{*}{50}  & 0.6665 & 0.6717 & 1.68  & 0.5587 & 0.5096 \\
                              & JT     & & 0.7471 & 0.7310 & 0.70     & 0.5722 & 0.5188 \\
         & PFT    & \multirow{2}{*}{20}  & 0.5994 & 0.5998 & 1.51  & 0.4573 & 0.4472 \\
         & JT     & & 0.7244 & 0.7073 & 1.18  & 0.5428 & 0.4930  \\
         & PFT    & \multirow{2}{*}{10}  & 0.6081 & 0.6095 & 1.55    & 0.4904 & 0.4762 \\
         & JT     & & 0.6735 & 0.6594 & 1.00     & 0.4733 & 0.4446 \\
         \midrule
\multirow{8}{*}{Rotation}     & PFT    & \multirow{2}{*}{100} & 0.7167 & 0.7046 & 7.88  & 0.5683 & 0.5234 \\
         & JT     & & 0.7512 & 0.7419 & 1.03  & 0.5768 & 0.5313 \\
         & PFT    & \multirow{2}{*}{50}  & 0.655  & 0.6307 & 6.38  & 0.5086 & 0.4682 \\
         & JT     & & 0.7225 & 0.7007 & 0.51  & 0.5798 & 0.5361 \\
         & PFT    & \multirow{2}{*}{20}  & 0.6441 & 0.6308 & 6.25  & 0.4985 & 0.4659 \\
         & JT     & & 0.7115 & 0.6717 & 0.80  & 0.5639 & 0.5122 \\
         & PFT    & \multirow{2}{*}{10}  & 0.6281 & 0.6038 & 6.20  & 0.4914 & 0.4525 \\
         & JT     & & 0.6717 & 0.6595 & 1.48  & 0.5070 & 0.4900   \\
         \midrule
\multirow{8}{*}{SimCLR}       & PFT    & \multirow{2}{*}{100} & 0.7941 & 0.7843 & 2.76  & 0.7742 & 0.7729 \\
         & JT     & & 0.7408 & 0.726  & 0.51  & 0.7697 & 0.7695 \\
         & PFT    & \multirow{2}{*}{50}  & 0.7848 & 0.7747 & 2.91  & 0.7784 & 0.7771 \\
         & JT     & & 0.7441 & 0.7406 & 0.41  & 0.6946 & 0.6939 \\
         & PFT    & \multirow{2}{*}{20}  & 0.7896 & 0.7813 & 2.71  & 0.7898 & 0.7893 \\
         & JT     & & 0.7152 & 0.7094 & 0.41  & 0.7137 & 0.7137 \\
         & PFT    & \multirow{2}{*}{10}  & 0.7830 & 0.7795 & 2.68  & 0.8037 & 0.8038 \\
         & JT     & & 0.6915 & 0.6787 & 0.56  & 0.6771 & 0.6768 \\
         \midrule
\multirow{8}{*}{BYOL}         & PFT    & \multirow{2}{*}{100} & 0.7762 & 0.7712 & 2.28  & 0.7909 & 0.7909 \\
         & JT     & & 0.6972 & 0.6884 & 0.43  & 0.6561 & 0.6539 \\
         & PFT    & \multirow{2}{*}{50}  & 0.7673 & 0.7621 & 2.20   & 0.7946 & 0.7946 \\
         & JT     & & 0.6811 & 0.6839 & 0.78  & 0.6721 & 0.6715 \\
         & PFT    & \multirow{2}{*}{20}  & 0.7674 & 0.7504 & 2.36  & 0.7511 & 0.7486 \\
         & JT     & & 0.6906 & 0.6919 & 1.18  & 0.6880  & 0.6877 \\
         & PFT    & \multirow{2}{*}{10}  & 0.7698 & 0.7636 & 2.20    & 0.7997 & 0.7997 \\
         & JT     & & 0.6535 & 0.6566 & 0.40    & 0.5533 & 0.5330  \\
         \midrule
\multirow{8}{*}{MoCo}         & PFT    & \multirow{2}{*}{100} & 0.784  & 0.7592 & 11.40   & 0.7633 & 0.7601 \\
         & JT     & & 0.7998 & 0.7840  & 7.91  & 0.7631 & 0.7615 \\
         & PFT    & \multirow{2}{*}{50}  & 0.7725 & 0.7721 & 11.07 & 0.7889 & 0.7889 \\
         & JT     & & 0.7628 & 0.7597 & 5.11  & 0.7470  & 0.7470  \\
         & PFT    & \multirow{2}{*}{20}  & 0.7767 & 0.7714 & 10.98 & 0.7685 & 0.7685 \\
         & JT     & & 0.7505 & 0.7380  & 5.08  & 0.7459 & 0.7458 \\
         & PFT    & \multirow{2}{*}{10}  & 0.7698 & 0.7636 & 10.98 & 0.7759 & 0.7759 \\
         & JT     & & 0.6356 & 0.6350  & 2.31  & 0.5393 & 0.4921 \\
         \midrule
\multirow{8}{*}{MAE}          & PFT    & \multirow{2}{*}{100} & 0.7341 & 0.7042 & 4.85    & 0.7051 & 0.6985 \\
         & JT     & & 0.6859 & 0.6904 & 1.15    & 0.7019 & 0.7011 \\
         & PFT    & \multirow{2}{*}{50}  & 0.7398 & 0.7252 & 4.48  & 0.7315 & 0.7304 \\
         & JT     & & 0.6952 & 0.6683 & 1.03  & 0.684  & 0.6839 \\
         & PFT    & \multirow{2}{*}{20}  & 0.7244 & 0.712  & 4.40     & 0.7391 & 0.7389 \\
         & JT     & & 0.6901 & 0.6766 & 1.21  & 0.6727 & 0.6707 \\
         & PFT    & \multirow{2}{*}{10}  & 0.7348 & 0.7219 & 4.45    & 0.7552 & 0.755  \\
         & JT     & & 0.6705 & 0.6645 & 1.83  & 0.5804 & 0.5614 \\
         \midrule
\multirow{8}{*}{DINO}         & PFT    & \multirow{2}{*}{100} & 0.7791 & 0.7749 & 7.68  & 0.7847 & 0.7845 \\
         & JT     & & 0.7722 & 0.7671 & 1.20     & 0.7475 & 0.7471 \\
         & PFT    & \multirow{2}{*}{50}  & 0.7714 & 0.7419 & 7.43  & 0.7714 & 0.7707 \\
         & JT     & & 0.7421 & 0.7556 & 1.00    & 0.7588 & 0.7523 \\
         & PFT    & \multirow{2}{*}{20}  & 0.7855 & 0.7710  & 7.23  & 0.7853 & 0.7853 \\
         & JT     & & 0.7271 & 0.7276 & 0.98  & 0.739  & 0.7381 \\
         & PFT    & \multirow{2}{*}{10}  & 0.7662 & 0.7632 & 7.06  & 0.7757 & 0.7749 \\
         & JT     & & 0.7094 & 0.7229 & 0.88  & 0.7437 & 0.7417 \\
         \midrule
\multirow{8}{*}{Barlow Twins} & PFT    & \multirow{2}{*}{100} & 0.7169 & 0.6885 & 2.16  & 0.6924 & 0.6845 \\
         & JT     & & 0.6258 & 0.6130  & 0.53  & 0.5278 & 0.4661 \\
         & PFT    & \multirow{2}{*}{50}  & 0.7288 & 0.7174 & 2.08  & 0.7283 & 0.7281 \\
         & JT     & & 0.6357 & 0.6284 & 1.33  & 0.5423 & 0.5100   \\
         & PFT    & \multirow{2}{*}{20}  & 0.7127 & 0.6989 & 1.96  & 0.7379 & 0.7377 \\
         & JT     & & 0.6241 & 0.6204 & 0.41  & 0.5317 & 0.4992 \\
         & PFT    & \multirow{2}{*}{10}  & 0.7254 & 0.7132 & 1.95    & 0.7614 & 0.7614 \\
         & JT     & & 0.5780  & 0.5781 & 0.71  & 0.5614 & 0.5500   \\

        \bottomrule
    \end{tabular}
    }
    \label{tab:CrisisMMD_DMD_res}
\end{table}

\clearpage

\subsection{Chest X-Ray Classification}
\label{appendix:jsrt_cls}

\begin{figure*}[h]
\resizebox{\textwidth}{!}{
\begin{tikzpicture}
 
\definecolor{pftpink}{RGB}{255,20,147}
\definecolor{jtlime}{RGB}{50,200,50}
\definecolor{gridc}{RGB}{200,200,200}
 
\newcommand{\drawpanel}[1]{%
  \foreach \v/\lab in {0.20/0.20, 0.30/0.30, 0.40/0.40, 0.50/0.50, 0.60/0.60}{
    \pgfmathsetmacro{\rr}{\v/0.70*5}
    \draw[gridc,thin]
      (90:\rr cm)--(45:\rr cm)--(0:\rr cm)--(-45:\rr cm)--
      (-90:\rr cm)--(-135:\rr cm)--(180:\rr cm)--(135:\rr cm)--cycle;
  }
  \foreach \i in {0,...,7}{
    \pgfmathsetmacro{\ang}{90-\i*45}
    \draw[gridc,thin] (0,0)--(\ang:5cm);
  }
  \node[font=\Huge,above]       at (90:5.5cm)   {Barlow Twins};
  \node[font=\Huge,above right] at (45:5.5cm)   {BYOL};
  \node[font=\Huge,right]       at (0:5.5cm)    {SimCLR};
  \node[font=\Huge,below right] at (-45:5.5cm)  {Colorization};
  \node[font=\Huge,below]       at (-90:5.5cm)  {Rotation};
  \node[font=\Huge,below left]  at (-135:5.5cm) {MAE};
  \node[font=\Huge,left]        at (180:5.5cm)  {DINO};
  \node[font=\Huge,above left]  at (135:5.5cm)  {MoCo};
}

\begin{scope}[xshift=0cm]
  \drawpanel{}
  \pgfmathsetmacro{\Pax}{0.4167/0.7*5*cos(90)}  \pgfmathsetmacro{\Pay}{0.4167/0.7*5*sin(90)}
  \pgfmathsetmacro{\Pbx}{0.4375/0.7*5*cos(45)}  \pgfmathsetmacro{\Pby}{0.4375/0.7*5*sin(45)}
  \pgfmathsetmacro{\Pcx}{0.6042/0.7*5*cos(0)}   \pgfmathsetmacro{\Pcy}{0.6042/0.7*5*sin(0)}
  \pgfmathsetmacro{\Pdx}{0.4792/0.7*5*cos(-45)} \pgfmathsetmacro{\Pdy}{0.4792/0.7*5*sin(-45)}
  \pgfmathsetmacro{\Pex}{0.4583/0.7*5*cos(-90)} \pgfmathsetmacro{\Pey}{0.4583/0.7*5*sin(-90)}
  \pgfmathsetmacro{\Pfx}{0.3958/0.7*5*cos(-135)}\pgfmathsetmacro{\Pfy}{0.3958/0.7*5*sin(-135)}
  \pgfmathsetmacro{\Pgx}{0.4375/0.7*5*cos(180)} \pgfmathsetmacro{\Pgy}{0.4375/0.7*5*sin(180)}
  \pgfmathsetmacro{\Phx}{0.2917/0.7*5*cos(135)} \pgfmathsetmacro{\Phy}{0.2917/0.7*5*sin(135)}
  \fill[pftpink,opacity=0.35]
    (\Pax cm,\Pay cm)--(\Pbx cm,\Pby cm)--(\Pcx cm,\Pcy cm)--(\Pdx cm,\Pdy cm)--
    (\Pex cm,\Pey cm)--(\Pfx cm,\Pfy cm)--(\Pgx cm,\Pgy cm)--(\Phx cm,\Phy cm)--cycle;
  \draw[pftpink,thick,dashed]
    (\Pax cm,\Pay cm)--(\Pbx cm,\Pby cm)--(\Pcx cm,\Pcy cm)--(\Pdx cm,\Pdy cm)--
    (\Pex cm,\Pey cm)--(\Pfx cm,\Pfy cm)--(\Pgx cm,\Pgy cm)--(\Phx cm,\Phy cm)--cycle;
  \pgfmathsetmacro{\Jax}{0.525/0.7*5*cos(90)}   \pgfmathsetmacro{\Jay}{0.525/0.7*5*sin(90)}
  \pgfmathsetmacro{\Jbx}{0.425/0.7*5*cos(45)}   \pgfmathsetmacro{\Jby}{0.425/0.7*5*sin(45)}
  \pgfmathsetmacro{\Jcx}{0.6/0.7*5*cos(0)}      \pgfmathsetmacro{\Jcy}{0.6/0.7*5*sin(0)}
  \pgfmathsetmacro{\Jdx}{0.5/0.7*5*cos(-45)}    \pgfmathsetmacro{\Jdy}{0.5/0.7*5*sin(-45)}
  \pgfmathsetmacro{\Jex}{0.4792/0.7*5*cos(-90)} \pgfmathsetmacro{\Jey}{0.4792/0.7*5*sin(-90)}
  \pgfmathsetmacro{\Jfx}{0.5/0.7*5*cos(-135)}   \pgfmathsetmacro{\Jfy}{0.5/0.7*5*sin(-135)}
  \pgfmathsetmacro{\Jgx}{0.4792/0.7*5*cos(180)} \pgfmathsetmacro{\Jgy}{0.4792/0.7*5*sin(180)}
  \pgfmathsetmacro{\Jhx}{0.35/0.7*5*cos(135)}   \pgfmathsetmacro{\Jhy}{0.35/0.7*5*sin(135)}
  \fill[jtlime,opacity=0.35]
    (\Jax cm,\Jay cm)--(\Jbx cm,\Jby cm)--(\Jcx cm,\Jcy cm)--(\Jdx cm,\Jdy cm)--
    (\Jex cm,\Jey cm)--(\Jfx cm,\Jfy cm)--(\Jgx cm,\Jgy cm)--(\Jhx cm,\Jhy cm)--cycle;
  \draw[jtlime,thick]
    (\Jax cm,\Jay cm)--(\Jbx cm,\Jby cm)--(\Jcx cm,\Jcy cm)--(\Jdx cm,\Jdy cm)--
    (\Jex cm,\Jey cm)--(\Jfx cm,\Jfy cm)--(\Jgx cm,\Jgy cm)--(\Jhx cm,\Jhy cm)--cycle;
  \node[font=\Huge\bfseries] at (0,-7cm) {100\% labeled};
\end{scope}

\begin{scope}[xshift=17cm]
  \drawpanel{}
  \pgfmathsetmacro{\Pax}{0.4792/0.7*5*cos(90)}  \pgfmathsetmacro{\Pay}{0.4792/0.7*5*sin(90)}
  \pgfmathsetmacro{\Pbx}{0.2917/0.7*5*cos(45)}  \pgfmathsetmacro{\Pby}{0.2917/0.7*5*sin(45)}
  \pgfmathsetmacro{\Pcx}{0.2917/0.7*5*cos(0)}   \pgfmathsetmacro{\Pcy}{0.2917/0.7*5*sin(0)}
  \pgfmathsetmacro{\Pdx}{0.4792/0.7*5*cos(-45)} \pgfmathsetmacro{\Pdy}{0.4792/0.7*5*sin(-45)}
  \pgfmathsetmacro{\Pex}{0.4375/0.7*5*cos(-90)} \pgfmathsetmacro{\Pey}{0.4375/0.7*5*sin(-90)}
  \pgfmathsetmacro{\Pfx}{0.3958/0.7*5*cos(-135)}\pgfmathsetmacro{\Pfy}{0.3958/0.7*5*sin(-135)}
  \pgfmathsetmacro{\Pgx}{0.3125/0.7*5*cos(180)} \pgfmathsetmacro{\Pgy}{0.3125/0.7*5*sin(180)}
  \pgfmathsetmacro{\Phx}{0.4375/0.7*5*cos(135)} \pgfmathsetmacro{\Phy}{0.4375/0.7*5*sin(135)}
  \fill[pftpink,opacity=0.35]
    (\Pax cm,\Pay cm)--(\Pbx cm,\Pby cm)--(\Pcx cm,\Pcy cm)--(\Pdx cm,\Pdy cm)--
    (\Pex cm,\Pey cm)--(\Pfx cm,\Pfy cm)--(\Pgx cm,\Pgy cm)--(\Phx cm,\Phy cm)--cycle;
  \draw[pftpink,thick,dashed]
    (\Pax cm,\Pay cm)--(\Pbx cm,\Pby cm)--(\Pcx cm,\Pcy cm)--(\Pdx cm,\Pdy cm)--
    (\Pex cm,\Pey cm)--(\Pfx cm,\Pfy cm)--(\Pgx cm,\Pgy cm)--(\Phx cm,\Phy cm)--cycle;
  \pgfmathsetmacro{\Jax}{0.35/0.7*5*cos(90)}    \pgfmathsetmacro{\Jay}{0.35/0.7*5*sin(90)}
  \pgfmathsetmacro{\Jbx}{0.575/0.7*5*cos(45)}   \pgfmathsetmacro{\Jby}{0.575/0.7*5*sin(45)}
  \pgfmathsetmacro{\Jcx}{0.35/0.7*5*cos(0)}     \pgfmathsetmacro{\Jcy}{0.35/0.7*5*sin(0)}
  \pgfmathsetmacro{\Jdx}{0.5625/0.7*5*cos(-45)} \pgfmathsetmacro{\Jdy}{0.5625/0.7*5*sin(-45)}
  \pgfmathsetmacro{\Jex}{0.4792/0.7*5*cos(-90)} \pgfmathsetmacro{\Jey}{0.4792/0.7*5*sin(-90)}
  \pgfmathsetmacro{\Jfx}{0.4792/0.7*5*cos(-135)}\pgfmathsetmacro{\Jfy}{0.4792/0.7*5*sin(-135)}
  \pgfmathsetmacro{\Jgx}{0.4583/0.7*5*cos(180)} \pgfmathsetmacro{\Jgy}{0.4583/0.7*5*sin(180)}
  \pgfmathsetmacro{\Jhx}{0.25/0.7*5*cos(135)}   \pgfmathsetmacro{\Jhy}{0.25/0.7*5*sin(135)}
  \fill[jtlime,opacity=0.35]
    (\Jax cm,\Jay cm)--(\Jbx cm,\Jby cm)--(\Jcx cm,\Jcy cm)--(\Jdx cm,\Jdy cm)--
    (\Jex cm,\Jey cm)--(\Jfx cm,\Jfy cm)--(\Jgx cm,\Jgy cm)--(\Jhx cm,\Jhy cm)--cycle;
  \draw[jtlime,thick]
    (\Jax cm,\Jay cm)--(\Jbx cm,\Jby cm)--(\Jcx cm,\Jcy cm)--(\Jdx cm,\Jdy cm)--
    (\Jex cm,\Jey cm)--(\Jfx cm,\Jfy cm)--(\Jgx cm,\Jgy cm)--(\Jhx cm,\Jhy cm)--cycle;
  \node[font=\Huge\bfseries] at (0,-7cm) {50\% labeled};
\end{scope}

\begin{scope}[xshift=34cm]
  \drawpanel{}
  \pgfmathsetmacro{\Pax}{0.5/0.7*5*cos(90)}     \pgfmathsetmacro{\Pay}{0.5/0.7*5*sin(90)}
  \pgfmathsetmacro{\Pbx}{0.4583/0.7*5*cos(45)}  \pgfmathsetmacro{\Pby}{0.4583/0.7*5*sin(45)}
  \pgfmathsetmacro{\Pcx}{0.4583/0.7*5*cos(0)}   \pgfmathsetmacro{\Pcy}{0.4583/0.7*5*sin(0)}
  \pgfmathsetmacro{\Pdx}{0.4583/0.7*5*cos(-45)} \pgfmathsetmacro{\Pdy}{0.4583/0.7*5*sin(-45)}
  \pgfmathsetmacro{\Pex}{0.4375/0.7*5*cos(-90)} \pgfmathsetmacro{\Pey}{0.4375/0.7*5*sin(-90)}
  \pgfmathsetmacro{\Pfx}{0.4167/0.7*5*cos(-135)}\pgfmathsetmacro{\Pfy}{0.4167/0.7*5*sin(-135)}
  \pgfmathsetmacro{\Pgx}{0.4375/0.7*5*cos(180)} \pgfmathsetmacro{\Pgy}{0.4375/0.7*5*sin(180)}
  \pgfmathsetmacro{\Phx}{0.3125/0.7*5*cos(135)} \pgfmathsetmacro{\Phy}{0.3125/0.7*5*sin(135)}
  \fill[pftpink,opacity=0.35]
    (\Pax cm,\Pay cm)--(\Pbx cm,\Pby cm)--(\Pcx cm,\Pcy cm)--(\Pdx cm,\Pdy cm)--
    (\Pex cm,\Pey cm)--(\Pfx cm,\Pfy cm)--(\Pgx cm,\Pgy cm)--(\Phx cm,\Phy cm)--cycle;
  \draw[pftpink,thick,dashed]
    (\Pax cm,\Pay cm)--(\Pbx cm,\Pby cm)--(\Pcx cm,\Pcy cm)--(\Pdx cm,\Pdy cm)--
    (\Pex cm,\Pey cm)--(\Pfx cm,\Pfy cm)--(\Pgx cm,\Pgy cm)--(\Phx cm,\Phy cm)--cycle;
  \pgfmathsetmacro{\Jax}{0.45/0.7*5*cos(90)}    \pgfmathsetmacro{\Jay}{0.45/0.7*5*sin(90)}
  \pgfmathsetmacro{\Jbx}{0.375/0.7*5*cos(45)}   \pgfmathsetmacro{\Jby}{0.375/0.7*5*sin(45)}
  \pgfmathsetmacro{\Jcx}{0.4/0.7*5*cos(0)}      \pgfmathsetmacro{\Jcy}{0.4/0.7*5*sin(0)}
  \pgfmathsetmacro{\Jdx}{0.5833/0.7*5*cos(-45)} \pgfmathsetmacro{\Jdy}{0.5833/0.7*5*sin(-45)}
  \pgfmathsetmacro{\Jex}{0.4792/0.7*5*cos(-90)} \pgfmathsetmacro{\Jey}{0.4792/0.7*5*sin(-90)}
  \pgfmathsetmacro{\Jfx}{0.6042/0.7*5*cos(-135)}\pgfmathsetmacro{\Jfy}{0.6042/0.7*5*sin(-135)}
  \pgfmathsetmacro{\Jgx}{0.4792/0.7*5*cos(180)} \pgfmathsetmacro{\Jgy}{0.4792/0.7*5*sin(180)}
  \pgfmathsetmacro{\Jhx}{0.425/0.7*5*cos(135)}  \pgfmathsetmacro{\Jhy}{0.425/0.7*5*sin(135)}
  \fill[jtlime,opacity=0.35]
    (\Jax cm,\Jay cm)--(\Jbx cm,\Jby cm)--(\Jcx cm,\Jcy cm)--(\Jdx cm,\Jdy cm)--
    (\Jex cm,\Jey cm)--(\Jfx cm,\Jfy cm)--(\Jgx cm,\Jgy cm)--(\Jhx cm,\Jhy cm)--cycle;
  \draw[jtlime,thick]
    (\Jax cm,\Jay cm)--(\Jbx cm,\Jby cm)--(\Jcx cm,\Jcy cm)--(\Jdx cm,\Jdy cm)--
    (\Jex cm,\Jey cm)--(\Jfx cm,\Jfy cm)--(\Jgx cm,\Jgy cm)--(\Jhx cm,\Jhy cm)--cycle;
  \node[font=\Huge\bfseries] at (0,-7cm) {20\% labeled};
\end{scope}

\begin{scope}[xshift=51cm]
  \drawpanel{}
  \pgfmathsetmacro{\Pax}{0.4167/0.7*5*cos(90)}  \pgfmathsetmacro{\Pay}{0.4167/0.7*5*sin(90)}
  \pgfmathsetmacro{\Pbx}{0.5/0.7*5*cos(45)}     \pgfmathsetmacro{\Pby}{0.5/0.7*5*sin(45)}
  \pgfmathsetmacro{\Pcx}{0.4792/0.7*5*cos(0)}   \pgfmathsetmacro{\Pcy}{0.4792/0.7*5*sin(0)}
  \pgfmathsetmacro{\Pdx}{0.4375/0.7*5*cos(-45)} \pgfmathsetmacro{\Pdy}{0.4375/0.7*5*sin(-45)}
  \pgfmathsetmacro{\Pex}{0.4583/0.7*5*cos(-90)} \pgfmathsetmacro{\Pey}{0.4583/0.7*5*sin(-90)}
  \pgfmathsetmacro{\Pfx}{0.4792/0.7*5*cos(-135)}\pgfmathsetmacro{\Pfy}{0.4792/0.7*5*sin(-135)}
  \pgfmathsetmacro{\Pgx}{0.5417/0.7*5*cos(180)} \pgfmathsetmacro{\Pgy}{0.5417/0.7*5*sin(180)}
  \pgfmathsetmacro{\Phx}{0.6042/0.7*5*cos(135)} \pgfmathsetmacro{\Phy}{0.6042/0.7*5*sin(135)}
  \fill[pftpink,opacity=0.35]
    (\Pax cm,\Pay cm)--(\Pbx cm,\Pby cm)--(\Pcx cm,\Pcy cm)--(\Pdx cm,\Pdy cm)--
    (\Pex cm,\Pey cm)--(\Pfx cm,\Pfy cm)--(\Pgx cm,\Pgy cm)--(\Phx cm,\Phy cm)--cycle;
  \draw[pftpink,thick,dashed]
    (\Pax cm,\Pay cm)--(\Pbx cm,\Pby cm)--(\Pcx cm,\Pcy cm)--(\Pdx cm,\Pdy cm)--
    (\Pex cm,\Pey cm)--(\Pfx cm,\Pfy cm)--(\Pgx cm,\Pgy cm)--(\Phx cm,\Phy cm)--cycle;
  \pgfmathsetmacro{\Jax}{0.35/0.7*5*cos(90)}    \pgfmathsetmacro{\Jay}{0.35/0.7*5*sin(90)}
  \pgfmathsetmacro{\Jbx}{0.4/0.7*5*cos(45)}     \pgfmathsetmacro{\Jby}{0.4/0.7*5*sin(45)}
  \pgfmathsetmacro{\Jcx}{0.425/0.7*5*cos(0)}    \pgfmathsetmacro{\Jcy}{0.425/0.7*5*sin(0)}
  \pgfmathsetmacro{\Jdx}{0.4792/0.7*5*cos(-45)} \pgfmathsetmacro{\Jdy}{0.4792/0.7*5*sin(-45)}
  \pgfmathsetmacro{\Jex}{0.3125/0.7*5*cos(-90)} \pgfmathsetmacro{\Jey}{0.3125/0.7*5*sin(-90)}
  \pgfmathsetmacro{\Jfx}{0.5417/0.7*5*cos(-135)}\pgfmathsetmacro{\Jfy}{0.5417/0.7*5*sin(-135)}
  \pgfmathsetmacro{\Jgx}{0.4792/0.7*5*cos(180)} \pgfmathsetmacro{\Jgy}{0.4792/0.7*5*sin(180)}
  \pgfmathsetmacro{\Jhx}{0.45/0.7*5*cos(135)}   \pgfmathsetmacro{\Jhy}{0.45/0.7*5*sin(135)}
  \fill[jtlime,opacity=0.35]
    (\Jax cm,\Jay cm)--(\Jbx cm,\Jby cm)--(\Jcx cm,\Jcy cm)--(\Jdx cm,\Jdy cm)--
    (\Jex cm,\Jey cm)--(\Jfx cm,\Jfy cm)--(\Jgx cm,\Jgy cm)--(\Jhx cm,\Jhy cm)--cycle;
  \draw[jtlime,thick]
    (\Jax cm,\Jay cm)--(\Jbx cm,\Jby cm)--(\Jcx cm,\Jcy cm)--(\Jdx cm,\Jdy cm)--
    (\Jex cm,\Jey cm)--(\Jfx cm,\Jfy cm)--(\Jgx cm,\Jgy cm)--(\Jhx cm,\Jhy cm)--cycle;
  \node[font=\Huge\bfseries] at (0,-7cm) {10\% labeled};
\end{scope}

\begin{scope}[xshift=25.5cm, yshift=-8.5cm]
  \draw[jtlime,thick]         (-2cm,0)--(-.8cm,0);
  \node[font=\Huge,anchor=west] at (-.7cm,0) {JT};
  \draw[pftpink,thick,dashed] (2.5cm,0)--(3.7cm,0);
  \node[font=\Huge,anchor=west] at (3.8cm,0) {PFT};
\end{scope}
 
\end{tikzpicture}
}
\caption{Radar plots depicting PFT and JT performance across each of the explored SSL techniques for nodule classification on the JSRT dataset.}
\label{fig:radar_jsrt_cls}
\end{figure*}
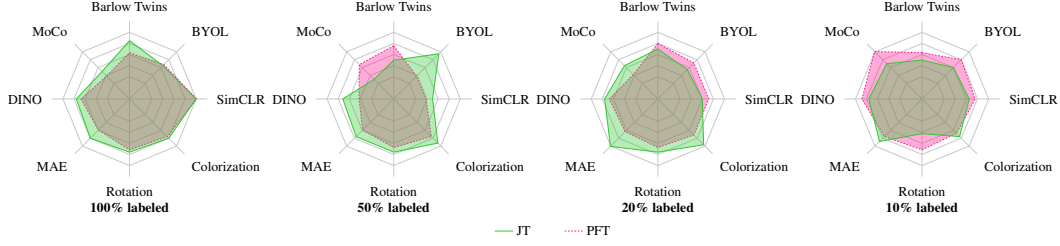

Fig.~\ref{fig:radar_jsrt_cls} shows radar plots of the PFT and JT performance of our eight SSL techniques on the multi-class chest x-ray classification task. Here, the classes being predicted are benign nodule, malignant nodule, and no nodule. Among all methods, colorization demonstrates the most consistent benefit from JT. Across every label fraction except the fully supervised setting, JT substantially improves both accuracy and F1 while simultaneously reducing training time. The gains are especially pronounced under limited supervision, where JT improves accuracy from 0.4583 to 0.5833 at 20\% labeled data and from 0.4792 to 0.5625 at 50\% labeled data. Similar improvements are observed in F1 score, indicating that the gains are not merely due to class imbalance effects.

MAE also benefits noticeably from JT, particularly in low-label regimes. At 20\% labeled data, JT improves accuracy from 0.4167 to 0.6042 and nearly doubles the F1 score relative to PFT. Unlike the segmentation experiments, where MAE remained comparatively weak, the reconstruction-based objective appears more compatible with image-level classification. This may indicate that masked reconstruction encourages global semantic reasoning that aligns more naturally with classification than with precise boundary localization. DINO exhibits a similar, though more moderate, trend, where JT consistently improves performance at moderate label fractions, particularly in the F1 score.

In contrast, the behavior of contrastive and redundancy-reduction methods is considerably less stable. SimCLR, MoCo, BYOL, and Barlow Twins exhibit highly inconsistent trends across label regimes, with JT sometimes improving performance and sometimes substantially degrading it. For example, BYOL JT achieves a large improvement at 50\% labeled data (accuracy: 0.2917 $\rightarrow$ 0.5750), yet underperforms PFT at 10\% and 20\% labels. Similarly, MoCo JT improves performance at 100\% and 20\% labels but collapses at 50\% and 10\% labels. These fluctuations suggest that the interaction between the supervised and self-supervised gradients is less stable for contrastive objectives in the JSRT classification setting.

Another important observation is that JT frequently reduces training time for lighter-weight SSL objectives such as colorization and rotation, but this efficiency advantage becomes inconsistent for larger contrastive frameworks. In several cases, including BYOL, DINO, and Barlow Twins, JT is actually more computationally expensive than PFT despite producing weaker performance. This contrasts with the segmentation experiments, where JT often achieved both faster convergence and higher accuracy. The discrepancy suggests that the computational efficiency benefits of JT may depend not only on the SSL framework itself, but also on the complexity and stability of the downstream task. Table~\ref{tab:jsrt_class} provides more detailed results on this task.

\begin{table}
    \centering
    \caption{Classification performance on JSRT under different training schemes.}
    \resizebox{0.4\textheight}{!}{
    \begin{tabular}{llcccc}
        \toprule
        \textbf{Framework} & \textbf{Config} & \textbf{Label (\%)} & \textbf{Accuracy} & \textbf{F1} & \textbf{Training Time (s)}\\
        \midrule
        \multirow{8}{*}{\textbf{Colorization}} 
        & PFT & \multirow{2}{*}{100\%} & 0.4792 & 0.3238 & 112.8900 \\
        & JT & & 0.5000 & 0.3370 & 101.4200 \\
        & PFT & \multirow{2}{*}{50\%} & 0.4792 & 0.2574 & 100.8400 \\
        & JT & & 0.5625 & 0.3969 & 87.6300 \\
        & PFT & \multirow{2}{*}{20\%} & 0.4583 & 0.2743 & 88.0200 \\
        & JT & & 0.5833 & 0.4079 & 77.1000 \\
        & PFT & \multirow{2}{*}{10\%} & 0.4375 & 0.3113 & 95.2400 \\
        & JT & & 0.4792 & 0.4010 & 76.0100 \\
        \midrule
        \multirow{8}{*}{\textbf{Rotation}} 
        & PFT & \multirow{2}{*}{100\%} & 0.4583 & 0.2993 & 100.8700 \\
        & JT & & 0.4792 & 0.4010 & 87.7600 \\
        & PFT & \multirow{2}{*}{50\%} & 0.4375 & 0.2890 & 84.9500 \\
        & JT & & 0.4792 & 0.4010 & 74.9200 \\
        & PFT & \multirow{2}{*}{20\%} & 0.4375 & 0.2925 & 75.2000 \\
        & JT & & 0.4792 & 0.4010 & 65.1300 \\
        & PFT & \multirow{2}{*}{10\%} & 0.4583 & 0.2717 & 69.1700 \\
        & JT & & 0.3125 & 0.2191 & 62.1300 \\
        \midrule
        \multirow{8}{*}{\textbf{SimCLR}} 
        & PFT & \multirow{2}{*}{100\%} & 0.6042 & 0.5395 & 275.3956 \\
        & JT & & 0.6000 & 0.4530 & 175.9404 \\
        & PFT & \multirow{2}{*}{50\%} & 0.2917 & 0.1726 & 244.2773 \\
        & JT & & 0.3500 & 0.1728 & 239.7501 \\
        & PFT & \multirow{2}{*}{20\%} & 0.4583 & 0.3113 & 224.5410 \\
        & JT & & 0.4000 & 0.3172 & 265.2517 \\
        & PFT & \multirow{2}{*}{10\%} & 0.4792 & 0.4010 & 253.5416 \\
        & JT & & 0.4250 & 0.3589 & 236.9074 \\
        \midrule
        \multirow{8}{*}{\textbf{BYOL}} 
        & PFT & \multirow{2}{*}{100\%} & 0.4375 & 0.2928 & 268.8473 \\
        & JT & & 0.4250 & 0.3121 & 164.6826 \\
        & PFT & \multirow{2}{*}{50\%} & 0.2917 & 0.2186 & 240.2865 \\
        & JT & &  0.5750 & 0.3247 & 265.1637 \\
        & PFT & \multirow{2}{*}{20\%} & 0.4583 & 0.2592 & 221.6456 \\
        & JT & & 0.3750 & 0.3135 & 258.2225 \\
        & PFT & \multirow{2}{*}{10\%} & 0.5000 & 0.2410 & 217.2265 \\
        & JT & & 0.4000 & 0.3116 & 229.8280 \\
        \midrule
        \multirow{8}{*}{\textbf{MoCo}} 
        & PFT & \multirow{2}{*}{100\%} & 0.2917 &  0.1726 & 276.0354 \\
        & JT & & 0.3500 & 0.1761 & 172.8037 \\
        & PFT & \multirow{2}{*}{50\%} & 0.4375 & 0.3189 & 247.0806 \\
        & JT & & 0.2500 & 0.1601 & 233.0023 \\
        & PFT & \multirow{2}{*}{20\%} & 0.3125 & 0.1686 & 223.9100 \\
        & JT & & 0.4250 & 0.1988 & 227.8016 \\
        & PFT & \multirow{2}{*}{10\%} & 0.6042 & 0.4382 & 220.8522 \\
        & JT & & 0.4500 & 0.3299 & 226.8130 \\
        \midrule
        \multirow{8}{*}{\textbf{MAE}} 
        & PFT & \multirow{2}{*}{100\%} & 0.3958 & 0.2508 & 221.3400 \\
        & JT & & 0.5000 & 0.3221 & 229.6600 \\
        & PFT & \multirow{2}{*}{50\%} & 0.3958 & 0.2453 & 179.1000 \\
        & JT & & 0.4792 & 0.4010 & 168.4500 \\
        & PFT & \multirow{2}{*}{20\%} & 0.4167 & 0.2676 & 144.0800 \\
        & JT & & 0.6042 & 0.5147 & 155.8300 \\
        & PFT & \multirow{2}{*}{10\%} & 0.4792 & 0.4010 & 121.5900 \\
        & JT & & 0.5417 & 0.3339 & 147.5900 \\
        \midrule
        \multirow{8}{*}{\textbf{DINO}} 
        & PFT & \multirow{2}{*}{100\%} & 0.4375 & 0.2401 & 247.4900 \\
        & JT & & 0.4792 & 0.4010 & 271.5200 \\
        & PFT & \multirow{2}{*}{50\%} & 0.3125 & 0.1818 & 233.2000 \\
        & JT & & 0.4583 & 0.3939 & 265.0900 \\
        & PFT & \multirow{2}{*}{20\%} & 0.4375 & 0.2593 & 221.5100 \\
        & JT & & 0.4792 & 0.4010 & 241.3900 \\
        & PFT & \multirow{2}{*}{10\%} & 0.5417 & 0.4486 & 217.7300 \\
        & JT & & 0.4792 & 0.4010 & 230.4600 \\
        \midrule
        \multirow{8}{*}{\textbf{Barlow Twins}} 
        & PFT & \multirow{2}{*}{100\%} & 0.4167 & 0.2800 & 268.1816 \\
        & JT & & 0.5250 & 0.4091 & 158.9600 \\
        & PFT & \multirow{2}{*}{50\%} & 0.4792 & 0.3085 & 239.5399 \\
        & JT & & 0.3500 & 0.2773 & 271.8517 \\
        & PFT & \multirow{2}{*}{20\%} & 0.5000 & 0.3393 & 217.3146 \\
        & JT & & 0.4500 & 0.3194 & 240.7588 \\
        & PFT & \multirow{2}{*}{10\%} & 0.4167 & 0.2119 & 215.6108 \\
        & JT & & 0.3500 & 0.2046 & 239.1821 \\
        \bottomrule
    \end{tabular}
    }
    \label{tab:jsrt_class}
\end{table}

\clearpage

\subsection{Dermatology Classification}
\label{appendix:isic_cls}

\begin{figure}[h]
\resizebox{\textwidth}{!}{
\begin{tikzpicture}

\definecolor{pftpink}{RGB}{255,20,147}
\definecolor{jtlime}{RGB}{50,200,50}
\definecolor{gridc}{RGB}{200,200,200}

\newcommand{\drawpanel}[1]{%
  \foreach \v/\lab in {0.50/0.50, 0.55/0.55, 0.60/0.60, 0.65/0.65, 0.70/0.70}{
    \pgfmathsetmacro{\rr}{\v/0.70*5}
    \draw[gridc,thin]
      (90:\rr cm)--(45:\rr cm)--(0:\rr cm)--(-45:\rr cm)--
      (-90:\rr cm)--(-135:\rr cm)--(180:\rr cm)--(135:\rr cm)--cycle;
  }
  \foreach \i in {0,...,7}{
    \pgfmathsetmacro{\ang}{90-\i*45}
    \draw[gridc,thin] (0,0)--(\ang:5cm);
  }
  \node[font=\Huge,above]       at (90:5.5cm)   {Barlow Twins};
  \node[font=\Huge,above right] at (45:5.5cm)   {BYOL};
  \node[font=\Huge,right]       at (0:5.5cm)    {SimCLR};
  \node[font=\Huge,below right] at (-45:5.5cm)  {Colorization};
  \node[font=\Huge,below]       at (-90:5.5cm)  {Rotation};
  \node[font=\Huge,below left]  at (-135:5.5cm) {MAE};
  \node[font=\Huge,left]        at (180:5.5cm)  {DINO};
  \node[font=\Huge,above left]  at (135:5.5cm)  {MoCo};
}

\begin{scope}[xshift=0cm]
  \drawpanel{}
  \pgfmathsetmacro{\Pax}{0.579/0.70*5*cos(90)}   \pgfmathsetmacro{\Pay}{0.579/0.70*5*sin(90)}
  \pgfmathsetmacro{\Pbx}{0.6345/0.70*5*cos(45)}  \pgfmathsetmacro{\Pby}{0.6345/0.70*5*sin(45)}
  \pgfmathsetmacro{\Pcx}{0.6696/0.70*5*cos(0)}   \pgfmathsetmacro{\Pcy}{0.6696/0.70*5*sin(0)}
  \pgfmathsetmacro{\Pdx}{0.4966/0.70*5*cos(-45)} \pgfmathsetmacro{\Pdy}{0.4966/0.70*5*sin(-45)}
  \pgfmathsetmacro{\Pex}{0.6617/0.70*5*cos(-90)} \pgfmathsetmacro{\Pey}{0.6617/0.70*5*sin(-90)}
  \pgfmathsetmacro{\Pfx}{0.5694/0.70*5*cos(-135)}\pgfmathsetmacro{\Pfy}{0.5694/0.70*5*sin(-135)}
  \pgfmathsetmacro{\Pgx}{0.574/0.70*5*cos(180)}  \pgfmathsetmacro{\Pgy}{0.574/0.70*5*sin(180)}
  \pgfmathsetmacro{\Phx}{0.6376/0.70*5*cos(135)} \pgfmathsetmacro{\Phy}{0.6376/0.70*5*sin(135)}
  \fill[pftpink,opacity=0.35]
    (\Pax cm,\Pay cm)--(\Pbx cm,\Pby cm)--(\Pcx cm,\Pcy cm)--(\Pdx cm,\Pdy cm)--
    (\Pex cm,\Pey cm)--(\Pfx cm,\Pfy cm)--(\Pgx cm,\Pgy cm)--(\Phx cm,\Phy cm)--cycle;
  \draw[pftpink,thick,dashed]
    (\Pax cm,\Pay cm)--(\Pbx cm,\Pby cm)--(\Pcx cm,\Pcy cm)--(\Pdx cm,\Pdy cm)--
    (\Pex cm,\Pey cm)--(\Pfx cm,\Pfy cm)--(\Pgx cm,\Pgy cm)--(\Phx cm,\Phy cm)--cycle;
  \pgfmathsetmacro{\Jax}{0.5/0.70*5*cos(90)}     \pgfmathsetmacro{\Jay}{0.5/0.70*5*sin(90)}
  \pgfmathsetmacro{\Jbx}{0.5/0.70*5*cos(45)}     \pgfmathsetmacro{\Jby}{0.5/0.70*5*sin(45)}
  \pgfmathsetmacro{\Jcx}{0.5829/0.70*5*cos(0)}   \pgfmathsetmacro{\Jcy}{0.5829/0.70*5*sin(0)}
  \pgfmathsetmacro{\Jdx}{0.5169/0.70*5*cos(-45)} \pgfmathsetmacro{\Jdy}{0.5169/0.70*5*sin(-45)}
  \pgfmathsetmacro{\Jex}{0.6625/0.70*5*cos(-90)} \pgfmathsetmacro{\Jey}{0.6625/0.70*5*sin(-90)}
  \pgfmathsetmacro{\Jfx}{0.567/0.70*5*cos(-135)} \pgfmathsetmacro{\Jfy}{0.567/0.70*5*sin(-135)}
  \pgfmathsetmacro{\Jgx}{0.6023/0.70*5*cos(180)} \pgfmathsetmacro{\Jgy}{0.6023/0.70*5*sin(180)}
  \pgfmathsetmacro{\Jhx}{0.5686/0.70*5*cos(135)} \pgfmathsetmacro{\Jhy}{0.5686/0.70*5*sin(135)}
  \fill[jtlime,opacity=0.35]
    (\Jax cm,\Jay cm)--(\Jbx cm,\Jby cm)--(\Jcx cm,\Jcy cm)--(\Jdx cm,\Jdy cm)--
    (\Jex cm,\Jey cm)--(\Jfx cm,\Jfy cm)--(\Jgx cm,\Jgy cm)--(\Jhx cm,\Jhy cm)--cycle;
  \draw[jtlime,thick]
    (\Jax cm,\Jay cm)--(\Jbx cm,\Jby cm)--(\Jcx cm,\Jcy cm)--(\Jdx cm,\Jdy cm)--
    (\Jex cm,\Jey cm)--(\Jfx cm,\Jfy cm)--(\Jgx cm,\Jgy cm)--(\Jhx cm,\Jhy cm)--cycle;
  \node[font=\Huge\bfseries] at (0,-7cm) {100\% labeled};
\end{scope}

\begin{scope}[xshift=17cm]
  \drawpanel{}
  \pgfmathsetmacro{\Pax}{0.5746/0.70*5*cos(90)}  \pgfmathsetmacro{\Pay}{0.5746/0.70*5*sin(90)}
  \pgfmathsetmacro{\Pbx}{0.6096/0.70*5*cos(45)}  \pgfmathsetmacro{\Pby}{0.6096/0.70*5*sin(45)}
  \pgfmathsetmacro{\Pcx}{0.5339/0.70*5*cos(0)}   \pgfmathsetmacro{\Pcy}{0.5339/0.70*5*sin(0)}
  \pgfmathsetmacro{\Pdx}{0.4966/0.70*5*cos(-45)} \pgfmathsetmacro{\Pdy}{0.4966/0.70*5*sin(-45)}
  \pgfmathsetmacro{\Pex}{0.6245/0.70*5*cos(-90)} \pgfmathsetmacro{\Pey}{0.6245/0.70*5*sin(-90)}
  \pgfmathsetmacro{\Pfx}{0.51/0.70*5*cos(-135)}  \pgfmathsetmacro{\Pfy}{0.51/0.70*5*sin(-135)}
  \pgfmathsetmacro{\Pgx}{0.6289/0.70*5*cos(180)} \pgfmathsetmacro{\Pgy}{0.6289/0.70*5*sin(180)}
  \pgfmathsetmacro{\Phx}{0.5976/0.70*5*cos(135)} \pgfmathsetmacro{\Phy}{0.5976/0.70*5*sin(135)}
  \fill[pftpink,opacity=0.35]
    (\Pax cm,\Pay cm)--(\Pbx cm,\Pby cm)--(\Pcx cm,\Pcy cm)--(\Pdx cm,\Pdy cm)--
    (\Pex cm,\Pey cm)--(\Pfx cm,\Pfy cm)--(\Pgx cm,\Pgy cm)--(\Phx cm,\Phy cm)--cycle;
  \draw[pftpink,thick,dashed]
    (\Pax cm,\Pay cm)--(\Pbx cm,\Pby cm)--(\Pcx cm,\Pcy cm)--(\Pdx cm,\Pdy cm)--
    (\Pex cm,\Pey cm)--(\Pfx cm,\Pfy cm)--(\Pgx cm,\Pgy cm)--(\Phx cm,\Phy cm)--cycle;
  \pgfmathsetmacro{\Jax}{0.5/0.70*5*cos(90)}     \pgfmathsetmacro{\Jay}{0.5/0.70*5*sin(90)}
  \pgfmathsetmacro{\Jbx}{0.5319/0.70*5*cos(45)}  \pgfmathsetmacro{\Jby}{0.5319/0.70*5*sin(45)}
  \pgfmathsetmacro{\Jcx}{0.583/0.70*5*cos(0)}    \pgfmathsetmacro{\Jcy}{0.583/0.70*5*sin(0)}
  \pgfmathsetmacro{\Jdx}{0.5505/0.70*5*cos(-45)} \pgfmathsetmacro{\Jdy}{0.5505/0.70*5*sin(-45)}
  \pgfmathsetmacro{\Jex}{0.558/0.70*5*cos(-90)}  \pgfmathsetmacro{\Jey}{0.558/0.70*5*sin(-90)}
  \pgfmathsetmacro{\Jfx}{0.5201/0.70*5*cos(-135)}\pgfmathsetmacro{\Jfy}{0.5201/0.70*5*sin(-135)}
  \pgfmathsetmacro{\Jgx}{0.5989/0.70*5*cos(180)} \pgfmathsetmacro{\Jgy}{0.5989/0.70*5*sin(180)}
  \pgfmathsetmacro{\Jhx}{0.5452/0.70*5*cos(135)} \pgfmathsetmacro{\Jhy}{0.5452/0.70*5*sin(135)}
  \fill[jtlime,opacity=0.35]
    (\Jax cm,\Jay cm)--(\Jbx cm,\Jby cm)--(\Jcx cm,\Jcy cm)--(\Jdx cm,\Jdy cm)--
    (\Jex cm,\Jey cm)--(\Jfx cm,\Jfy cm)--(\Jgx cm,\Jgy cm)--(\Jhx cm,\Jhy cm)--cycle;
  \draw[jtlime,thick]
    (\Jax cm,\Jay cm)--(\Jbx cm,\Jby cm)--(\Jcx cm,\Jcy cm)--(\Jdx cm,\Jdy cm)--
    (\Jex cm,\Jey cm)--(\Jfx cm,\Jfy cm)--(\Jgx cm,\Jgy cm)--(\Jhx cm,\Jhy cm)--cycle;
  \node[font=\Huge\bfseries] at (0,-7cm) {50\% labeled};
\end{scope}

\begin{scope}[xshift=34cm]
  \drawpanel{}
  \pgfmathsetmacro{\Pax}{0.5245/0.70*5*cos(90)}  \pgfmathsetmacro{\Pay}{0.5245/0.70*5*sin(90)}
  \pgfmathsetmacro{\Pbx}{0.574/0.70*5*cos(45)}   \pgfmathsetmacro{\Pby}{0.574/0.70*5*sin(45)}
  \pgfmathsetmacro{\Pcx}{0.5653/0.70*5*cos(0)}   \pgfmathsetmacro{\Pcy}{0.5653/0.70*5*sin(0)}
  \pgfmathsetmacro{\Pdx}{0.5/0.70*5*cos(-45)}    \pgfmathsetmacro{\Pdy}{0.5/0.70*5*sin(-45)}
  \pgfmathsetmacro{\Pex}{0.6114/0.70*5*cos(-90)} \pgfmathsetmacro{\Pey}{0.6114/0.70*5*sin(-90)}
  \pgfmathsetmacro{\Pfx}{0.5184/0.70*5*cos(-135)}\pgfmathsetmacro{\Pfy}{0.5184/0.70*5*sin(-135)}
  \pgfmathsetmacro{\Pgx}{0.6096/0.70*5*cos(180)} \pgfmathsetmacro{\Pgy}{0.6096/0.70*5*sin(180)}
  \pgfmathsetmacro{\Phx}{0.6524/0.70*5*cos(135)} \pgfmathsetmacro{\Phy}{0.6524/0.70*5*sin(135)}
  \fill[pftpink,opacity=0.35]
    (\Pax cm,\Pay cm)--(\Pbx cm,\Pby cm)--(\Pcx cm,\Pcy cm)--(\Pdx cm,\Pdy cm)--
    (\Pex cm,\Pey cm)--(\Pfx cm,\Pfy cm)--(\Pgx cm,\Pgy cm)--(\Phx cm,\Phy cm)--cycle;
  \draw[pftpink,thick,dashed]
    (\Pax cm,\Pay cm)--(\Pbx cm,\Pby cm)--(\Pcx cm,\Pcy cm)--(\Pdx cm,\Pdy cm)--
    (\Pex cm,\Pey cm)--(\Pfx cm,\Pfy cm)--(\Pgx cm,\Pgy cm)--(\Phx cm,\Phy cm)--cycle;
  \pgfmathsetmacro{\Jax}{0.5/0.70*5*cos(90)}     \pgfmathsetmacro{\Jay}{0.5/0.70*5*sin(90)}
  \pgfmathsetmacro{\Jbx}{0.5217/0.70*5*cos(45)}  \pgfmathsetmacro{\Jby}{0.5217/0.70*5*sin(45)}
  \pgfmathsetmacro{\Jcx}{0.5548/0.70*5*cos(0)}   \pgfmathsetmacro{\Jcy}{0.5548/0.70*5*sin(0)}
  \pgfmathsetmacro{\Jdx}{0.532/0.70*5*cos(-45)}  \pgfmathsetmacro{\Jdy}{0.532/0.70*5*sin(-45)}
  \pgfmathsetmacro{\Jex}{0.5577/0.70*5*cos(-90)} \pgfmathsetmacro{\Jey}{0.5577/0.70*5*sin(-90)}
  \pgfmathsetmacro{\Jfx}{0.5/0.70*5*cos(-135)}   \pgfmathsetmacro{\Jfy}{0.5/0.70*5*sin(-135)}
  \pgfmathsetmacro{\Jgx}{0.5639/0.70*5*cos(180)} \pgfmathsetmacro{\Jgy}{0.5639/0.70*5*sin(180)}
  \pgfmathsetmacro{\Jhx}{0.5552/0.70*5*cos(135)} \pgfmathsetmacro{\Jhy}{0.5552/0.70*5*sin(135)}
  \fill[jtlime,opacity=0.35]
    (\Jax cm,\Jay cm)--(\Jbx cm,\Jby cm)--(\Jcx cm,\Jcy cm)--(\Jdx cm,\Jdy cm)--
    (\Jex cm,\Jey cm)--(\Jfx cm,\Jfy cm)--(\Jgx cm,\Jgy cm)--(\Jhx cm,\Jhy cm)--cycle;
  \draw[jtlime,thick]
    (\Jax cm,\Jay cm)--(\Jbx cm,\Jby cm)--(\Jcx cm,\Jcy cm)--(\Jdx cm,\Jdy cm)--
    (\Jex cm,\Jey cm)--(\Jfx cm,\Jfy cm)--(\Jgx cm,\Jgy cm)--(\Jhx cm,\Jhy cm)--cycle;
  \node[font=\Huge\bfseries] at (0,-7cm) {20\% labeled};
\end{scope}

\begin{scope}[xshift=51cm]
  \drawpanel{}
  \pgfmathsetmacro{\Pax}{0.4886/0.70*5*cos(90)}  \pgfmathsetmacro{\Pay}{0.4886/0.70*5*sin(90)}
  \pgfmathsetmacro{\Pbx}{0.5769/0.70*5*cos(45)}  \pgfmathsetmacro{\Pby}{0.5769/0.70*5*sin(45)}
  \pgfmathsetmacro{\Pcx}{0.5775/0.70*5*cos(0)}   \pgfmathsetmacro{\Pcy}{0.5775/0.70*5*sin(0)}
  \pgfmathsetmacro{\Pdx}{0.4966/0.70*5*cos(-45)} \pgfmathsetmacro{\Pdy}{0.4966/0.70*5*sin(-45)}
  \pgfmathsetmacro{\Pex}{0.5/0.70*5*cos(-90)}    \pgfmathsetmacro{\Pey}{0.5/0.70*5*sin(-90)}
  \pgfmathsetmacro{\Pfx}{0.5235/0.70*5*cos(-135)}\pgfmathsetmacro{\Pfy}{0.5235/0.70*5*sin(-135)}
  \pgfmathsetmacro{\Pgx}{0.5067/0.70*5*cos(180)} \pgfmathsetmacro{\Pgy}{0.5067/0.70*5*sin(180)}
  \pgfmathsetmacro{\Phx}{0.585/0.70*5*cos(135)}  \pgfmathsetmacro{\Phy}{0.585/0.70*5*sin(135)}
  \fill[pftpink,opacity=0.35]
    (\Pax cm,\Pay cm)--(\Pbx cm,\Pby cm)--(\Pcx cm,\Pcy cm)--(\Pdx cm,\Pdy cm)--
    (\Pex cm,\Pey cm)--(\Pfx cm,\Pfy cm)--(\Pgx cm,\Pgy cm)--(\Phx cm,\Phy cm)--cycle;
  \draw[pftpink,thick,dashed]
    (\Pax cm,\Pay cm)--(\Pbx cm,\Pby cm)--(\Pcx cm,\Pcy cm)--(\Pdx cm,\Pdy cm)--
    (\Pex cm,\Pey cm)--(\Pfx cm,\Pfy cm)--(\Pgx cm,\Pgy cm)--(\Phx cm,\Phy cm)--cycle;
  \pgfmathsetmacro{\Jax}{0.5/0.70*5*cos(90)}     \pgfmathsetmacro{\Jay}{0.5/0.70*5*sin(90)}
  \pgfmathsetmacro{\Jbx}{0.5617/0.70*5*cos(45)}  \pgfmathsetmacro{\Jby}{0.5617/0.70*5*sin(45)}
  \pgfmathsetmacro{\Jcx}{0.5368/0.70*5*cos(0)}   \pgfmathsetmacro{\Jcy}{0.5368/0.70*5*sin(0)}
  \pgfmathsetmacro{\Jdx}{0.6008/0.70*5*cos(-45)} \pgfmathsetmacro{\Jdy}{0.6008/0.70*5*sin(-45)}
  \pgfmathsetmacro{\Jex}{0.5688/0.70*5*cos(-90)} \pgfmathsetmacro{\Jey}{0.5688/0.70*5*sin(-90)}
  \pgfmathsetmacro{\Jfx}{0.5524/0.70*5*cos(-135)}\pgfmathsetmacro{\Jfy}{0.5524/0.70*5*sin(-135)}
  \pgfmathsetmacro{\Jgx}{0.5482/0.70*5*cos(180)} \pgfmathsetmacro{\Jgy}{0.5482/0.70*5*sin(180)}
  \pgfmathsetmacro{\Jhx}{0.542/0.70*5*cos(135)}  \pgfmathsetmacro{\Jhy}{0.542/0.70*5*sin(135)}
  \fill[jtlime,opacity=0.35]
    (\Jax cm,\Jay cm)--(\Jbx cm,\Jby cm)--(\Jcx cm,\Jcy cm)--(\Jdx cm,\Jdy cm)--
    (\Jex cm,\Jey cm)--(\Jfx cm,\Jfy cm)--(\Jgx cm,\Jgy cm)--(\Jhx cm,\Jhy cm)--cycle;
  \draw[jtlime,thick]
    (\Jax cm,\Jay cm)--(\Jbx cm,\Jby cm)--(\Jcx cm,\Jcy cm)--(\Jdx cm,\Jdy cm)--
    (\Jex cm,\Jey cm)--(\Jfx cm,\Jfy cm)--(\Jgx cm,\Jgy cm)--(\Jhx cm,\Jhy cm)--cycle;
  \node[font=\Huge\bfseries] at (0,-7cm) {10\% labeled};
\end{scope}

\begin{scope}[xshift=25.5cm, yshift=-8.5cm]
  \draw[jtlime,thick]      (-2cm,0)--(-.8cm,0);
  \node[font=\Huge,anchor=west] at (-.7cm,0) {JT};
  \draw[pftpink,thick,dashed] (2.5cm,0)--(3.7cm,0);
  \node[font=\Huge,anchor=west] at (3.8cm,0) {PFT};
\end{scope}

\end{tikzpicture}
}
\caption{Radar plots depicting the accuracy of skin lesion classification on the ISIC dataset using both PFT and JT training paradigms across eight representative SSL techniques.}
\label{fig:radar_isic_cls}
\end{figure}
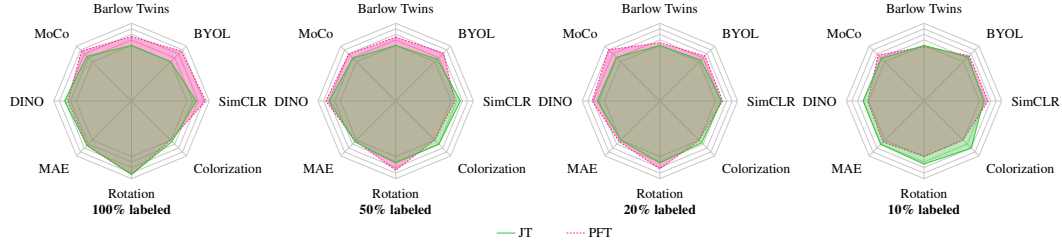

Fig.~\ref{fig:radar_isic_cls} highlights the classification behavior of different SSL methods across varying label fractions on the ISIC-2016 dataset. Table~\ref{tab:isic_cls} shows that PFT generally provides stronger and more stable accuracy and F1 performance for contrastive methods such as SimCLR, BYOL, MoCo, and DINO, especially at higher label fractions, where the frozen pretrained features already provide strong class separation. At full supervision, for instance, MoCo and DINO under PFT reach F1 scores of 0.6390 and 0.5812, respectively, exceeding their JT counterparts. In contrast, reconstruction-oriented and auxiliary-task methods such as Colorization, Rotation, and MAE become more competitive under JT, particularly when labels are scarce. Colorization improves from 0.4431 to 0.5964 F1 at 10\% labels, and MAE from 0.5016 to 0.5565, suggesting that joint optimization helps these objectives recover task-relevant structure that linear probing alone cannot. Unlike the segmentation experiments, the classification results exhibit substantially larger performance variation across SSL objectives, particularly due to the strong class imbalance of the dataset. The results further show that low-label settings amplify performance instability across methods, indicating that representation quality and minority-class sensitivity play a critical role in dermatology classification performance.

\begin{table}
    \centering
    \caption{Classification performance on ISIC-2016 under different training schemes.}
    \resizebox{0.4\textheight}{!}{
    \begin{tabular}{llcccc}
        \toprule
        \textbf{Framework} & \textbf{Config} & \textbf{Label (\%)} & \textbf{Accuracy} & \textbf{F1} & \textbf{Training Time (s)}\\
        \midrule
\multirow{8}{*}{Colorization} & PFT & \multirow{2}{*}{100} & 0.4966 & 0.4431 & 1260 \\
          & JT  &  & 0.5169 & 0.5071 & 300  \\
          & PFT & \multirow{2}{*}{50}  & 0.4966 & 0.4431 & 1200 \\
          & JT  &  & 0.5505 & 0.5450 & 240  \\
          & PFT & \multirow{2}{*}{20}  & 0.5000 & 0.4448 & 960  \\
          & JT  &  & 0.5320 & 0.5195 & 300  \\
          & PFT & \multirow{2}{*}{10}  & 0.4966 & 0.4431 & 1140 \\
          & JT  &  & 0.6008 & 0.5964 & 360  \\
          \midrule
\multirow{8}{*}{Rotation}  
        & PFT & \multirow{2}{*}{100} & 0.6617 & 0.6098 & 1080 \\
          & JT  &  & 0.6625 & 0.6679 & 600  \\
          & PFT & \multirow{2}{*}{50}  & 0.6245 & 0.6252 & 1140 \\
          & JT  &  & 0.5580 & 0.5610 & 660  \\
          & PFT & \multirow{2}{*}{20}  & 0.6114 & 0.5196 & 1140 \\
          & JT  &  & 0.5577 & 0.5656 & 1320 \\
          & PFT & \multirow{2}{*}{10}  & 0.5000 & 0.4448 & 960  \\
          & JT  &  & 0.5688 & 0.5695 & 1080 \\
          \midrule
\multirow{8}{*}{SimCLR}     
        & PFT & \multirow{2}{*}{100} & 0.6656 & 0.5496 & 1560 \\
          & JT  &  & 0.5829 & 0.5803 & 540  \\
          & PFT & \multirow{2}{*}{50}  & 0.5339 & 0.3070 & 1680 \\
          & JT  &  & 0.5830 & 0.5781 & 780  \\
          & PFT & \multirow{2}{*}{20}  & 0.5653 & 0.5716 & 1740 \\
          & JT  &  & 0.5548 & 0.3880 & 540  \\
          & PFT & \multirow{2}{*}{10}  & 0.6046 & 0.5302 & 1620 \\
          & JT  &  & 0.5368 & 0.5243 & 360  \\
          \midrule
\multirow{8}{*}{BYOL}    
        & PFT & \multirow{2}{*}{100} & 0.6345 & 0.5517 & 1380 \\
          & JT  &  & 0.5000 & 0.4451 & 360  \\
          & PFT & \multirow{2}{*}{50}  & 0.6096 & 0.6010 & 1620 \\
          & JT  &  & 0.5319 & 0.5192 & 300  \\
          & PFT & \multirow{2}{*}{20}  & 0.5740 & 0.5812 & 1440 \\
          & JT  &  & 0.5217 & 0.4940 & 420  \\
          & PFT & \multirow{2}{*}{10}  & 0.5769 & 0.3614 & 1380 \\
          & JT  &  & 0.5617 & 0.5629 & 420  \\
        \midrule
\multirow{8}{*}{MoCo}   
        & PFT & \multirow{2}{*}{100} & 0.6376 & 0.6390 & 2940 \\
          & JT  &  & 0.5686 & 0.5759 & 360  \\
          & PFT & \multirow{2}{*}{50}  & 0.5976 & 0.4433 & 3060 \\
          & JT  &  & 0.5452 & 0.5404 & 540  \\
          & PFT & \multirow{2}{*}{20}  & 0.6524 & 0.5837 & 3000 \\
          & JT  &  & 0.5552 & 0.5567 & 480  \\
          & PFT & \multirow{2}{*}{10}  & 0.5850 & 0.5701 & 3300 \\
          & JT  &  & 0.5420 & 0.5392 & 360  \\
          \midrule
\multirow{8}{*}{MAE}   
        & PFT & \multirow{2}{*}{100} & 0.5694 & 0.5713 & 960  \\
          & JT  &  & 0.5670 & 0.5744 & 300  \\
          & PFT & \multirow{2}{*}{50}  & 0.5100 & 0.4701 & 1080 \\
          & JT  &  & 0.5201 & 0.4926 & 420  \\
          & PFT & \multirow{2}{*}{20}  & 0.5184 & 0.4912 & 1020 \\
          & JT  &  & 0.5000 & 0.4451 & 60   \\
          & PFT & \multirow{2}{*}{10}  & 0.5235 & 0.5016 & 900  \\
          & JT  &  & 0.5524 & 0.5565 & 480  \\
          \midrule
\multirow{8}{*}{DINO}     
        & PFT & \multirow{2}{*}{100} & 0.5740 & 0.5812 & 3660 \\
          & JT  &  & 0.6023 & 0.6123 & 480  \\
          & PFT & \multirow{2}{*}{50}  & 0.6289 & 0.5794 & 3900 \\
          & JT  &  & 0.5989 & 0.6115 & 600  \\
          & PFT & \multirow{2}{*}{20}  & 0.6096 & 0.6010 & 3840 \\
          & JT  &  & 0.5639 & 0.5704 & 840  \\
          & PFT & \multirow{2}{*}{10}  & 0.5067 & 0.3446 & 3660 \\
          & JT  &  & 0.5482 & 0.5430 & 360  \\
        \midrule
\multirow{8}{*}{Barlow Twins} 
        & PFT & \multirow{2}{*}{100} & 0.5790 & 0.5875 & 1440 \\
          & JT  &  & 0.5000 & 0.4451 & 360  \\
          & PFT & \multirow{2}{*}{50}  & 0.5746 & 0.4103 & 1380 \\
          & JT  &  & 0.5000 & 0.4451 & 420  \\
          & PFT & \multirow{2}{*}{20}  & 0.5245 & 0.4874 & 1440 \\
          & JT  &  & 0.5000 & 0.4451 & 420  \\
          & PFT & \multirow{2}{*}{10}  & 0.4886 & 0.4482 & 1260 \\
          & JT  &  & 0.5000 & 0.4451 & 360  \\  
          \bottomrule
    \end{tabular}
    }
    \label{tab:isic_cls}
\end{table}

\clearpage

\section{Object Detection Performance}
\label{appendix:coco}

\begin{table}[h]
\centering
\caption{Object detection performance on COCO under different training schemes.}
\resizebox{\linewidth}{!}{
\begin{tabular}{l l c c c c c c}
\toprule
\textbf{Framework} & \textbf{Config} & \textbf{Label(\%)} & \textbf{P} & \textbf{R} & \textbf{mAP} & \textbf{mAP50--95} & \textbf{Training Time (hrs)}\\
\midrule
\multirow{4}{*}{\textbf{SimCLR}} 
 & PFT & \multirow{2}{*}{100\%} & 0.688 & 0.555 & 0.612 & 0.450 & 54.378\\
 & JT &  & 0.698 & 0.547 & 0.608 & 0.449 & 42.269\\
 & PFT & \multirow{2}{*}{10\%} & 0.497 & 0.377 & 0.382 & 0.257 & 16.843\\
 & JT &   & 0.507 & 0.370 & 0.379 & 0.257 & 4.415\\
\midrule
\multirow{4}{*}{\textbf{BYOL}} 
 & PFT & \multirow{2}{*}{100\%} & 0.694 & 0.552 & 0.607 & 0.447 & 46.004\\
 & JT &   & 0.694 & 0.557 & 0.611 & 0.447 & 37.744\\
 & PFT & \multirow{2}{*}{10\%} & 0.495 & 0.368 & 0.374 & 0.254 & 13.094\\
 & JT &   & 0.502 & 0.374 & 0.383 & 0.259 & 4.339\\
\midrule
\multirow{4}{*}{\textbf{MoCo}} 
 & PFT &\multirow{2}{*}{100\%} & 0.697 & 0.552 & 0.608 & 0.446 & 43.334\\
 & JT &   & 0.682 & 0.560 & 0.611 & 0.447 & 37.642\\
 & PFT & \multirow{2}{*}{10\%} & 0.491 & 0.373 & 0.374 & 0.253 & 9.823\\
 & JT &  & 0.496 & 0.380 & 0.381 & 0.258 & 4.424\\
\midrule
\multirow{4}{*}{\textbf{MAE}} 
 & PFT & \multirow{2}{*}{100\%} & 0.689 & 0.560 & 0.608 & 0.445 & 42.318 \\
 & JT &   & 0.675 & 0.557 & 0.605 & 0.444 & 34.972 \\
 & PFT & \multirow{2}{*}{10\%} & 0.494 & 0.375 & 0.376 & 0.252 & 11.654\\
 & JT & & 0.482 & 0.376 & 0.375 & 0.253 & 9.271\\
\midrule
\multirow{4}{*}{\textbf{DINO}} 
 & PFT & \multirow{2}{*}{100\%} & 0.706 & 0.553 & 0.611 & 0.449 & 54.754\\
 & JT &   & 0.710 & 0.548 & 0.608 & 0.446 & 37.565\\
 & PFT & \multirow{2}{*}{10\%} & 0.509 & 0.367 & 0.377 & 0.255 & 21.330\\
 & JT &   & 0.503 & 0.369 & 0.377 & 0.254 & 4.288\\
\midrule
\multirow{4}{*}{\makecell[l]{\textbf{Barlow}\\\textbf{Twins}}}
 & PFT & \multirow{2}{*}{100\%} & 0.699 & 0.560 & 0.610 & 0.449 & 46.522\\
 & JT &   & 0.692 & 0.556 & 0.611 & 0.450 & 39.006 \\
 & PFT & \multirow{2}{*}{10\%} & 0.517 & 0.363 & 0.381 & 0.259 & 12.992\\
 & JT &   & 0.504 & 0.368 & 0.377 & 0.254 & 4.294\\
\bottomrule
\end{tabular}
}
\label{tab:coco_ssl_results}
\end{table}

Table~\ref{tab:coco_ssl_results} reports the results of the six evaluated SSL methods under fully-labeled and limited-labeled settings for both the PFT and JT training paradigms. Unlike the classification results on datasets like CIFAR-10 and CrisisMMD, the differences in performance between the PFT and JT are much narrower. The best-performing SSL methods under the JT paradigm are BYOL and MoCo, improving on average over PFT under both 100\% and 10\% labeled data settings. The performance is much more varied for the other SSL methods, with PFT oftentimes slightly edging out JT in terms of performance. However, it should be noted that a major benefit of the JT paradigm here is the reduction of training time to achieve comparable or improved results.

\clearpage

\begin{table*}[t]
\centering
\caption{Cross-dataset generalization from COCO to PASCAL VOC2012 under different training regimes and label fractions.}
\vspace{0.5em}
\resizebox{\linewidth}{!}{
\begin{tabular}{l l c c c c c c c c c}
\toprule
\multirow{2}{*}{\textbf{Framework}} & \multirow{2}{*}{\textbf{Config}} & \multirow{2}{*}{\textbf{Labels}} &
\multicolumn{4}{c}{\textbf{COCO}} &
\multicolumn{4}{c}{\textbf{PascalVOC}} \\
\cmidrule(lr){4-7} \cmidrule(lr){8-11}
& & & \textbf{P} & \textbf{R} & \textbf{mAP} & \textbf{mAP50--95} &
\textbf{P} & \textbf{R} & \textbf{mAP} & \textbf{mAP50--95} \\
\midrule

\multirow{4}{*}{\textbf{SimCLR}}
 & PFT & \multirow{2}{*}{100\%} & 0.688 & 0.555 & 0.612 & 0.450 & 0.805 & 0.797 & 0.855 & 0.680 \\
 & JT  &                        & 0.698 & 0.547 & 0.608 & 0.449 & 0.796 & 0.805 & 0.854 & 0.682 \\
 & PFT & \multirow{2}{*}{10\%}  & 0.497 & 0.377 & 0.382 & 0.257 & 0.680 & 0.600 & 0.658 & 0.468 \\
 & JT  &                        & 0.507 & 0.370 & 0.379 & 0.257 & 0.678 & 0.593 & 0.652 & 0.461 \\
\midrule

\multirow{4}{*}{\textbf{BYOL}}
 & PFT & \multirow{2}{*}{100\%} & 0.694 & 0.552 & 0.607 & 0.447 & 0.796 & 0.807 & 0.855 & 0.681 \\
 & JT  &                        & 0.694 & 0.557 & 0.611 & 0.447 & 0.802 & 0.807 & 0.853 & 0.681 \\
 & PFT & \multirow{2}{*}{10\%}  & 0.495 & 0.368 & 0.374 & 0.254 & 0.666 & 0.614 & 0.662 & 0.469 \\
 & JT  &                        & 0.502 & 0.374 & 0.383 & 0.259 & 0.684 & 0.594 & 0.656 & 0.462 \\
\midrule

\multirow{4}{*}{\textbf{MoCo}}
 & PFT & \multirow{2}{*}{100\%} & 0.697 & 0.552 & 0.608 & 0.446 & 0.797 & 0.805 & 0.855 & 0.684 \\
 & JT  &                        & 0.682 & 0.560 & 0.611 & 0.447 & 0.794 & 0.802 & 0.855 & 0.682 \\
 & PFT & \multirow{2}{*}{10\%}  & 0.491 & 0.373 & 0.374 & 0.253 & 0.680 & 0.601 & 0.660 & 0.466 \\
 & JT  &                        & 0.496 & 0.380 & 0.381 & 0.258 & 0.679 & 0.612 & 0.667 & 0.469 \\
\midrule

\multirow{2}{*}{\textbf{MAE}}
 & PFT & \multirow{2}{*}{100\%}  & 0.494 & 0.375 & 0.376 & 0.252 & 0.799 & 0.804 & 0.855 & 0.681 \\
 & JT  &                        & 0.482 & 0.376 & 0.375 & 0.253 & 0.803 & 0.808 & 0.856 & 0.683 \\
 & PFT & \multirow{2}{*}{10\%}  & 0.494 & 0.375 & 0.376 & 0.252 & 0.688 & 0.612 & 0.661 & 0.462 \\
 & JT  &                        & 0.482 & 0.376 & 0.375 & 0.253 & 0.698 & 0.598 & 0.666 & 0.471 \\
\midrule

\multirow{4}{*}{\textbf{DINO}}
 & PFT & \multirow{2}{*}{100\%} & 0.706 & 0.553 & 0.611 & 0.449 & 0.806 & 0.797 & 0.855 & 0.682 \\
 & JT  &                        & 0.710 & 0.548 & 0.608 & 0.446 & 0.790 & 0.807 & 0.854 & 0.679 \\
 & PFT & \multirow{2}{*}{10\%}  & 0.509 & 0.367 & 0.377 & 0.255 & 0.682 & 0.609 & 0.666 & 0.469\\
 & JT  &                        & 0.503 & 0.369 & 0.377 & 0.254 & 0.690 & 0.601 & 0.664 & 0.466 \\
\midrule

\multirow{4}{*}{\makecell[l]{\textbf{Barlow}\\\textbf{Twins}}}
 & PFT & \multirow{2}{*}{100\%} & 0.699 & 0.560 & 0.610 & 0.449 & 0.794 & 0.801 & 0.855 & 0.681 \\
 & JT  &                        & 0.692 & 0.556 & 0.611 & 0.450 & 0.795 & 0.811 & 0.855 & 0.681 \\
 & PFT & \multirow{2}{*}{10\%}  & 0.517 & 0.363 & 0.381 & 0.259 & 0.681 & 0.604 & 0.665 & 0.472 \\
 & JT  &                        & 0.504 & 0.368 & 0.377 & 0.254 & 0.683 & 0.598 & 0.660 & 0.466 \\
\bottomrule
\end{tabular}
}
\label{tab:pascalvoc_results}
\end{table*}

Table~\ref{tab:pascalvoc_results} reports the generalizability of the object detection models, which were trained on COCO, when tested on PASCAL VOC2012. Given that the models were initially trained to predict 80 classes, and given that PASCAL VOC2012 only contains 20 classes, not all of which are identical to their use in COCO, the labels first had to be remapped (e.g., aeroplane $\rightarrow$ airplane) before they could be used by the YOLOv12 model. As shown in the table, in terms of generalization ability, there is little variance in performance between the PFT and JT training paradigms. This demonstrates the efficacy of JT in achieving comparable results with drastically reduced training times. Interestingly, the COCO-trained models tend to perform better on the 20 classes from PASCAL VOC2012, indicating that this dataset may be made up of easier-to-identify classes compared to COCO.

\clearpage

\section{Segmentation Results}

\subsection{Chest X-ray Segmentation}
\label{appendix:jsrt_seg}

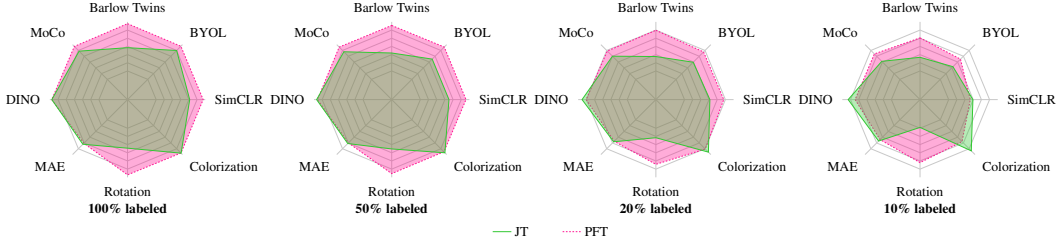
\begin{figure*}[h]
\resizebox{\textwidth}{!}{
\begin{tikzpicture}
 
\definecolor{pftpink}{RGB}{255,20,147}
\definecolor{jtlime}{RGB}{50,200,50}
\definecolor{gridc}{RGB}{200,200,200}
 
\newcommand{\drawpanel}[1]{%
  \foreach \v/\lab in {0.35/0.35, 0.45/0.45, 0.55/0.55, 0.65/0.65, 0.75/0.75, 0.85/0.85}{
    \pgfmathsetmacro{\rr}{\v/0.95*5}
    \draw[gridc,thin]
      (90:\rr cm)--(45:\rr cm)--(0:\rr cm)--(-45:\rr cm)--
      (-90:\rr cm)--(-135:\rr cm)--(180:\rr cm)--(135:\rr cm)--cycle;
  }
  \foreach \i in {0,...,7}{
    \pgfmathsetmacro{\ang}{90-\i*45}
    \draw[gridc,thin] (0,0)--(\ang:5cm);
  }
  \node[font=\Huge,above]       at (90:5.5cm)   {Barlow Twins};
  \node[font=\Huge,above right] at (45:5.5cm)   {BYOL};
  \node[font=\Huge,right]       at (0:5.5cm)    {SimCLR};
  \node[font=\Huge,below right] at (-45:5.5cm)  {Colorization};
  \node[font=\Huge,below]       at (-90:5.5cm)  {Rotation};
  \node[font=\Huge,below left]  at (-135:5.5cm) {MAE};
  \node[font=\Huge,left]        at (180:5.5cm)  {DINO};
  \node[font=\Huge,above left]  at (135:5.5cm)  {MoCo};
}

\begin{scope}[xshift=0cm]
  \drawpanel{}
  \pgfmathsetmacro{\Pax}{0.9285/0.95*5*cos(90)}  \pgfmathsetmacro{\Pay}{0.9285/0.95*5*sin(90)}
  \pgfmathsetmacro{\Pbx}{0.9225/0.95*5*cos(45)}  \pgfmathsetmacro{\Pby}{0.9225/0.95*5*sin(45)}
  \pgfmathsetmacro{\Pcx}{0.9249/0.95*5*cos(0)}   \pgfmathsetmacro{\Pcy}{0.9249/0.95*5*sin(0)}
  \pgfmathsetmacro{\Pdx}{0.9242/0.95*5*cos(-45)} \pgfmathsetmacro{\Pdy}{0.9242/0.95*5*sin(-45)}
  \pgfmathsetmacro{\Pex}{0.9203/0.95*5*cos(-90)} \pgfmathsetmacro{\Pey}{0.9203/0.95*5*sin(-90)}
  \pgfmathsetmacro{\Pfx}{0.7554/0.95*5*cos(-135)}\pgfmathsetmacro{\Pfy}{0.7554/0.95*5*sin(-135)}
  \pgfmathsetmacro{\Pgx}{0.9269/0.95*5*cos(180)} \pgfmathsetmacro{\Pgy}{0.9269/0.95*5*sin(180)}
  \pgfmathsetmacro{\Phx}{0.9192/0.95*5*cos(135)} \pgfmathsetmacro{\Phy}{0.9192/0.95*5*sin(135)}
  \fill[pftpink,opacity=0.35]
    (\Pax cm,\Pay cm)--(\Pbx cm,\Pby cm)--(\Pcx cm,\Pcy cm)--(\Pdx cm,\Pdy cm)--
    (\Pex cm,\Pey cm)--(\Pfx cm,\Pfy cm)--(\Pgx cm,\Pgy cm)--(\Phx cm,\Phy cm)--cycle;
  \draw[pftpink,thick,dashed]
    (\Pax cm,\Pay cm)--(\Pbx cm,\Pby cm)--(\Pcx cm,\Pcy cm)--(\Pdx cm,\Pdy cm)--
    (\Pex cm,\Pey cm)--(\Pfx cm,\Pfy cm)--(\Pgx cm,\Pgy cm)--(\Phx cm,\Phy cm)--cycle;
  \pgfmathsetmacro{\Jax}{0.6348/0.95*5*cos(90)}  \pgfmathsetmacro{\Jay}{0.6348/0.95*5*sin(90)}
  \pgfmathsetmacro{\Jbx}{0.8492/0.95*5*cos(45)}  \pgfmathsetmacro{\Jby}{0.8492/0.95*5*sin(45)}
  \pgfmathsetmacro{\Jcx}{0.7602/0.95*5*cos(0)}   \pgfmathsetmacro{\Jcy}{0.7602/0.95*5*sin(0)}
  \pgfmathsetmacro{\Jdx}{0.9263/0.95*5*cos(-45)} \pgfmathsetmacro{\Jdy}{0.9263/0.95*5*sin(-45)}
  \pgfmathsetmacro{\Jex}{0.594/0.95*5*cos(-90)}  \pgfmathsetmacro{\Jey}{0.594/0.95*5*sin(-90)}
  \pgfmathsetmacro{\Jfx}{0.7723/0.95*5*cos(-135)}\pgfmathsetmacro{\Jfy}{0.7723/0.95*5*sin(-135)}
  \pgfmathsetmacro{\Jgx}{0.9258/0.95*5*cos(180)} \pgfmathsetmacro{\Jgy}{0.9258/0.95*5*sin(180)}
  \pgfmathsetmacro{\Jhx}{0.8377/0.95*5*cos(135)} \pgfmathsetmacro{\Jhy}{0.8377/0.95*5*sin(135)}
  \fill[jtlime,opacity=0.35]
    (\Jax cm,\Jay cm)--(\Jbx cm,\Jby cm)--(\Jcx cm,\Jcy cm)--(\Jdx cm,\Jdy cm)--
    (\Jex cm,\Jey cm)--(\Jfx cm,\Jfy cm)--(\Jgx cm,\Jgy cm)--(\Jhx cm,\Jhy cm)--cycle;
  \draw[jtlime,thick]
    (\Jax cm,\Jay cm)--(\Jbx cm,\Jby cm)--(\Jcx cm,\Jcy cm)--(\Jdx cm,\Jdy cm)--
    (\Jex cm,\Jey cm)--(\Jfx cm,\Jfy cm)--(\Jgx cm,\Jgy cm)--(\Jhx cm,\Jhy cm)--cycle;
  \node[font=\Huge\bfseries] at (0,-7cm) {100\% labeled};
\end{scope}

\begin{scope}[xshift=17cm]
  \drawpanel{}
  \pgfmathsetmacro{\Pax}{0.9066/0.95*5*cos(90)}  \pgfmathsetmacro{\Pay}{0.9066/0.95*5*sin(90)}
  \pgfmathsetmacro{\Pbx}{0.9088/0.95*5*cos(45)}  \pgfmathsetmacro{\Pby}{0.9088/0.95*5*sin(45)}
  \pgfmathsetmacro{\Pcx}{0.9084/0.95*5*cos(0)}   \pgfmathsetmacro{\Pcy}{0.9084/0.95*5*sin(0)}
  \pgfmathsetmacro{\Pdx}{0.9028/0.95*5*cos(-45)} \pgfmathsetmacro{\Pdy}{0.9028/0.95*5*sin(-45)}
  \pgfmathsetmacro{\Pex}{0.9051/0.95*5*cos(-90)} \pgfmathsetmacro{\Pey}{0.9051/0.95*5*sin(-90)}
  \pgfmathsetmacro{\Pfx}{0.749/0.95*5*cos(-135)} \pgfmathsetmacro{\Pfy}{0.749/0.95*5*sin(-135)}
  \pgfmathsetmacro{\Pgx}{0.9082/0.95*5*cos(180)} \pgfmathsetmacro{\Pgy}{0.9082/0.95*5*sin(180)}
  \pgfmathsetmacro{\Phx}{0.908/0.95*5*cos(135)}  \pgfmathsetmacro{\Phy}{0.908/0.95*5*sin(135)}
  \fill[pftpink,opacity=0.35]
    (\Pax cm,\Pay cm)--(\Pbx cm,\Pby cm)--(\Pcx cm,\Pcy cm)--(\Pdx cm,\Pdy cm)--
    (\Pex cm,\Pey cm)--(\Pfx cm,\Pfy cm)--(\Pgx cm,\Pgy cm)--(\Phx cm,\Phy cm)--cycle;
  \draw[pftpink,thick,dashed]
    (\Pax cm,\Pay cm)--(\Pbx cm,\Pby cm)--(\Pcx cm,\Pcy cm)--(\Pdx cm,\Pdy cm)--
    (\Pex cm,\Pey cm)--(\Pfx cm,\Pfy cm)--(\Pgx cm,\Pgy cm)--(\Phx cm,\Phy cm)--cycle;
  \pgfmathsetmacro{\Jax}{0.5697/0.95*5*cos(90)}  \pgfmathsetmacro{\Jay}{0.5697/0.95*5*sin(90)}
  \pgfmathsetmacro{\Jbx}{0.7002/0.95*5*cos(45)}  \pgfmathsetmacro{\Jby}{0.7002/0.95*5*sin(45)}
  \pgfmathsetmacro{\Jcx}{0.7002/0.95*5*cos(0)}   \pgfmathsetmacro{\Jcy}{0.7002/0.95*5*sin(0)}
  \pgfmathsetmacro{\Jdx}{0.92/0.95*5*cos(-45)}   \pgfmathsetmacro{\Jdy}{0.92/0.95*5*sin(-45)}
  \pgfmathsetmacro{\Jex}{0.6034/0.95*5*cos(-90)} \pgfmathsetmacro{\Jey}{0.6034/0.95*5*sin(-90)}
  \pgfmathsetmacro{\Jfx}{0.7597/0.95*5*cos(-135)}\pgfmathsetmacro{\Jfy}{0.7597/0.95*5*sin(-135)}
  \pgfmathsetmacro{\Jgx}{0.9181/0.95*5*cos(180)} \pgfmathsetmacro{\Jgy}{0.9181/0.95*5*sin(180)}
  \pgfmathsetmacro{\Jhx}{0.8272/0.95*5*cos(135)} \pgfmathsetmacro{\Jhy}{0.8272/0.95*5*sin(135)}
  \fill[jtlime,opacity=0.35]
    (\Jax cm,\Jay cm)--(\Jbx cm,\Jby cm)--(\Jcx cm,\Jcy cm)--(\Jdx cm,\Jdy cm)--
    (\Jex cm,\Jey cm)--(\Jfx cm,\Jfy cm)--(\Jgx cm,\Jgy cm)--(\Jhx cm,\Jhy cm)--cycle;
  \draw[jtlime,thick]
    (\Jax cm,\Jay cm)--(\Jbx cm,\Jby cm)--(\Jcx cm,\Jcy cm)--(\Jdx cm,\Jdy cm)--
    (\Jex cm,\Jey cm)--(\Jfx cm,\Jfy cm)--(\Jgx cm,\Jgy cm)--(\Jhx cm,\Jhy cm)--cycle;
  \node[font=\Huge\bfseries] at (0,-7cm) {50\% labeled};
\end{scope}
 
\begin{scope}[xshift=34cm]
  \drawpanel{}
  \pgfmathsetmacro{\Pax}{0.8474/0.95*5*cos(90)}  \pgfmathsetmacro{\Pay}{0.8474/0.95*5*sin(90)}
  \pgfmathsetmacro{\Pbx}{0.8244/0.95*5*cos(45)}  \pgfmathsetmacro{\Pby}{0.8244/0.95*5*sin(45)}
  \pgfmathsetmacro{\Pcx}{0.8299/0.95*5*cos(0)}   \pgfmathsetmacro{\Pcy}{0.8299/0.95*5*sin(0)}
  \pgfmathsetmacro{\Pdx}{0.8546/0.95*5*cos(-45)} \pgfmathsetmacro{\Pdy}{0.8546/0.95*5*sin(-45)}
  \pgfmathsetmacro{\Pex}{0.7915/0.95*5*cos(-90)} \pgfmathsetmacro{\Pey}{0.7915/0.95*5*sin(-90)}
  \pgfmathsetmacro{\Pfx}{0.7206/0.95*5*cos(-135)}\pgfmathsetmacro{\Pfy}{0.7206/0.95*5*sin(-135)}
  \pgfmathsetmacro{\Pgx}{0.8526/0.95*5*cos(180)} \pgfmathsetmacro{\Pgy}{0.8526/0.95*5*sin(180)}
  \pgfmathsetmacro{\Phx}{0.8397/0.95*5*cos(135)} \pgfmathsetmacro{\Phy}{0.8397/0.95*5*sin(135)}
  \fill[pftpink,opacity=0.35]
    (\Pax cm,\Pay cm)--(\Pbx cm,\Pby cm)--(\Pcx cm,\Pcy cm)--(\Pdx cm,\Pdy cm)--
    (\Pex cm,\Pey cm)--(\Pfx cm,\Pfy cm)--(\Pgx cm,\Pgy cm)--(\Phx cm,\Phy cm)--cycle;
  \draw[pftpink,thick,dashed]
    (\Pax cm,\Pay cm)--(\Pbx cm,\Pby cm)--(\Pcx cm,\Pcy cm)--(\Pdx cm,\Pdy cm)--
    (\Pex cm,\Pey cm)--(\Pfx cm,\Pfy cm)--(\Pgx cm,\Pgy cm)--(\Phx cm,\Phy cm)--cycle;
  \pgfmathsetmacro{\Jax}{0.5266/0.95*5*cos(90)}  \pgfmathsetmacro{\Jay}{0.5266/0.95*5*sin(90)}
  \pgfmathsetmacro{\Jbx}{0.646/0.95*5*cos(45)}   \pgfmathsetmacro{\Jby}{0.646/0.95*5*sin(45)}
  \pgfmathsetmacro{\Jcx}{0.6603/0.95*5*cos(0)}   \pgfmathsetmacro{\Jcy}{0.6603/0.95*5*sin(0)}
  \pgfmathsetmacro{\Jdx}{0.9085/0.95*5*cos(-45)} \pgfmathsetmacro{\Jdy}{0.9085/0.95*5*sin(-45)}
  \pgfmathsetmacro{\Jex}{0.4661/0.95*5*cos(-90)} \pgfmathsetmacro{\Jey}{0.4661/0.95*5*sin(-90)}
  \pgfmathsetmacro{\Jfx}{0.7297/0.95*5*cos(-135)}\pgfmathsetmacro{\Jfy}{0.7297/0.95*5*sin(-135)}
  \pgfmathsetmacro{\Jgx}{0.9025/0.95*5*cos(180)} \pgfmathsetmacro{\Jgy}{0.9025/0.95*5*sin(180)}
  \pgfmathsetmacro{\Jhx}{0.7506/0.95*5*cos(135)} \pgfmathsetmacro{\Jhy}{0.7506/0.95*5*sin(135)}
  \fill[jtlime,opacity=0.35]
    (\Jax cm,\Jay cm)--(\Jbx cm,\Jby cm)--(\Jcx cm,\Jcy cm)--(\Jdx cm,\Jdy cm)--
    (\Jex cm,\Jey cm)--(\Jfx cm,\Jfy cm)--(\Jgx cm,\Jgy cm)--(\Jhx cm,\Jhy cm)--cycle;
  \draw[jtlime,thick]
    (\Jax cm,\Jay cm)--(\Jbx cm,\Jby cm)--(\Jcx cm,\Jcy cm)--(\Jdx cm,\Jdy cm)--
    (\Jex cm,\Jey cm)--(\Jfx cm,\Jfy cm)--(\Jgx cm,\Jgy cm)--(\Jhx cm,\Jhy cm)--cycle;
  \node[font=\Huge\bfseries] at (0,-7cm) {20\% labeled};
\end{scope}

\begin{scope}[xshift=51cm]
  \drawpanel{}
  \pgfmathsetmacro{\Pax}{0.755/0.95*5*cos(90)}   \pgfmathsetmacro{\Pay}{0.755/0.95*5*sin(90)}
  \pgfmathsetmacro{\Pbx}{0.6993/0.95*5*cos(45)}  \pgfmathsetmacro{\Pby}{0.6993/0.95*5*sin(45)}
  \pgfmathsetmacro{\Pcx}{0.6144/0.95*5*cos(0)}   \pgfmathsetmacro{\Pcy}{0.6144/0.95*5*sin(0)}
  \pgfmathsetmacro{\Pdx}{0.7261/0.95*5*cos(-45)} \pgfmathsetmacro{\Pdy}{0.7261/0.95*5*sin(-45)}
  \pgfmathsetmacro{\Pex}{0.7707/0.95*5*cos(-90)} \pgfmathsetmacro{\Pey}{0.7707/0.95*5*sin(-90)}
  \pgfmathsetmacro{\Pfx}{0.6913/0.95*5*cos(-135)}\pgfmathsetmacro{\Pfy}{0.6913/0.95*5*sin(-135)}
  \pgfmathsetmacro{\Pgx}{0.7939/0.95*5*cos(180)} \pgfmathsetmacro{\Pgy}{0.7939/0.95*5*sin(180)}
  \pgfmathsetmacro{\Phx}{0.7897/0.95*5*cos(135)} \pgfmathsetmacro{\Phy}{0.7897/0.95*5*sin(135)}
  \fill[pftpink,opacity=0.35]
    (\Pax cm,\Pay cm)--(\Pbx cm,\Pby cm)--(\Pcx cm,\Pcy cm)--(\Pdx cm,\Pdy cm)--
    (\Pex cm,\Pey cm)--(\Pfx cm,\Pfy cm)--(\Pgx cm,\Pgy cm)--(\Phx cm,\Phy cm)--cycle;
  \draw[pftpink,thick,dashed]
    (\Pax cm,\Pay cm)--(\Pbx cm,\Pby cm)--(\Pcx cm,\Pcy cm)--(\Pdx cm,\Pdy cm)--
    (\Pex cm,\Pey cm)--(\Pfx cm,\Pfy cm)--(\Pgx cm,\Pgy cm)--(\Phx cm,\Phy cm)--cycle;
  \pgfmathsetmacro{\Jax}{0.5171/0.95*5*cos(90)}  \pgfmathsetmacro{\Jay}{0.5171/0.95*5*sin(90)}
  \pgfmathsetmacro{\Jbx}{0.5689/0.95*5*cos(45)}  \pgfmathsetmacro{\Jby}{0.5689/0.95*5*sin(45)}
  \pgfmathsetmacro{\Jcx}{0.6464/0.95*5*cos(0)}   \pgfmathsetmacro{\Jcy}{0.6464/0.95*5*sin(0)}
  \pgfmathsetmacro{\Jdx}{0.8862/0.95*5*cos(-45)} \pgfmathsetmacro{\Jdy}{0.8862/0.95*5*sin(-45)}
  \pgfmathsetmacro{\Jex}{0.3401/0.95*5*cos(-90)} \pgfmathsetmacro{\Jey}{0.3401/0.95*5*sin(-90)}
  \pgfmathsetmacro{\Jfx}{0.7164/0.95*5*cos(-135)}\pgfmathsetmacro{\Jfy}{0.7164/0.95*5*sin(-135)}
  \pgfmathsetmacro{\Jgx}{0.8787/0.95*5*cos(180)} \pgfmathsetmacro{\Jgy}{0.8787/0.95*5*sin(180)}
  \pgfmathsetmacro{\Jhx}{0.6616/0.95*5*cos(135)} \pgfmathsetmacro{\Jhy}{0.6616/0.95*5*sin(135)}
  \fill[jtlime,opacity=0.35]
    (\Jax cm,\Jay cm)--(\Jbx cm,\Jby cm)--(\Jcx cm,\Jcy cm)--(\Jdx cm,\Jdy cm)--
    (\Jex cm,\Jey cm)--(\Jfx cm,\Jfy cm)--(\Jgx cm,\Jgy cm)--(\Jhx cm,\Jhy cm)--cycle;
  \draw[jtlime,thick]
    (\Jax cm,\Jay cm)--(\Jbx cm,\Jby cm)--(\Jcx cm,\Jcy cm)--(\Jdx cm,\Jdy cm)--
    (\Jex cm,\Jey cm)--(\Jfx cm,\Jfy cm)--(\Jgx cm,\Jgy cm)--(\Jhx cm,\Jhy cm)--cycle;
  \node[font=\Huge\bfseries] at (0,-7cm) {10\% labeled};
\end{scope}
 
\begin{scope}[xshift=25.5cm, yshift=-8.5cm]
  \draw[jtlime,thick]         (-2cm,0)--(-.8cm,0);
  \node[font=\Huge,anchor=west] at (-.7cm,0) {JT};
  \draw[pftpink,thick,dashed] (2.5cm,0)--(3.7cm,0);
  \node[font=\Huge,anchor=west] at (3.8cm,0) {PFT};
\end{scope}
 
\end{tikzpicture}
}
\caption{Radar plots depicting PFT and JT performance across each of the explored SSL techniques for lung segmentation on the JSRT dataset.}
\label{fig:radar_jsrt_seg}
\end{figure*}

Fig.~\ref{fig:radar_jsrt_seg} shows radar plots of the PFT and JT performance of our eight SSL techniques on the multi-class chest x-ray segmentation task. Unlike the more uniform trends observed on natural image benchmarks, the JSRT results reveal that jointly optimizing supervised and self-supervised objectives can either substantially improve segmentation quality or severely degrade it depending on the SSL framework. These findings reinforce the central claim of this work that the interaction between SSL and supervised learning is highly task- and objective-dependent.

Utilizing colorization as the pretext task for JT demonstrates strong performance, as does the use of DINO. This strong performance is most evident when looking at the performance of JT under the lowest label setting for these two techniques. When trained on 10\% labeled data using colorization as a pretext, JT improves Dice from 0.7261 in the PFT setting to 0.8862 and mIoU from 0.6393 to 0.8103, while also reducing training time. DINO shows a similar trend, with JT consistently outperforming PFT across all label fractions, including a large improvement at 10\% labeled data (Dice: 0.7939 $\rightarrow$ 0.8787).

\begin{figure}[h]
\centering
\resizebox{\linewidth}{!}{
\begin{tabular}{r c c c c c c c c}
& & & & {\large Input Image} & {\large Ground Truth} & & & \\
\smallskip
& & & &
\includegraphics[width=0.22\linewidth]{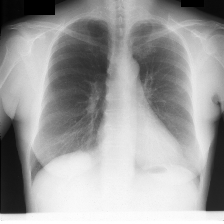} &
\includegraphics[width=0.22\linewidth]{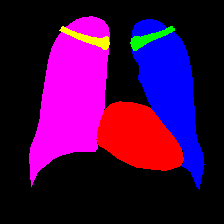} & & & \\
\medskip
\raisebox{-.5\height}{\large Colorization} &
\raisebox{-.5\height}{\includegraphics[width=0.22\linewidth]{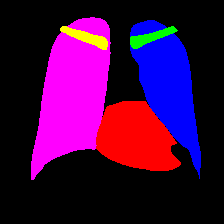}} &
\raisebox{-.5\height}{\includegraphics[width=0.22\linewidth]{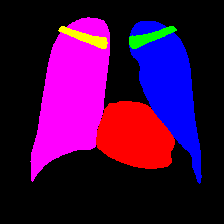}} &
\raisebox{-.5\height}{\includegraphics[width=0.22\linewidth]{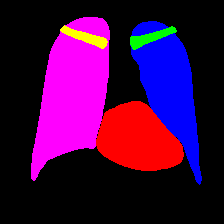}} &
\raisebox{-.5\height}{\includegraphics[width=0.22\linewidth]{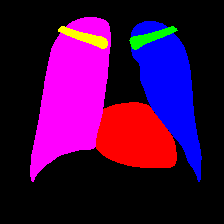}} &
\raisebox{-.5\height}{\includegraphics[width=0.22\linewidth]{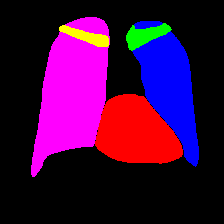}} &
\raisebox{-.5\height}{\includegraphics[width=0.22\linewidth]{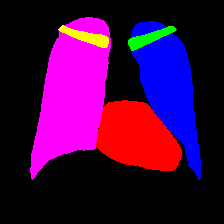}} &
\raisebox{-.5\height}{\includegraphics[width=0.22\linewidth]{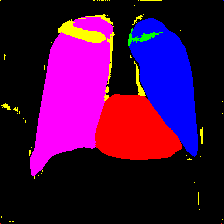}} &
\raisebox{-.5\height}{\includegraphics[width=0.22\linewidth]{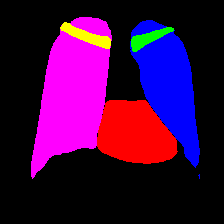}} \\
\medskip
\raisebox{-.5\height}{\large Rotation} &
\raisebox{-.5\height}{\includegraphics[width=0.22\linewidth]{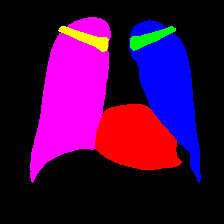}} &
\raisebox{-.5\height}{\includegraphics[width=0.22\linewidth]{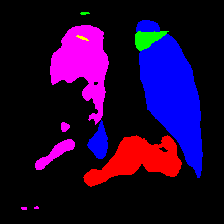}} &
\raisebox{-.5\height}{\includegraphics[width=0.22\linewidth]{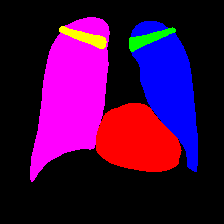}} &
\raisebox{-.5\height}{\includegraphics[width=0.22\linewidth]{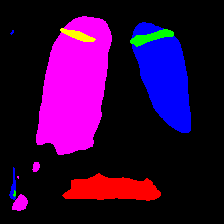}} &
\raisebox{-.5\height}{\includegraphics[width=0.22\linewidth]{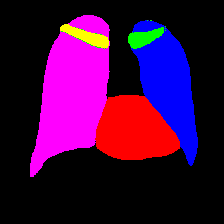}} &
\raisebox{-.5\height}{\includegraphics[width=0.22\linewidth]{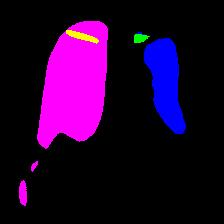}} &
\raisebox{-.5\height}{\includegraphics[width=0.22\linewidth]{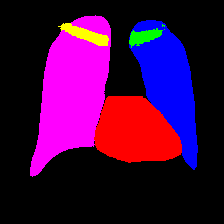}} &
\raisebox{-.5\height}{\includegraphics[width=0.22\linewidth]{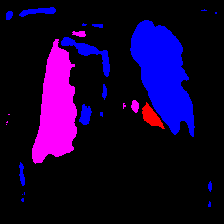}} \\
\medskip
\raisebox{-.5\height}{\large SimCLR} &
\raisebox{-.5\height}{\includegraphics[width=0.22\linewidth]{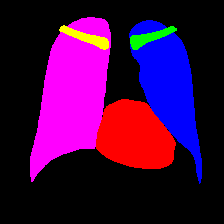}} &
\raisebox{-.5\height}{\includegraphics[width=0.22\linewidth]{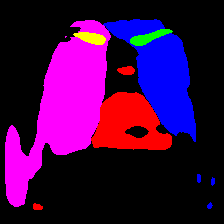}} &
\raisebox{-.5\height}{\includegraphics[width=0.22\linewidth]{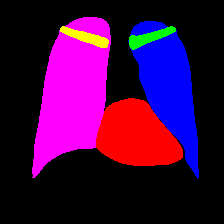}} &
\raisebox{-.5\height}{\includegraphics[width=0.22\linewidth]{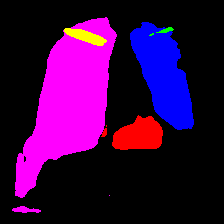}} &
\raisebox{-.5\height}{\includegraphics[width=0.22\linewidth]{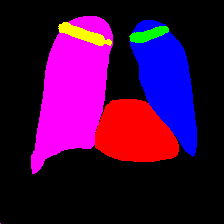}} &
\raisebox{-.5\height}{\includegraphics[width=0.22\linewidth]{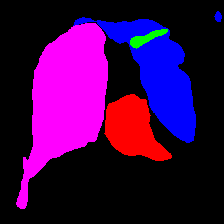}} &
\raisebox{-.5\height}{\includegraphics[width=0.22\linewidth]{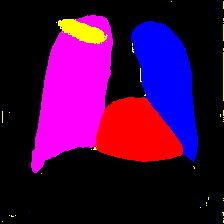}} &
\raisebox{-.5\height}{\includegraphics[width=0.22\linewidth]{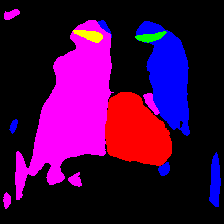}} \\
\medskip
\raisebox{-.5\height}{\large BYOL} &
\raisebox{-.5\height}{\includegraphics[width=0.22\linewidth]{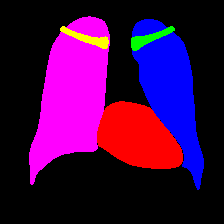}} &
\raisebox{-.5\height}{\includegraphics[width=0.22\linewidth]{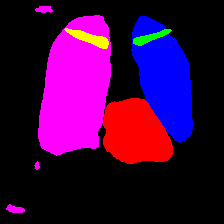}} &
\raisebox{-.5\height}{\includegraphics[width=0.22\linewidth]{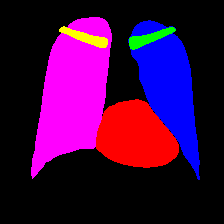}} &
\raisebox{-.5\height}{\includegraphics[width=0.22\linewidth]{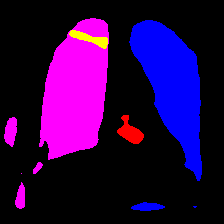}} &
\raisebox{-.5\height}{\includegraphics[width=0.22\linewidth]{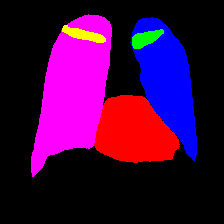}} &
\raisebox{-.5\height}{\includegraphics[width=0.22\linewidth]{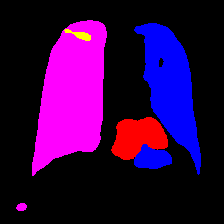}} &
\raisebox{-.5\height}{\includegraphics[width=0.22\linewidth]{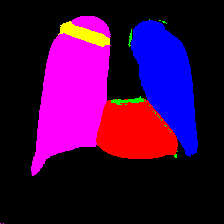}} &
\raisebox{-.5\height}{\includegraphics[width=0.22\linewidth]{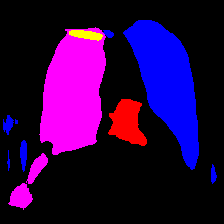}} \\
\medskip
\raisebox{-.5\height}{\large MoCo} &
\raisebox{-.5\height}{\includegraphics[width=0.22\linewidth]{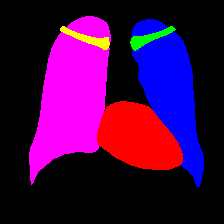}} &
\raisebox{-.5\height}{\includegraphics[width=0.22\linewidth]{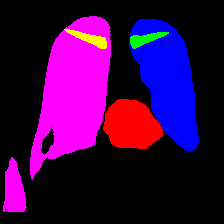}} &
\raisebox{-.5\height}{\includegraphics[width=0.22\linewidth]{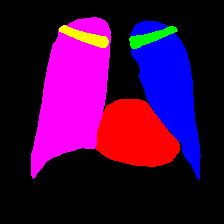}} &
\raisebox{-.5\height}{\includegraphics[width=0.22\linewidth]{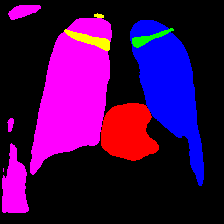}} &
\raisebox{-.5\height}{\includegraphics[width=0.22\linewidth]{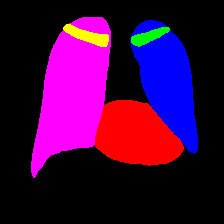}} &
\raisebox{-.5\height}{\includegraphics[width=0.22\linewidth]{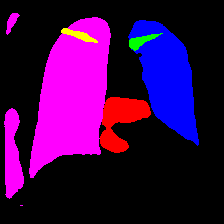}} &
\raisebox{-.5\height}{\includegraphics[width=0.22\linewidth]{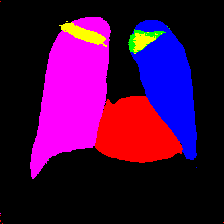}} &
\raisebox{-.5\height}{\includegraphics[width=0.22\linewidth]{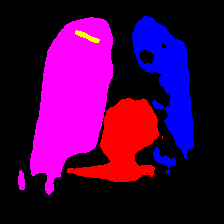}} \\
\medskip
\raisebox{-.5\height}{\large MAE} &
\raisebox{-.5\height}{\includegraphics[width=0.22\linewidth]{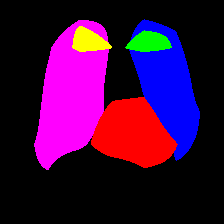}} &
\raisebox{-.5\height}{\includegraphics[width=0.22\linewidth]{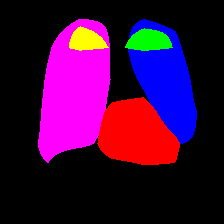}} &
\raisebox{-.5\height}{\includegraphics[width=0.22\linewidth]{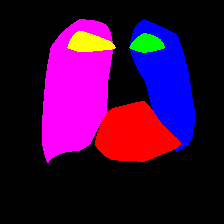}} &
\raisebox{-.5\height}{\includegraphics[width=0.22\linewidth]{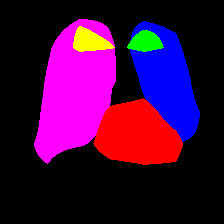}} &
\raisebox{-.5\height}{\includegraphics[width=0.22\linewidth]{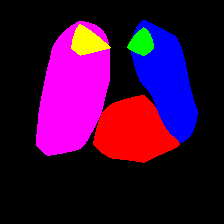}} &
\raisebox{-.5\height}{\includegraphics[width=0.22\linewidth]{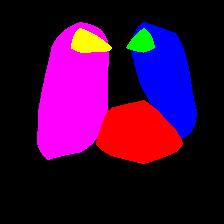}} &
\raisebox{-.5\height}{\includegraphics[width=0.22\linewidth]{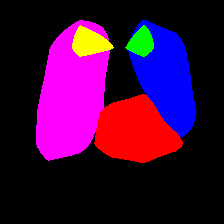}} &
\raisebox{-.5\height}{\includegraphics[width=0.22\linewidth]{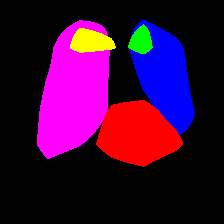}} \\
\medskip
\raisebox{-.5\height}{\large DINO} &
\raisebox{-.5\height}{\includegraphics[width=0.22\linewidth]{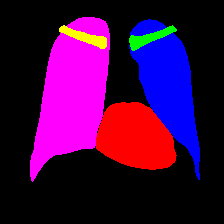}} &
\raisebox{-.5\height}{\includegraphics[width=0.22\linewidth]{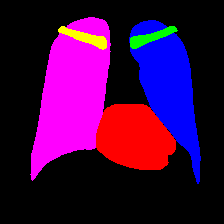}} &
\raisebox{-.5\height}{\includegraphics[width=0.22\linewidth]{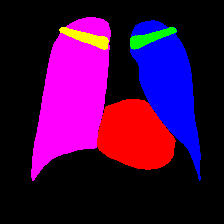}} &
\raisebox{-.5\height}{\includegraphics[width=0.22\linewidth]{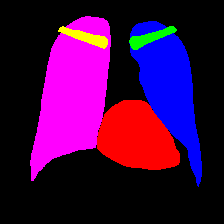}} &
\raisebox{-.5\height}{\includegraphics[width=0.22\linewidth]{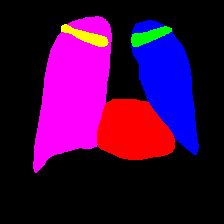}} &
\raisebox{-.5\height}{\includegraphics[width=0.22\linewidth]{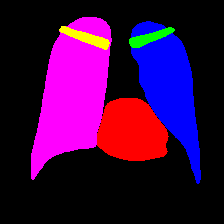}} &
\raisebox{-.5\height}{\includegraphics[width=0.22\linewidth]{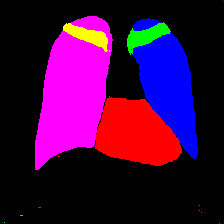}} &
\raisebox{-.5\height}{\includegraphics[width=0.22\linewidth]{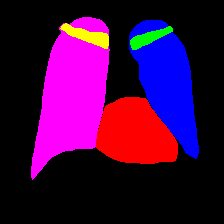}} \\
\medskip
\raisebox{-.5\height}{\large Barlow Twins} &
\raisebox{-.5\height}{\includegraphics[width=0.22\linewidth]{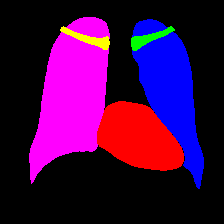}} &
\raisebox{-.5\height}{\includegraphics[width=0.22\linewidth]{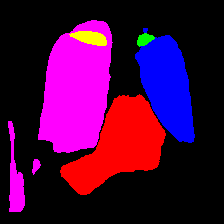}} &
\raisebox{-.5\height}{\includegraphics[width=0.22\linewidth]{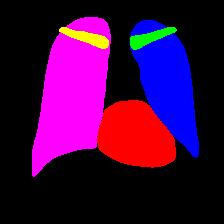}} &
\raisebox{-.5\height}{\includegraphics[width=0.22\linewidth]{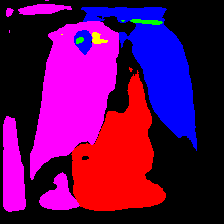}} &
\raisebox{-.5\height}{\includegraphics[width=0.22\linewidth]{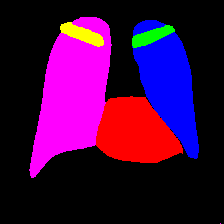}} &
\raisebox{-.5\height}{\includegraphics[width=0.22\linewidth]{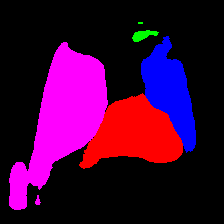}} &
\raisebox{-.5\height}{\includegraphics[width=0.22\linewidth]{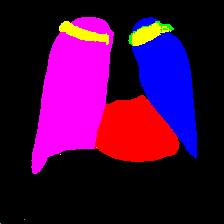}} &
\raisebox{-.5\height}{\includegraphics[width=0.22\linewidth]{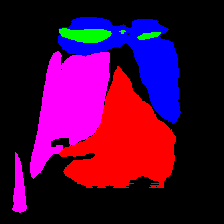}} \\
& & {\large 100\%} & & {\large 50\%} & & {\large 20\%} & & {\large 10\%} \\
\end{tabular}
}
\caption{Predicted segmentation masks vs. the ground truth masks on the JSRT dataset across each of the examined SSL techniques and labeled fractions. Color code: red (heart), pink (left lung), blue (right lung), yellow (left clavicle), green (right clavicle).}
\label{fig:seg_results_jsrt}
\end{figure}

In contrast, several SSL objectives exhibit strong optimization interference under JT. Rotation prediction represents the clearest failure case, where JT causes a severe collapse in segmentation performance despite strong PFT results. At 10\% labels, Dice drops from 0.7707 under PFT to 0.3401 under JT. Similar degradation is observed for SimCLR, BYOL, MoCo, and Barlow Twins, where JT consistently underperforms PFT across most label fractions. A detailed numerical breakdown of segmentation performance across the examined techniques can be seen in Table~\ref{tab:jsrt_seg}. Fig.~\ref{fig:seg_results_jsrt} visualizes the predicted segmentation masks across the five classes of the JSRT dataset (heart, left and right clavicles, left and right lungs) for each of the SSL techniques and labeled fractions.

\clearpage

\begin{table}
    \centering
    \caption{Segmentation performance on JSRT under different training schemes.}
    \resizebox{0.4\textheight}{!}{
    \begin{tabular}{llcccc}
        \toprule
        \textbf{Framework} & \textbf{Config} & \textbf{Label (\%)} & \textbf{Dice} & \textbf{mIoU} & \textbf{Training Time (s)}\\
        \midrule
        \multirow{8}{*}{\textbf{Colorization}} 
        & PFT & \multirow{2}{*}{100\%} & 0.9242 & 0.8642 & 274.9857 \\
        & JT & & 0.9263 & 0.8678 & 255.3800\\
        & PFT & \multirow{2}{*}{50\%} & 0.9028 & 0.8368 & 248.2321\\
        & JT & & 0.9200 & 0.8582 & 224.3300 \\
        & PFT & \multirow{2}{*}{20\%} & 0.8546 & 0.7698 & 257.0643\\
        & JT & & 0.9085 & 0.8414 & 212.6400 \\
        & PFT & \multirow{2}{*}{10\%} & 0.7261 & 0.6393 & 246.2301\\
        & JT & & 0.8862 & 0.8103 & 200.1700 \\
        \midrule
        \multirow{8}{*}{\textbf{Rotation}} 
        & PFT & \multirow{2}{*}{100\%} & 0.9203 & 0.8580 & 261.7501 \\
        & JT & & 0.5940 & 0.4757 & 297.5800 \\
        & PFT & \multirow{2}{*}{50\%} & 0.9051 & 0.8355 & 246.8294 \\
        & JT & & 0.6034 & 0.5051 & 282.1100 \\
        & PFT & \multirow{2}{*}{20\%} & 0.7915 & 0.7031 & 237.6206 \\
        & JT & & 0.4661 & 0.3682 & 181.5900 \\
        & PFT & \multirow{2}{*}{10\%} & 0.7707 & 0.6791 & 224.3539 \\
        & JT & & 0.3401 & 0.2714 & 166.0300 \\
        \midrule
        \multirow{8}{*}{\textbf{SimCLR}} 
        & PFT & \multirow{2}{*}{100\%} & 0.9249 & 0.8655 & 331.3500 \\
        & JT & & 0.7602 & 0.6484 & 453.4335 \\
        & PFT & \multirow{2}{*}{50\%} & 0.9084 & 0.8409 & 299.2333 \\
        & JT & & 0.7002 & 0.5905 & 308.6110 \\
        & PFT & \multirow{2}{*}{20\%} & 0.8299 & 0.7445 & 286.9790 \\
        & JT & & 0.6603 & 0.5635 & 233.9699 \\
        & PFT & \multirow{2}{*}{10\%} & 0.6144 & 0.5641 & 276.9653 \\
        & JT & & 0.6464 & 0.5206 & 199.2107 \\
        \midrule
        \multirow{8}{*}{\textbf{BYOL}} 
        & PFT & \multirow{2}{*}{100\%} & 0.9225 & 0.8628 & 349.5800 \\
        & JT & & 0.8492 & 0.7566 & 408.2511 \\
        & PFT & \multirow{2}{*}{50\%} & 0.9088 & 0.7002 & 325.1443 \\
        & JT & & 0.7002 & 0.5905 & 308.3321 \\
        & PFT & \multirow{2}{*}{20\%} & 0.8244 & 0.7375 & 300.1652 \\
        & JT & & 0.6460 & 0.5425 & 253.8184 \\
        & PFT & \multirow{2}{*}{10\%} & 0.6993 & 0.6300 & 301.4861 \\
        & JT & & 0.5689 & 0.4793 & 215.6505 \\
        \midrule
        \multirow{8}{*}{\textbf{MoCo}} 
        & PFT & \multirow{2}{*}{100\%} & 0.9192 & 0.8578 & 315.8700 \\
        & JT & & 0.8377 & 0.7405 & 420.9029 \\
        & PFT & \multirow{2}{*}{50\%} & 0.9080 & 0.8397 & 284.9684 \\
        & JT & & 0.8272 & 0.7273 & 315.4798 \\
        & PFT & \multirow{2}{*}{20\%} & 0.8397 & 0.7581 & 284.8934 \\
        & JT & & 0.7506 & 0.6446 & 254.6207 \\
        & PFT & \multirow{2}{*}{10\%} & 0.7897 & 0.6964 & 272.5408 \\
        & JT & & 0.6616 & 0.5615 & 224.2699 \\
        \midrule
        \multirow{8}{*}{\textbf{MAE}} 
        & PFT & \multirow{2}{*}{100\%} & 0.7554 & 0.6498 & 532.2200 \\
        & JT & & 0.7723 & 0.6689 & 447.9000 \\
        & PFT & \multirow{2}{*}{50\%} & 0.7490 & 0.6498 & 505.6600 \\
        & JT & & 0.7597 & 0.6526 & 971.6223 \\
        & PFT & \multirow{2}{*}{20\%} & 0.7206 & 0.6087 & 481.4300 \\
        & JT & & 0.7297 & 0.6157 & 317.9478 \\
        & PFT & \multirow{2}{*}{10\%} & 0.6913 & 0.5776 & 569.1400 \\
        & JT & & 0.7164 & 0.5998 & 277.3327 \\
        \midrule
        \multirow{8}{*}{\textbf{DINO}} 
        & PFT & \multirow{2}{*}{100\%} & 0.9269 & 0.8685 & 438.8100 \\
        & JT & & 0.9258 & 0.8669 & 718.7600 \\
        & PFT & \multirow{2}{*}{50\%} & 0.9082 & 0.8410 & 410.7700 \\
        & JT & & 0.9181 & 0.8552 & 538.8200 \\
        & PFT & \multirow{2}{*}{20\%} & 0.8526 & 0.7690 & 389.6300 \\
        & JT & & 0.9025 & 0.8318 & 537.6600 \\
        & PFT & \multirow{2}{*}{10\%} & 0.7939 & 0.7030 & 382.2300 \\
        & JT & & 0.8787 & 0.7997 & 506.2700 \\
        \midrule
        \multirow{8}{*}{\textbf{Barlow Twins}} 
        & PFT & \multirow{2}{*}{100\%} & 0.9285 & 0.8720 & 324.5800 \\
        & JT & & 0.6348 & 0.5184 & 389.1663 \\
        & PFT & \multirow{2}{*}{50\%} & 0.9066 & 0.8387 & 308.4739 \\
        & JT & & 0.5697 & 0.4423 & 281.9649 \\
        & PFT & \multirow{2}{*}{20\%} & 0.8474 & 0.7652 & 291.6109 \\
        & JT & & 0.5266 & 0.4340 & 227.8010 \\
        & PFT & \multirow{2}{*}{10\%} & 0.7550 & 0.6662 & 287.0245 \\
        & JT & & 0.5171 & 0.4062 & 205.1080 \\
        \bottomrule
    \end{tabular}
    }
    \label{tab:jsrt_seg}
\end{table}

\clearpage

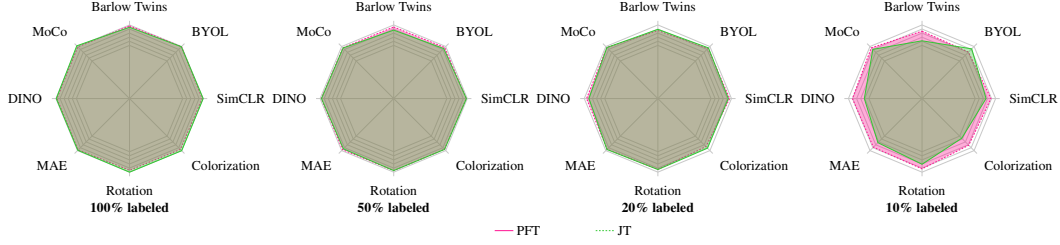
\begin{figure}
\resizebox{\textwidth}{!}{
\begin{tikzpicture}
 
\definecolor{pftpink}{RGB}{255,20,147}
\definecolor{jtlime}{RGB}{50,200,50}
\definecolor{gridc}{RGB}{200,200,200}

\newcommand{\drawpanel}[1]{%
  \foreach \v/\lab in {0.65/0.65, 0.70/0.70, 0.75/0.75, 0.80/0.80, 0.85/0.85, 0.90/0.90}{
    \pgfmathsetmacro{\rr}{\v/0.95*5}
    \draw[gridc,thin]
      (90:\rr cm)--(45:\rr cm)--(0:\rr cm)--(-45:\rr cm)--
      (-90:\rr cm)--(-135:\rr cm)--(180:\rr cm)--(135:\rr cm)--cycle;
  }
  \foreach \i in {0,...,7}{
    \pgfmathsetmacro{\ang}{90-\i*45}
    \draw[gridc,thin] (0,0)--(\ang:5cm);
  }
  \node[font=\Huge,above]       at (90:5.5cm)   {Barlow Twins};
  \node[font=\Huge,above right] at (45:5.5cm)   {BYOL};
  \node[font=\Huge,right]       at (0:5.5cm)    {SimCLR};
  \node[font=\Huge,below right] at (-45:5.5cm)  {Colorization};
  \node[font=\Huge,below]       at (-90:5.5cm)  {Rotation};
  \node[font=\Huge,below left]  at (-135:5.5cm) {MAE};
  \node[font=\Huge,left]        at (180:5.5cm)  {DINO};
  \node[font=\Huge,above left]  at (135:5.5cm)  {MoCo};
}
 
\begin{scope}[xshift=0cm]
  \drawpanel{}
  \pgfmathsetmacro{\Pax}{0.8886/0.95*5*cos(90)}  \pgfmathsetmacro{\Pay}{0.8886/0.95*5*sin(90)}
  \pgfmathsetmacro{\Pbx}{0.8978/0.95*5*cos(45)}  \pgfmathsetmacro{\Pby}{0.8978/0.95*5*sin(45)}
  \pgfmathsetmacro{\Pcx}{0.8886/0.95*5*cos(0)}   \pgfmathsetmacro{\Pcy}{0.8886/0.95*5*sin(0)}
  \pgfmathsetmacro{\Pdx}{0.8798/0.95*5*cos(-45)} \pgfmathsetmacro{\Pdy}{0.8798/0.95*5*sin(-45)}
  \pgfmathsetmacro{\Pex}{0.8776/0.95*5*cos(-90)} \pgfmathsetmacro{\Pey}{0.8776/0.95*5*sin(-90)}
  \pgfmathsetmacro{\Pfx}{0.8883/0.95*5*cos(-135)}\pgfmathsetmacro{\Pfy}{0.8883/0.95*5*sin(-135)}
  \pgfmathsetmacro{\Pgx}{0.888/0.95*5*cos(180)}  \pgfmathsetmacro{\Pgy}{0.888/0.95*5*sin(180)}
  \pgfmathsetmacro{\Phx}{0.8946/0.95*5*cos(135)} \pgfmathsetmacro{\Phy}{0.8946/0.95*5*sin(135)}
  \fill[pftpink,opacity=0.35]
    (\Pax cm,\Pay cm)--(\Pbx cm,\Pby cm)--(\Pcx cm,\Pcy cm)--(\Pdx cm,\Pdy cm)--
    (\Pex cm,\Pey cm)--(\Pfx cm,\Pfy cm)--(\Pgx cm,\Pgy cm)--(\Phx cm,\Phy cm)--cycle;
  \draw[pftpink,thick,dashed]
    (\Pax cm,\Pay cm)--(\Pbx cm,\Pby cm)--(\Pcx cm,\Pcy cm)--(\Pdx cm,\Pdy cm)--
    (\Pex cm,\Pey cm)--(\Pfx cm,\Pfy cm)--(\Pgx cm,\Pgy cm)--(\Phx cm,\Phy cm)--cycle;
  \pgfmathsetmacro{\Jax}{0.8718/0.95*5*cos(90)}  \pgfmathsetmacro{\Jay}{0.8718/0.95*5*sin(90)}
  \pgfmathsetmacro{\Jbx}{0.8965/0.95*5*cos(45)}  \pgfmathsetmacro{\Jby}{0.8965/0.95*5*sin(45)}
  \pgfmathsetmacro{\Jcx}{0.9005/0.95*5*cos(0)}   \pgfmathsetmacro{\Jcy}{0.9005/0.95*5*sin(0)}
  \pgfmathsetmacro{\Jdx}{0.9019/0.95*5*cos(-45)} \pgfmathsetmacro{\Jdy}{0.9019/0.95*5*sin(-45)}
  \pgfmathsetmacro{\Jex}{0.9002/0.95*5*cos(-90)} \pgfmathsetmacro{\Jey}{0.9002/0.95*5*sin(-90)}
  \pgfmathsetmacro{\Jfx}{0.8948/0.95*5*cos(-135)}\pgfmathsetmacro{\Jfy}{0.8948/0.95*5*sin(-135)}
  \pgfmathsetmacro{\Jgx}{0.8959/0.95*5*cos(180)} \pgfmathsetmacro{\Jgy}{0.8959/0.95*5*sin(180)}
  \pgfmathsetmacro{\Jhx}{0.9123/0.95*5*cos(135)} \pgfmathsetmacro{\Jhy}{0.9123/0.95*5*sin(135)}
  \fill[jtlime,opacity=0.35]
    (\Jax cm,\Jay cm)--(\Jbx cm,\Jby cm)--(\Jcx cm,\Jcy cm)--(\Jdx cm,\Jdy cm)--
    (\Jex cm,\Jey cm)--(\Jfx cm,\Jfy cm)--(\Jgx cm,\Jgy cm)--(\Jhx cm,\Jhy cm)--cycle;
  \draw[jtlime,thick]
    (\Jax cm,\Jay cm)--(\Jbx cm,\Jby cm)--(\Jcx cm,\Jcy cm)--(\Jdx cm,\Jdy cm)--
    (\Jex cm,\Jey cm)--(\Jfx cm,\Jfy cm)--(\Jgx cm,\Jgy cm)--(\Jhx cm,\Jhy cm)--cycle;
  \node[font=\Huge\bfseries] at (0,-7cm) {100\% labeled};
\end{scope}

\begin{scope}[xshift=17cm]
  \drawpanel{}
  \pgfmathsetmacro{\Pax}{0.8752/0.95*5*cos(90)}  \pgfmathsetmacro{\Pay}{0.8752/0.95*5*sin(90)}
  \pgfmathsetmacro{\Pbx}{0.8827/0.95*5*cos(45)}  \pgfmathsetmacro{\Pby}{0.8827/0.95*5*sin(45)}
  \pgfmathsetmacro{\Pcx}{0.8867/0.95*5*cos(0)}   \pgfmathsetmacro{\Pcy}{0.8867/0.95*5*sin(0)}
  \pgfmathsetmacro{\Pdx}{0.8772/0.95*5*cos(-45)} \pgfmathsetmacro{\Pdy}{0.8772/0.95*5*sin(-45)}
  \pgfmathsetmacro{\Pex}{0.8809/0.95*5*cos(-90)} \pgfmathsetmacro{\Pey}{0.8809/0.95*5*sin(-90)}
  \pgfmathsetmacro{\Pfx}{0.8839/0.95*5*cos(-135)}\pgfmathsetmacro{\Pfy}{0.8839/0.95*5*sin(-135)}
  \pgfmathsetmacro{\Pgx}{0.8853/0.95*5*cos(180)} \pgfmathsetmacro{\Pgy}{0.8853/0.95*5*sin(180)}
  \pgfmathsetmacro{\Phx}{0.8784/0.95*5*cos(135)} \pgfmathsetmacro{\Phy}{0.8784/0.95*5*sin(135)}
  \fill[pftpink,opacity=0.35]
    (\Pax cm,\Pay cm)--(\Pbx cm,\Pby cm)--(\Pcx cm,\Pcy cm)--(\Pdx cm,\Pdy cm)--
    (\Pex cm,\Pey cm)--(\Pfx cm,\Pfy cm)--(\Pgx cm,\Pgy cm)--(\Phx cm,\Phy cm)--cycle;
  \draw[pftpink,thick,dashed]
    (\Pax cm,\Pay cm)--(\Pbx cm,\Pby cm)--(\Pcx cm,\Pcy cm)--(\Pdx cm,\Pdy cm)--
    (\Pex cm,\Pey cm)--(\Pfx cm,\Pfy cm)--(\Pgx cm,\Pgy cm)--(\Phx cm,\Phy cm)--cycle;
  \pgfmathsetmacro{\Jax}{0.837/0.95*5*cos(90)}   \pgfmathsetmacro{\Jay}{0.837/0.95*5*sin(90)}
  \pgfmathsetmacro{\Jbx}{0.865/0.95*5*cos(45)}   \pgfmathsetmacro{\Jby}{0.865/0.95*5*sin(45)}
  \pgfmathsetmacro{\Jcx}{0.8893/0.95*5*cos(0)}   \pgfmathsetmacro{\Jcy}{0.8893/0.95*5*sin(0)}
  \pgfmathsetmacro{\Jdx}{0.8794/0.95*5*cos(-45)} \pgfmathsetmacro{\Jdy}{0.8794/0.95*5*sin(-45)}
  \pgfmathsetmacro{\Jex}{0.8806/0.95*5*cos(-90)} \pgfmathsetmacro{\Jey}{0.8806/0.95*5*sin(-90)}
  \pgfmathsetmacro{\Jfx}{0.8698/0.95*5*cos(-135)}\pgfmathsetmacro{\Jfy}{0.8698/0.95*5*sin(-135)}
  \pgfmathsetmacro{\Jgx}{0.8834/0.95*5*cos(180)} \pgfmathsetmacro{\Jgy}{0.8834/0.95*5*sin(180)}
  \pgfmathsetmacro{\Jhx}{0.8751/0.95*5*cos(135)} \pgfmathsetmacro{\Jhy}{0.8751/0.95*5*sin(135)}
  \fill[jtlime,opacity=0.35]
    (\Jax cm,\Jay cm)--(\Jbx cm,\Jby cm)--(\Jcx cm,\Jcy cm)--(\Jdx cm,\Jdy cm)--
    (\Jex cm,\Jey cm)--(\Jfx cm,\Jfy cm)--(\Jgx cm,\Jgy cm)--(\Jhx cm,\Jhy cm)--cycle;
  \draw[jtlime,thick]
    (\Jax cm,\Jay cm)--(\Jbx cm,\Jby cm)--(\Jcx cm,\Jcy cm)--(\Jdx cm,\Jdy cm)--
    (\Jex cm,\Jey cm)--(\Jfx cm,\Jfy cm)--(\Jgx cm,\Jgy cm)--(\Jhx cm,\Jhy cm)--cycle;
  \node[font=\Huge\bfseries] at (0,-7cm) {50\% labeled};
\end{scope}
 
\begin{scope}[xshift=34cm]
  \drawpanel{}
  \pgfmathsetmacro{\Pax}{0.8319/0.95*5*cos(90)}  \pgfmathsetmacro{\Pay}{0.8319/0.95*5*sin(90)}
  \pgfmathsetmacro{\Pbx}{0.8666/0.95*5*cos(45)}  \pgfmathsetmacro{\Pby}{0.8666/0.95*5*sin(45)}
  \pgfmathsetmacro{\Pcx}{0.8779/0.95*5*cos(0)}   \pgfmathsetmacro{\Pcy}{0.8779/0.95*5*sin(0)}
  \pgfmathsetmacro{\Pdx}{0.8313/0.95*5*cos(-45)} \pgfmathsetmacro{\Pdy}{0.8313/0.95*5*sin(-45)}
  \pgfmathsetmacro{\Pex}{0.8666/0.95*5*cos(-90)} \pgfmathsetmacro{\Pey}{0.8666/0.95*5*sin(-90)}
  \pgfmathsetmacro{\Pfx}{0.8763/0.95*5*cos(-135)}\pgfmathsetmacro{\Pfy}{0.8763/0.95*5*sin(-135)}
  \pgfmathsetmacro{\Pgx}{0.8722/0.95*5*cos(180)} \pgfmathsetmacro{\Pgy}{0.8722/0.95*5*sin(180)}
  \pgfmathsetmacro{\Phx}{0.8824/0.95*5*cos(135)} \pgfmathsetmacro{\Phy}{0.8824/0.95*5*sin(135)}
  \fill[pftpink,opacity=0.35]
    (\Pax cm,\Pay cm)--(\Pbx cm,\Pby cm)--(\Pcx cm,\Pcy cm)--(\Pdx cm,\Pdy cm)--
    (\Pex cm,\Pey cm)--(\Pfx cm,\Pfy cm)--(\Pgx cm,\Pgy cm)--(\Phx cm,\Phy cm)--cycle;
  \draw[pftpink,thick,dashed]
    (\Pax cm,\Pay cm)--(\Pbx cm,\Pby cm)--(\Pcx cm,\Pcy cm)--(\Pdx cm,\Pdy cm)--
    (\Pex cm,\Pey cm)--(\Pfx cm,\Pfy cm)--(\Pgx cm,\Pgy cm)--(\Phx cm,\Phy cm)--cycle;
  \pgfmathsetmacro{\Jax}{0.8461/0.95*5*cos(90)}  \pgfmathsetmacro{\Jay}{0.8461/0.95*5*sin(90)}
  \pgfmathsetmacro{\Jbx}{0.8756/0.95*5*cos(45)}  \pgfmathsetmacro{\Jby}{0.8756/0.95*5*sin(45)}
  \pgfmathsetmacro{\Jcx}{0.8612/0.95*5*cos(0)}   \pgfmathsetmacro{\Jcy}{0.8612/0.95*5*sin(0)}
  \pgfmathsetmacro{\Jdx}{0.8584/0.95*5*cos(-45)} \pgfmathsetmacro{\Jdy}{0.8584/0.95*5*sin(-45)}
  \pgfmathsetmacro{\Jex}{0.868/0.95*5*cos(-90)}  \pgfmathsetmacro{\Jey}{0.868/0.95*5*sin(-90)}
  \pgfmathsetmacro{\Jfx}{0.8811/0.95*5*cos(-135)}\pgfmathsetmacro{\Jfy}{0.8811/0.95*5*sin(-135)}
  \pgfmathsetmacro{\Jgx}{0.8427/0.95*5*cos(180)} \pgfmathsetmacro{\Jgy}{0.8427/0.95*5*sin(180)}
  \pgfmathsetmacro{\Jhx}{0.8826/0.95*5*cos(135)} \pgfmathsetmacro{\Jhy}{0.8826/0.95*5*sin(135)}
  \fill[jtlime,opacity=0.35]
    (\Jax cm,\Jay cm)--(\Jbx cm,\Jby cm)--(\Jcx cm,\Jcy cm)--(\Jdx cm,\Jdy cm)--
    (\Jex cm,\Jey cm)--(\Jfx cm,\Jfy cm)--(\Jgx cm,\Jgy cm)--(\Jhx cm,\Jhy cm)--cycle;
  \draw[jtlime,thick]
    (\Jax cm,\Jay cm)--(\Jbx cm,\Jby cm)--(\Jcx cm,\Jcy cm)--(\Jdx cm,\Jdy cm)--
    (\Jex cm,\Jey cm)--(\Jfx cm,\Jfy cm)--(\Jgx cm,\Jgy cm)--(\Jhx cm,\Jhy cm)--cycle;
  \node[font=\Huge\bfseries] at (0,-7cm) {20\% labeled};
\end{scope}
 
\begin{scope}[xshift=51cm]
  \drawpanel{}
  \pgfmathsetmacro{\Pax}{0.8244/0.95*5*cos(90)}  \pgfmathsetmacro{\Pay}{0.8244/0.95*5*sin(90)}
  \pgfmathsetmacro{\Pbx}{0.8034/0.95*5*cos(45)}  \pgfmathsetmacro{\Pby}{0.8034/0.95*5*sin(45)}
  \pgfmathsetmacro{\Pcx}{0.8428/0.95*5*cos(0)}   \pgfmathsetmacro{\Pcy}{0.8428/0.95*5*sin(0)}
  \pgfmathsetmacro{\Pdx}{0.8076/0.95*5*cos(-45)} \pgfmathsetmacro{\Pdy}{0.8076/0.95*5*sin(-45)}
  \pgfmathsetmacro{\Pex}{0.8598/0.95*5*cos(-90)} \pgfmathsetmacro{\Pey}{0.8598/0.95*5*sin(-90)}
  \pgfmathsetmacro{\Pfx}{0.8411/0.95*5*cos(-135)}\pgfmathsetmacro{\Pfy}{0.8411/0.95*5*sin(-135)}
  \pgfmathsetmacro{\Pgx}{0.8517/0.95*5*cos(180)} \pgfmathsetmacro{\Pgy}{0.8517/0.95*5*sin(180)}
  \pgfmathsetmacro{\Phx}{0.8777/0.95*5*cos(135)} \pgfmathsetmacro{\Phy}{0.8777/0.95*5*sin(135)}
  \fill[pftpink,opacity=0.35]
    (\Pax cm,\Pay cm)--(\Pbx cm,\Pby cm)--(\Pcx cm,\Pcy cm)--(\Pdx cm,\Pdy cm)--
    (\Pex cm,\Pey cm)--(\Pfx cm,\Pfy cm)--(\Pgx cm,\Pgy cm)--(\Phx cm,\Phy cm)--cycle;
  \draw[pftpink,thick,dashed]
    (\Pax cm,\Pay cm)--(\Pbx cm,\Pby cm)--(\Pcx cm,\Pcy cm)--(\Pdx cm,\Pdy cm)--
    (\Pex cm,\Pey cm)--(\Pfx cm,\Pfy cm)--(\Pgx cm,\Pgy cm)--(\Phx cm,\Phy cm)--cycle;
  \pgfmathsetmacro{\Jax}{0.7062/0.95*5*cos(90)}  \pgfmathsetmacro{\Jay}{0.7062/0.95*5*sin(90)}
  \pgfmathsetmacro{\Jbx}{0.8594/0.95*5*cos(45)}  \pgfmathsetmacro{\Jby}{0.8594/0.95*5*sin(45)}
  \pgfmathsetmacro{\Jcx}{0.7828/0.95*5*cos(0)}   \pgfmathsetmacro{\Jcy}{0.7828/0.95*5*sin(0)}
  \pgfmathsetmacro{\Jdx}{0.6865/0.95*5*cos(-45)} \pgfmathsetmacro{\Jdy}{0.6865/0.95*5*sin(-45)}
  \pgfmathsetmacro{\Jex}{0.8035/0.95*5*cos(-90)} \pgfmathsetmacro{\Jey}{0.8035/0.95*5*sin(-90)}
  \pgfmathsetmacro{\Jfx}{0.7631/0.95*5*cos(-135)}\pgfmathsetmacro{\Jfy}{0.7631/0.95*5*sin(-135)}
  \pgfmathsetmacro{\Jgx}{0.7058/0.95*5*cos(180)} \pgfmathsetmacro{\Jgy}{0.7058/0.95*5*sin(180)}
  \pgfmathsetmacro{\Jhx}{0.8506/0.95*5*cos(135)} \pgfmathsetmacro{\Jhy}{0.8506/0.95*5*sin(135)}
  \fill[jtlime,opacity=0.35]
    (\Jax cm,\Jay cm)--(\Jbx cm,\Jby cm)--(\Jcx cm,\Jcy cm)--(\Jdx cm,\Jdy cm)--
    (\Jex cm,\Jey cm)--(\Jfx cm,\Jfy cm)--(\Jgx cm,\Jgy cm)--(\Jhx cm,\Jhy cm)--cycle;
  \draw[jtlime,thick]
    (\Jax cm,\Jay cm)--(\Jbx cm,\Jby cm)--(\Jcx cm,\Jcy cm)--(\Jdx cm,\Jdy cm)--
    (\Jex cm,\Jey cm)--(\Jfx cm,\Jfy cm)--(\Jgx cm,\Jgy cm)--(\Jhx cm,\Jhy cm)--cycle;
  \node[font=\Huge\bfseries] at (0,-7cm) {10\% labeled};
\end{scope}
 
\begin{scope}[xshift=25.5cm, yshift=-8.5cm]
  \draw[pftpink,thick]         (-2cm,0)--(-.8cm,0);
  \node[font=\Huge,anchor=west] at (-.7cm,0) {PFT};
  \draw[jtlime,thick,dashed] (4.5cm,0)--(5.7cm,0);
  \node[font=\Huge,anchor=west] at (5.8cm,0) {JT};
\end{scope}
 
\end{tikzpicture}
}
\caption{Radar plots depicting the Dice of skin lesion segmentation on the ISIC dataset using
both PFT and JT training paradigms across eight representative SSL techniques.}
\label{fig:radar_isic_dice}
\end{figure}

\subsection{Dermatology Segmentation}
\label{appendixisic}
Fig.~\ref{fig:radar_isic_dice} summarizes the ISIC-2016 segmentation trends across different label fractions, showing that JT generally improves or preserves segmentation performance for several SSL methods under moderate supervision, while performance differences become increasingly method-dependent in the extremely low-label regime. Table~\ref{tab:isic_seg} shows that the differences between PFT and JT are smaller than those observed in the classification experiments, although several consistent trends emerge. Under the fully labeled setting, JT improves segmentation performance for reconstruction-oriented methods such as Colorization, Rotation, SimCLR, MoCo, and MAE, with Rotation JT achieving one of the highest Dice scores (0.9002). MoCo and SimCLR also remain highly competitive under both paradigms, indicating strong transferability of contrastive representations for dense medical segmentation tasks. Under reduced label fractions, the behavior becomes increasingly method-dependent. For Colorization and Rotation, JT consistently preserves stronger Dice and mIoU performance compared with PFT at 50\%, 20\%, and 10\% labels while dramatically reducing training time. Similar trends are observed for MoCo and SimCLR, where JT often maintains comparable segmentation quality despite substantially faster optimization. In contrast, several methods, including DINO, MAE, and Barlow Twins, exhibit noticeable performance degradation under JT in the extremely low-label regime, suggesting that jointly optimizing supervised and self-supervised objectives can destabilize representation learning for certain SSL objectives when annotation availability becomes highly limited. The results further highlight the efficiency advantages of JT. Across nearly all SSL methods and label fractions, JT reduces training time by a large margin compared with PFT while frequently maintaining competitive Dice and mIoU scores. These findings suggest that, for medical image segmentation, JT can provide a favorable trade-off between computational efficiency and segmentation performance, although the relative benefit remains highly dependent on the underlying SSL objective and label availability.

\begin{table}
    \centering
    \caption{Segmentation performance on ISIC-2016 under different training schemes.}
    \resizebox{0.4\textheight}{!}{
    \begin{tabular}{llcccc}
        \toprule
        \textbf{Framework} & \textbf{Config} & \textbf{Label (\%)} & \textbf{Dice} & \textbf{mIoU} & \textbf{Training Time (s)}\\
        \midrule
\multirow{8}{*}{Colorization} 
        & PFT & \multirow{2}{*}{100} & 0.8798 & 0.8373 & 1380 \\
          & JT  &  & 0.9019 & 0.8472 & 420  \\
          & PFT & \multirow{2}{*}{50}  & 0.8772 & 0.8273 & 1500 \\
          & JT  &  & 0.8794 & 0.8236 & 240  \\
          & PFT & \multirow{2}{*}{20}  & 0.8313 & 0.7836 & 1500 \\
          & JT  &  & 0.8584 & 0.7944 & 240  \\
          & PFT & \multirow{2}{*}{10}  & 0.8076 & 0.7508 & 1200 \\
          & JT  &  & 0.6865 & 0.7200   & 180  \\
          \midrule
\multirow{8}{*}{Rotation}     
        & PFT & \multirow{2}{*}{100} & 0.8776 & 0.8399 & 1140 \\
          & JT  &  & 0.9002 & 0.8490  & 420  \\
          & PFT & \multirow{2}{*}{50}  & 0.8809 & 0.8284 & 1440 \\
          & JT  &  & 0.8806 & 0.8295 & 300  \\
          & PFT & \multirow{2}{*}{20}  & 0.8666 & 0.8119 & 1440 \\
          & JT  &  & 0.868  & 0.8118 & 240  \\
          & PFT & \multirow{2}{*}{10}  & 0.8598 & 0.7899 & 1200 \\
          & JT  &  & 0.8035 & 0.7512 & 240  \\
          \midrule
\multirow{8}{*}{SimCLR}       
        & PFT & \multirow{2}{*}{100} & 0.8886 & 0.8466 & 1320 \\
          & JT  &  & 0.9005 & 0.8569 & 840  \\
          & PFT & \multirow{2}{*}{50}  & 0.8867 & 0.8421 & 1320 \\
          & JT  &  & 0.8893 & 0.8381 & 720  \\
          & PFT & \multirow{2}{*}{20}  & 0.8779 & 0.8327 & 1320 \\
          & JT  &  & 0.8612 & 0.7955 & 540  \\
          & PFT & \multirow{2}{*}{10}  & 0.8428 & 0.7921 & 1140 \\
          & JT  &  & 0.7828 & 0.7177 & 360  \\
          \midrule
\multirow{8}{*}{BYOL}
        & PFT & \multirow{2}{*}{100} & 0.8978 & 0.8459 & 1500 \\
          & JT  &  & 0.8965 & 0.8434 & 420  \\
          & PFT & \multirow{2}{*}{50}  & 0.8827 & 0.8293 & 1440 \\
          & JT  &  & 0.8650  & 0.8094 & 660  \\
          & PFT & \multirow{2}{*}{20}  & 0.8666 & 0.8110  & 1380 \\
          & JT  &  & 0.8756 & 0.8145 & 660  \\
          & PFT & \multirow{2}{*}{10}  & 0.8034 & 0.7462 & 1260 \\
          & JT  &  & 0.8594 & 0.7790  & 660  \\
          \midrule
\multirow{8}{*}{MoCo}    
        & PFT & \multirow{2}{*}{100} & 0.8946 & 0.8540  & 2940 \\
          & JT  &  & 0.9123 & 0.8704 & 660  \\
          & PFT & \multirow{2}{*}{50}  & 0.8784 & 0.8423 & 3120 \\
          & JT  &  & 0.8751 & 0.8283 & 300  \\
          & PFT & \multirow{2}{*}{20}  & 0.8824 & 0.8421 & 3120 \\
          & JT  &  & 0.8826 & 0.8261 & 360  \\
          & PFT & \multirow{2}{*}{10}  & 0.8777 & 0.8201 & 2880 \\
          & JT  &  & 0.8506 & 0.7915 & 240  \\
          \midrule
\multirow{8}{*}{MAE}      
        & PFT & \multirow{2}{*}{100} & 0.8883 & 0.8440  & 1200 \\
          & JT  &  & 0.8948 & 0.8413 & 480  \\
          & PFT & \multirow{2}{*}{50}  & 0.8839 & 0.8370  & 1320 \\
          & JT  &  & 0.8698 & 0.8252 & 360  \\
          & PFT & \multirow{2}{*}{20}  & 0.8763 & 0.8174 & 1320 \\
          & JT  &  & 0.8811 & 0.8201 & 600  \\
          & PFT & \multirow{2}{*}{10}  & 0.8411 & 0.7750  & 1140 \\
          & JT  &  & 0.7631 & 0.7200   & 240  \\
          \midrule
\multirow{8}{*}{DINO}    
        & PFT & \multirow{2}{*}{100} & 0.888  & 0.8447 & 3600 \\
          & JT  &  & 0.8959 & 0.8354 & 840  \\
          & PFT & \multirow{2}{*}{50}  & 0.8853 & 0.8300   & 3900 \\
          & JT  &  & 0.8834 & 0.8245 & 660  \\
          & PFT & \multirow{2}{*}{20}  & 0.8722 & 0.8097 & 3900 \\
          & JT  &  & 0.8427 & 0.7701 & 540  \\
          & PFT & \multirow{2}{*}{10}  & 0.8517 & 0.7880  & 3600 \\
          & JT  &  & 0.7058 & 0.6146 & 480  \\
          \midrule
\multirow{8}{*}{Barlow Twins} 
        & PFT & \multirow{2}{*}{100} & 0.8886 & 0.845  & 1860 \\
          & JT  &  & 0.8718 & 0.8335 & 900  \\
          & PFT & \multirow{2}{*}{50}  & 0.8752 & 0.8313 & 1680 \\
          & JT  &  & 0.8370  & 0.8100   & 420  \\
          & PFT & \multirow{2}{*}{20}  & 0.8319 & 0.7902 & 1800 \\
          & JT  &  & 0.8461 & 0.7862 & 720  \\
          & PFT & \multirow{2}{*}{10}  & 0.8244 & 0.7633 & 1320 \\
          & JT  &  & 0.7062 & 0.6682 & 300  \\ 
          \bottomrule
    \end{tabular}
    }
    \label{tab:isic_seg}
\end{table}

\clearpage

\section{Image Quality Assessment}
\label{appendix:iqa}

\begin{table*}[h]
\centering
\scriptsize
\setlength{\tabcolsep}{3pt}
\caption{KADID-10K IQA Performance Comparison (SROCC/PLCC) Across Different SSL Methods and Label Regimes.}
\label{tab:kadid_results}

\resizebox{\textwidth}{!}{%

\begin{tabular}{lcc|cc|cc|cc|cc}
\hline

\multirow{2}{*}{SSL Method} &
\multirow{2}{*}{PFT} &
\multirow{2}{*}{JT} &
\multicolumn{2}{c|}{10\% Labels} &
\multicolumn{2}{c|}{20\% Labels} &
\multicolumn{2}{c|}{50\% Labels} &
\multicolumn{2}{c}{100\% Labels} \\

& & &
SROCC & PLCC &
SROCC & PLCC &
SROCC & PLCC &
SROCC & PLCC \\

\hline

\multirow{2}{*}{SimCLR}
& $\checkmark$ &  &
0.7844 & 0.7951 &
0.8138 & 0.8209 &
0.8460 & 0.8553 &
0.8601 & 0.8688 \\

&  & $\checkmark$
& 0.8279 & 0.8490
& 0.8370 & 0.8510
& 0.8618 & 0.8653
& 0.8634 & 0.8680 \\

\hline

\multirow{2}{*}{MOCO}
& $\checkmark$ &  &
0.8335 & 0.8429 &
0.9014 & 0.9050 &
0.9410 & 0.9446 &
0.9487 & 0.9521 \\

&  & $\checkmark$
& 0.6957 & 0.7079
& 0.7805 & 0.7923
& 0.8678 & 0.8706
& 0.8820 & 0.8806 \\

\hline

\multirow{2}{*}{BYOL}
& $\checkmark$ &  &
0.1199 & 0.1269 &
0.3101 & 0.3261 &
0.4644 & 0.4850 &
0.5642 & 0.5857 \\

&  & $\checkmark$
& 0.1023 & 0.1123
& 0.2934 & 0.3152
& 0.4436 & 0.4734
& 0.5535 & 0.5759 \\

\hline

\multirow{2}{*}{DINO}
& $\checkmark$ &  &
0.8223 & 0.8295 &
0.8872 & 0.8934 &
0.9294 & 0.9338 &
0.9458 & 0.9511 \\

&  & $\checkmark$
& 0.2260 & 0.2886
& 0.3235 & 0.3860
& 0.5450 & 0.5980
& 0.6420 & 0.6540 \\

\hline

\multirow{2}{*}{Barlow Twins}
& $\checkmark$ &  &
0.2960 & 0.3110 &
0.3560 & 0.3970 &
0.5510 & 0.5890 &
0.6650 & 0.6792 \\

&  & $\checkmark$
& 0.2710 & 0.2910
& 0.3150 & 0.3780
& 0.5220 & 0.5490
& 0.6540 & 0.6692 \\

\hline

\multirow{2}{*}{MAE}
& $\checkmark$ &  &
0.4123 & 0.4213 &
0.5239 & 0.5567 &
0.7341 & 0.7516 &
0.8592 & 0.8691 \\

&  & $\checkmark$
& 0.6154 & 0.6532
& 0.6921 & 0.7397
& 0.8763 & 0.8897
& 0.9083 & 0.9087 \\

\hline

\multirow{2}{*}{Rotation}
& $\checkmark$ &  &
0.4261 & 0.4473 &
0.5739 & 0.5931 &
0.7364 & 0.7791 &
0.8581 & 0.8594 \\

&  & $\checkmark$
& 0.4932 & 0.5276
& 0.5317 & 0.5831
& 0.7771 & 0.7913
& 0.8829 & 0.8863 \\

\hline

\multirow{2}{*}{Colorization}
& $\checkmark$ &  &
0.4268 & 0.4982 &
0.5231 & 0.5547 &
0.7853 & 0.7891 &
0.8025 & 0.8068 \\

&  & $\checkmark$
& 0.4123 & 0.4443
& 0.5294 & 0.5342
& 0.7931 & 0.8197
& 0.8417 & 0.8472 \\

\hline

\end{tabular}
}

\end{table*}

\subsection{SROCC Analysis} 
\begin{figure}[h]
    \centering
    \includegraphics[width=\linewidth]{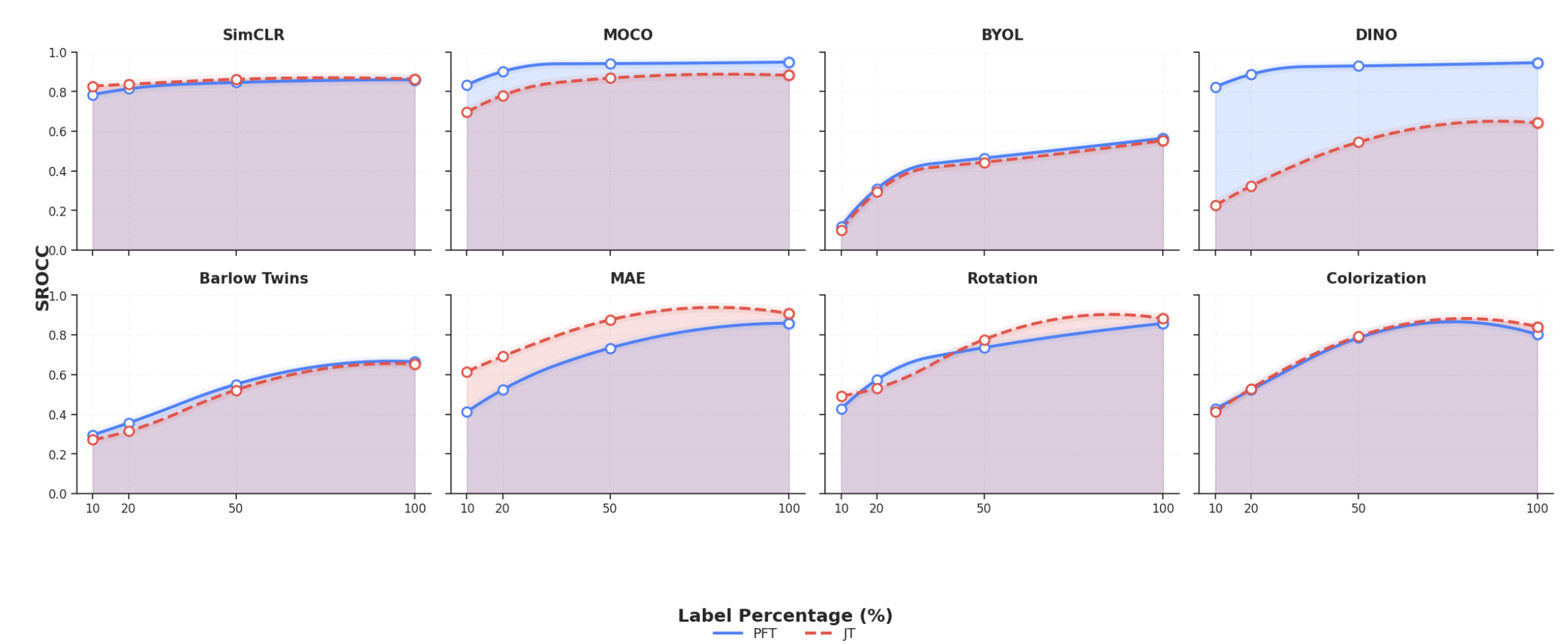}
    \caption{Comparison of SROCC performance of self-supervised learning (SSL) methods across different label regimes on the KADID-10K dataset under pretraining-finetuning (PFT) and joint training (JT) strategies.
}
    \label{fig:kadid_srocc}
\end{figure}

\begin{figure}[h]
    \centering
    \includegraphics[width=\linewidth]{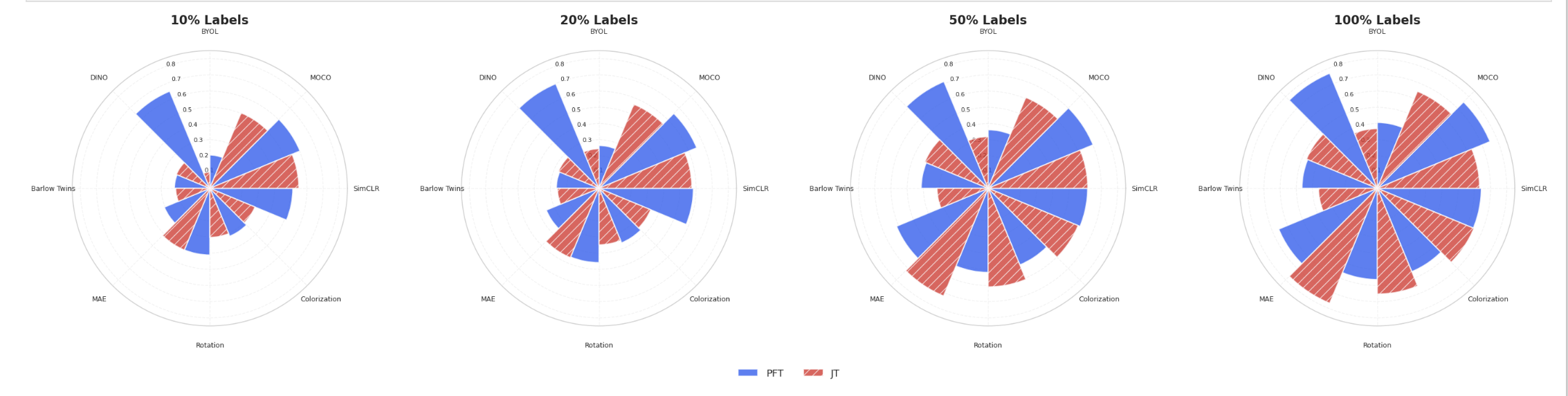}
    \caption{KonIQ SROCC comparison of SSL methods across different label regimes under PFT and JT training.}
    \label{fig:koniq_srocc}
\end{figure}

\begin{figure}[h]
    \centering
    \includegraphics[width=\linewidth]{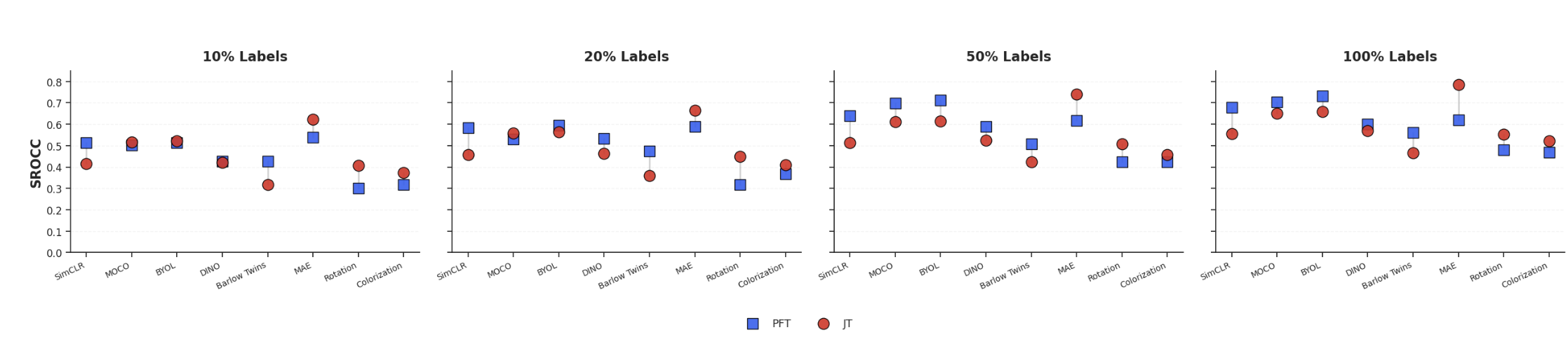}
    \caption{Comparison of SROCC and PLCC performance under clean and adversarial settings for MoCo- and DINO-based models, highlighting the impact of adversarial perturbations on correlation accuracy and robustness.
}
    \label{fig:ldct_sroc}
\end{figure}

Based on the SROCC results from Table~\ref{tab:kadid_results} and Table~\ref{tab:ldctiqa_results}, the three datasets exhibit distinct SSL adaptation behaviors for image quality evaluation. Fig.~\ref{fig:kadid_srocc} shows that on KADID-10K, contrastive methods such as MOCO and DINO achieve the strongest performance under PFT, particularly at higher label regimes, indicating the effectiveness of transferable semantic representations for synthetic distortion assessment. Similarly, from Fig.~\ref{fig:koniq_srocc}, KonIQ-10K demonstrates consistent performance improvements with increased label availability, where contrastive approaches remain competitive under PFT while reconstruction-oriented methods such as MAE show improved robustness under JT. In contrast, from Fig.~\ref{fig:ldct_sroc}, the LDCTIQA results reveal that reconstruction and auxiliary-task-based methods, including MAE, Rotation, and Colorization, generally benefit more from JT compared to PFT, especially in low-label settings, suggesting that preserving structural and distortion-sensitive representations is particularly important for medical image quality evaluation.

\begin{table}[t]
\centering
\scriptsize
\setlength{\tabcolsep}{3pt}
\caption{Performance comparison of different SSL methods on the LDCTIQA dataset under progressive fine-tuning (PFT) and joint training (JT) settings. The table reports SROCC and PLCC results across four label regimes (10\%, 20\%, 50\%, and 100\% labeled training data) using a ResNet-18 backbone. Higher values indicate better perceptual quality prediction performance.}
\label{tab:ldctiqa_results}

\resizebox{\textwidth}{!}{%

\begin{tabular}{lcc|cc|cc|cc|cc}
\hline

\multirow{2}{*}{SSL Method} &
\multirow{2}{*}{PFT} &
\multirow{2}{*}{JT} &
\multicolumn{2}{c|}{10\% Labels} &
\multicolumn{2}{c|}{20\% Labels} &
\multicolumn{2}{c|}{50\% Labels} &
\multicolumn{2}{c}{100\% Labels} \\

& & &
SROCC & PLCC &
SROCC & PLCC &
SROCC & PLCC &
SROCC & PLCC \\

\hline

\multirow{2}{*}{SimCLR}
& $\checkmark$ &  &
0.5139 & 0.5007 &
0.5833 & 0.5732 &
0.6397 & 0.5912 &
0.6773 & 0.6241 \\

&  & $\checkmark$
& 0.4168 & 0.4019
& 0.4582 & 0.4375
& 0.5134 & 0.5006
& 0.5543 & 0.5338 \\

\hline

\multirow{2}{*}{MOCO}
& $\checkmark$ &  &
0.5012 & 0.4866 &
0.5298 & 0.5094 &
0.6977 & 0.6725 &
0.7049 & 0.6896 \\

&  & $\checkmark$
& 0.5157 & 0.5013
& 0.5568 & 0.5362
& 0.6104 & 0.6009
& 0.6517 & 0.6308 \\

\hline

\multirow{2}{*}{BYOL}
& $\checkmark$ &  &
0.5129 & 0.5060 &
0.5936 & 0.5487 &
0.7122 & 0.6848 &
0.7325 & 0.7056 \\

&  & $\checkmark$
& 0.5209 & 0.5021
& 0.5624 & 0.5416
& 0.6152 & 0.6023
& 0.6581 & 0.6365 \\

\hline

\multirow{2}{*}{DINO}
& $\checkmark$ &  &
0.4273 & 0.4009 &
0.5321 & 0.5171 &
0.5879 & 0.5742 &
0.6013 & 0.5981 \\

&  & $\checkmark$
& 0.4206 & 0.4012
& 0.4627 & 0.4413
& 0.5253 & 0.5038
& 0.5689 & 0.5467 \\

\hline

\multirow{2}{*}{Barlow Twins}
& $\checkmark$ &  &
0.4279 & 0.4138 &
0.4736 & 0.4525 &
0.5087 & 0.4871 &
0.5619 & 0.5017 \\

&  & $\checkmark$
& 0.3169 & 0.3023
& 0.3598 & 0.3384
& 0.4235 & 0.4027
& 0.4663 & 0.4448 \\

\hline

\multirow{2}{*}{MAE}
& $\checkmark$ &  &
0.5398 & 0.5122 &
0.5896 & 0.5432 &
0.6170 & 0.5742 &
0.6209 & 0.6017 \\

&  & $\checkmark$
& 0.6213 & 0.6036
& 0.6648 & 0.6427
& 0.7387 & 0.7129
& 0.7861 & 0.7614 \\

\hline

\multirow{2}{*}{Rotation}
& $\checkmark$ &  &
0.3014 & 0.3125 &
0.3179 & 0.3365 &
0.4236 & 0.4631 &
0.4796 & 0.4905 \\

&  & $\checkmark$
& 0.4058 & 0.4276
& 0.4492 & 0.4715
& 0.5089 & 0.5304
& 0.5518 & 0.5736 \\

\hline

\multirow{2}{*}{Colorization}
& $\checkmark$ &  &
0.3173 & 0.3334 &
0.3681 & 0.3867 &
0.4237 & 0.4399 &
0.4693 & 0.4873 \\

&  & $\checkmark$
& 0.3746 & 0.3910
& 0.4097 & 0.4342
& 0.4563 & 0.4781
& 0.5228 & 0.5298 \\

\hline

\end{tabular}
}

\end{table}

\begin{figure}[t]
    \centering
    \includegraphics[width=\linewidth]{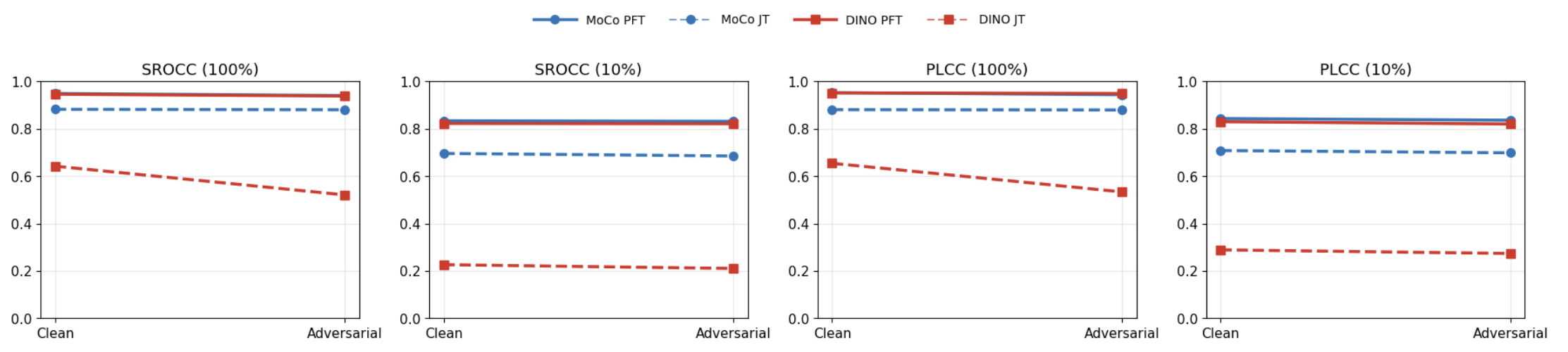}
    \caption{Comparison of SROCC and PLCC performance under clean and adversarial conditions for MoCo- and DINO-based models, highlighting the impact of adversarial perturbations on prediction reliability and correlation accuracy}
    \label{fig:iqa_adv}
\end{figure}

\subsection{PLCC Analysis}
\begin{figure}[h]
    \centering
    \includegraphics[width=\linewidth]{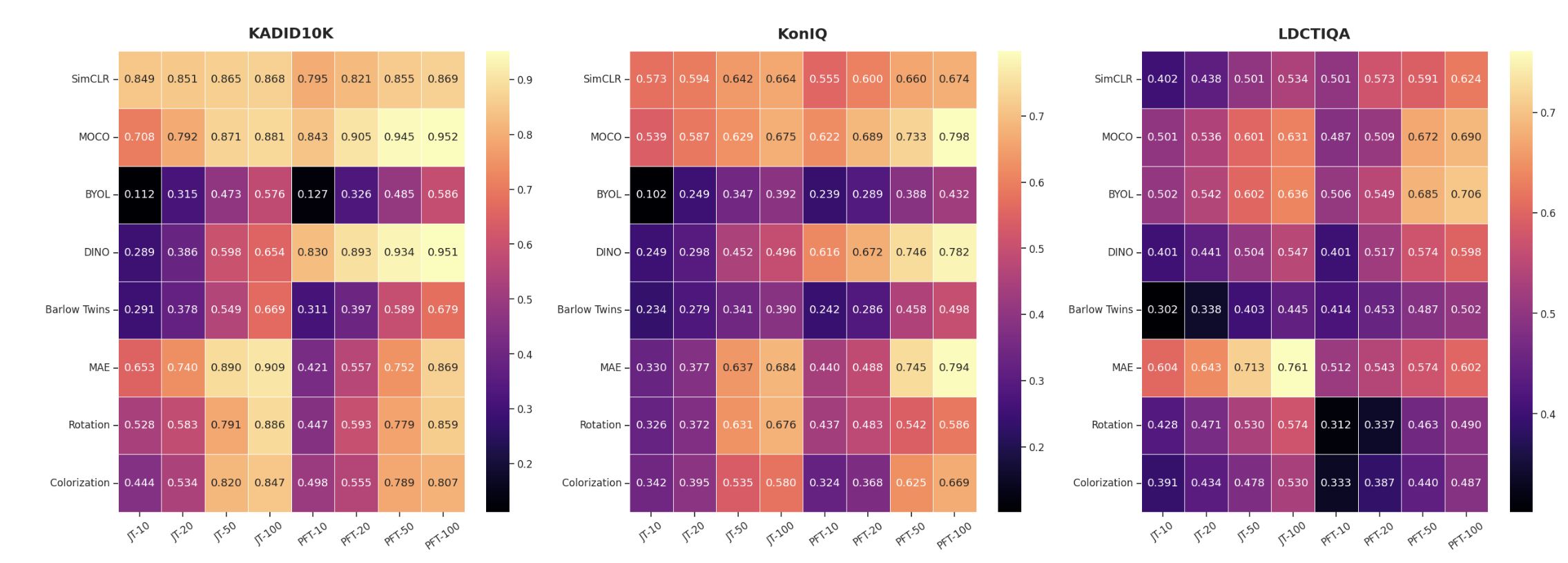}
    \caption{Cross-Dataset PLCC Performance Analysis of SSL Methods for Image Quality Assessment}
    \label{fig:kadid_plcc}
\end{figure}

\begin{figure}[t]
    \centering
    \includegraphics[width=\linewidth]{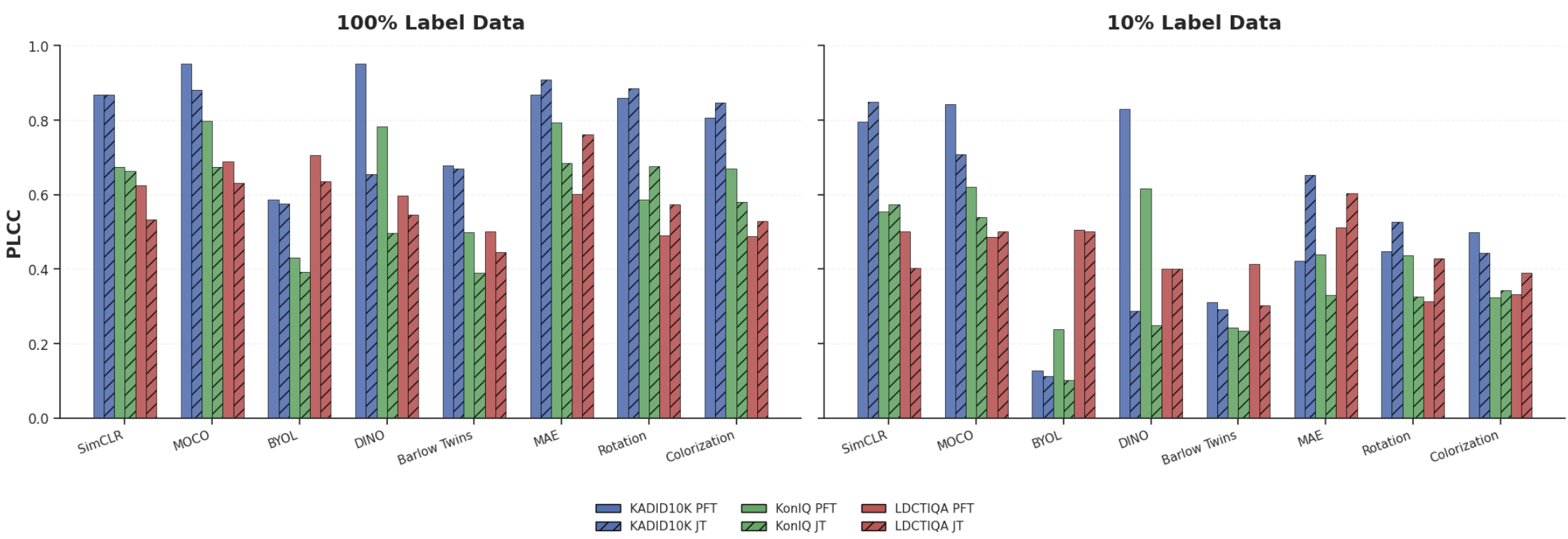}
    \caption{Cross-Dataset Image Quality Assessment PLCC Comparison of SSL Methods Under 10\% and 100\% Label Regimes with PFT and JT Training Strategies}
    \label{fig:iqa_plcc}
\end{figure}

Based on the PLCC results, similar adaptation trends can be observed across the three datasets for SSL-based image quality evaluation. From Fig.~\ref{fig:kadid_plcc}, on KADID-10K, contrastive methods such as MOCO and DINO achieve the highest PLCC performance under PFT, particularly at higher label regimes, indicating strong perceptual correlation with subjective quality scores. KonIQ-10K (Fig.~\ref{fig:iqa_plcc}) also demonstrates improved PLCC performance with increased label availability, where contrastive approaches remain highly effective under PFT while reconstruction-oriented methods such as MAE exhibit stronger robustness under JT. In contrast, the LDCTIQA results show that reconstruction and auxiliary-task-based SSL methods, including MAE, Rotation, and Colorization, generally benefit more from JT, especially under limited supervision, highlighting the importance of preserving structural and low-level perceptual representations for medical image quality evaluation.

\subsection{IQA: JT or PFT?}
We evaluate the robustness of SSL–based image quality assessment models on the KADID-10k under an extremely low-magnitude adversarial setting, where perturbations are restricted to modifying only one or two pixels per image (Fig.~\ref{fig:iqa_adv}). Across different evaluated models, such as MoCo and DINO, we observe that such localized perturbations do not meaningfully degrade performance, with both rank-based and correlation-based metrics remaining largely stable. This indicates that SSL-based IQA models are inherently robust to sparse, pixel-level noise, likely because their learned representations emphasize global structural and semantic information rather than isolated local variations.

PFT is generally more effective for image quality assessment scenarios where sufficient labeled data is available, and the SSL backbone is based on contrastive representation learning methods such as MOCO, DINO, and SimCLR. The experimental results on KADID-10K and KonIQ-10K show that PFT consistently achieves stronger SROCC and PLCC performance in high-label regimes, indicating that preserving pretrained semantic feature representations is beneficial for perceptual quality prediction, particularly for synthetic and authentic natural-image distortions.

In contrast, JT becomes more advantageous in low-label settings and for reconstruction or auxiliary-task-based SSL approaches such as MAE, Rotation, and Colorization. The results, particularly on LDCTIQA, demonstrate that JT better preserves structural, low-level, and distortion-sensitive representations that are important for medical and perceptual quality evaluation. These findings suggest that JT is preferable when label availability is limited or when the downstream IQA task relies heavily on fine-grained structural fidelity rather than high-level semantic representations.

\begin{figure}[h]
    \centering
    \includegraphics[width=\linewidth]{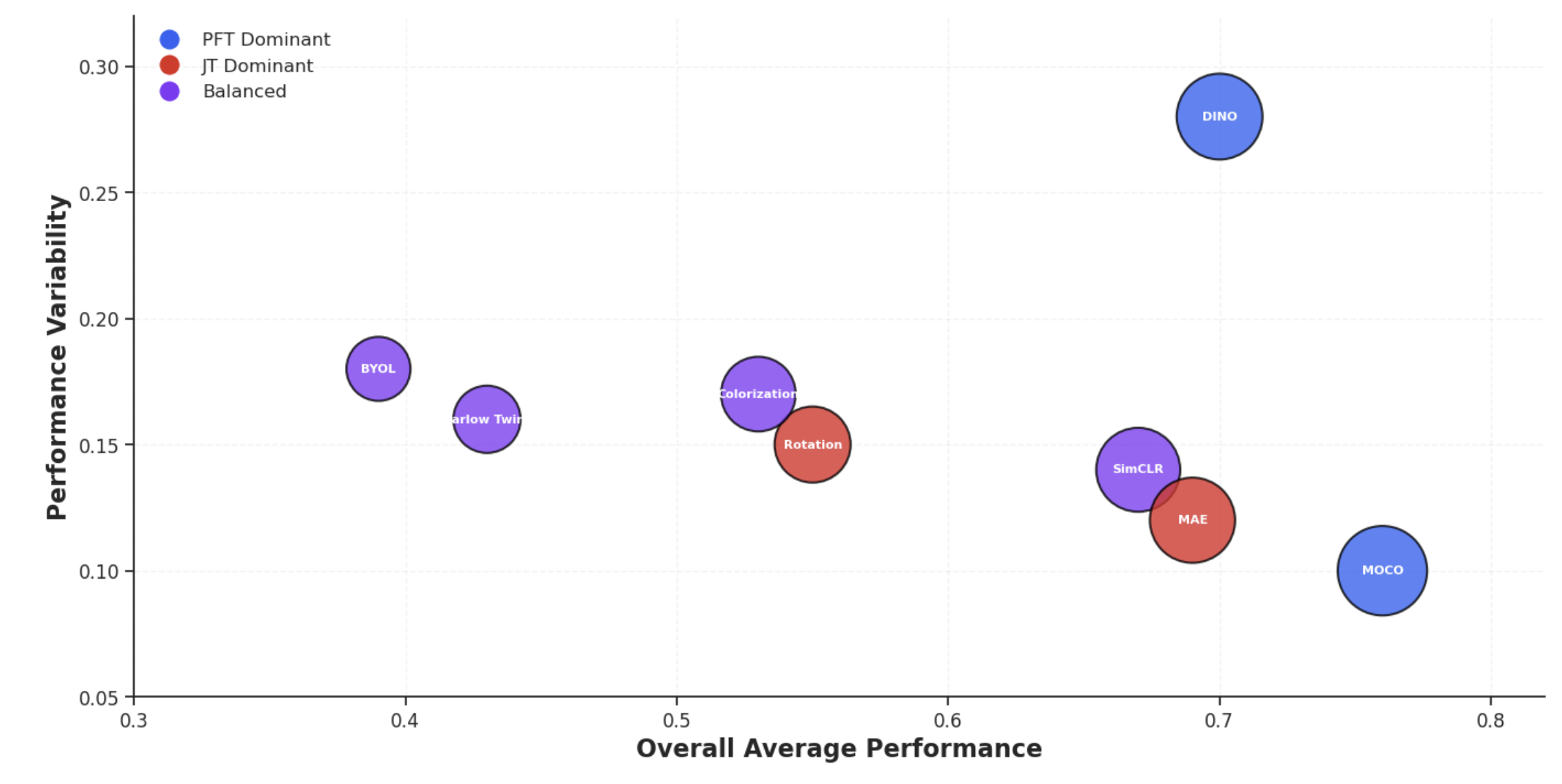}
    \caption{Overall SSL Benchmark Landscape Across IQA Datasets}
    \label{fig:ssl_iqa}
\end{figure}

\subsection{Overall Analysis}
The overall benchmark analysis across KADID-10K, KonIQ-10K, and LDCTIQA in Fig.~\ref{fig:ssl_iqa} demonstrates that MOCO achieves the strongest average performance with an overall score of approximately 0.76, followed by DINO (0.70), MAE (0.69), and SimCLR (0.67). Among these methods, MOCO and DINO show strong dominance under PFT, indicating the effectiveness of contrastive representation learning for image quality evaluation. In contrast, MAE exhibits comparatively stronger performance under JT, particularly in structurally sensitive and low-label scenarios. Furthermore, methods such as BYOL and Barlow Twins achieve comparatively lower overall scores, suggesting reduced robustness and generalization capability across diverse IQA settings.

\clearpage

\section{Earthscape Dataset Performance}
\label{appendix:earthscape}

\begin{figure}[t]
    \centering
    \includegraphics[width=\linewidth]{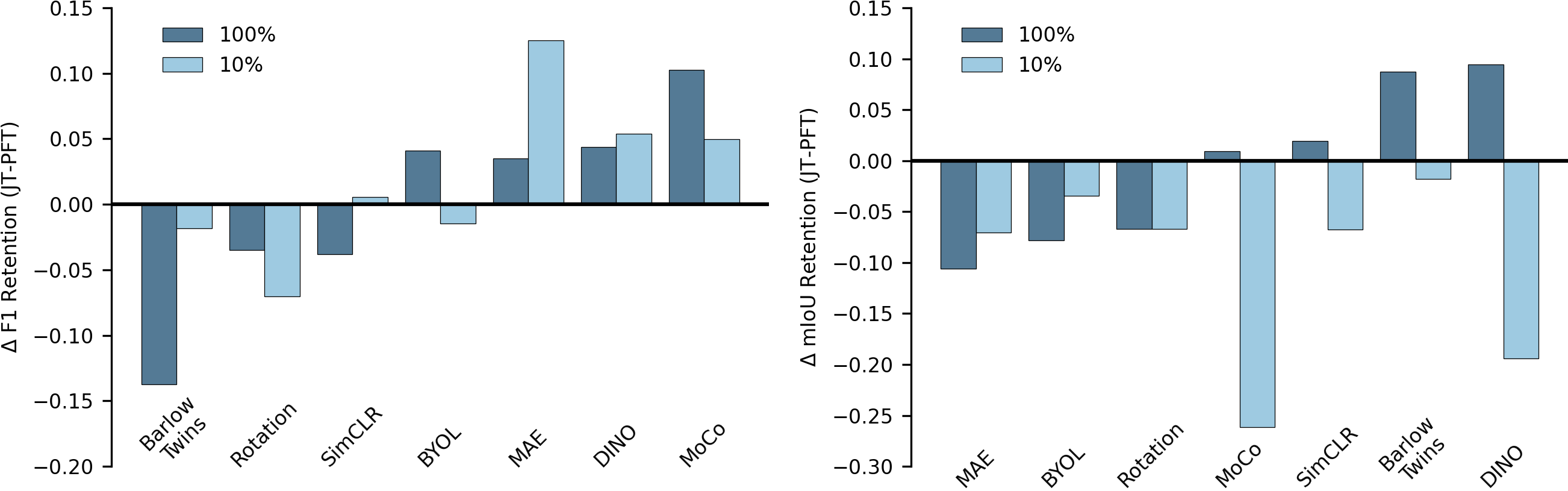}
    \caption{Change in geographic domain-shift retention under joint training ($\mathrm{JT}-\mathrm{PFT}$) across SSL methods on EarthScape for multilabel classification (left) and semantic segmentation (right) under full (100\%) and limited (10\%) supervision. Retention is computed as cross-domain performance divided by in-domain performance using F1 for classification and mean IoU for segmentation. Positive values indicate greater retention under JT relative to PFT, whereas negative values indicate stronger retention under PFT.}
    \label{fig:earthscape_shift}
\end{figure}

\begin{table*}[h]
\centering
\caption{Multilabel classification performance on EarthScape using DEM inputs under PFT and JT training regimes with full (100\%) and limited (10\%) supervision. Results are reported for both in-domain and cross-domain test sets using accuracy and F1 score, along with training time (hours).}
\begin{tabular}{l l c c c c c c}
\toprule
\multirow{2}{*}{\textbf{Framework}} & \multirow{2}{*}{\textbf{Config}} & \multirow{2}{*}{\textbf{Labels}} &
\multicolumn{2}{c}{\textbf{In-domain}} &
\multicolumn{2}{c}{\textbf{Cross-domain}} & \multirow{2}{2cm}{\textbf{\centering{Training Time (hrs)}}} \\
\cmidrule(lr){4-5} \cmidrule(lr){6-7}
& & & \textbf{Acc.} & \textbf{F1} & \textbf{Acc.} & \textbf{F1} \\
\midrule

\multirow{4}{*}{\textbf{Barlow Twins}}
 & PFT & \multirow{2}{*}{100\%} & 0.5975 & 0.8153 & 0.4772 & 0.7790 & 0.98 \\
 & JT  &                        & 0.5704 & 0.7929 & 0.3772 & 0.7401 & 7.12 \\
 & PFT & \multirow{2}{*}{10\%}  & 0.5816 & 0.7981 & 0.4038 & 0.7514 & 0.94 \\
 & JT  &                        & 0.5758 & 0.7906 & 0.3893 & 0.7367 & 1.02 \\
\midrule

\multirow{4}{*}{\textbf{BYOL}}
 & PFT & \multirow{2}{*}{100\%} & 0.5827 & 0.7878 & 0.4514 & 0.7314 & 1.04 \\
 & JT  &                        & 0.6073 & 0.7943 & 0.4954 & 0.7311 & 5.96 \\
 & PFT & \multirow{2}{*}{10\%}  & 0.5631 & 0.7381 & 0.431 & 0.7241 & 0.97 \\
 & JT  &                        & 0.6022 & 0.8089 & 0.452 & 0.7357 & 1.05 \\
\midrule

\multirow{4}{*}{\textbf{DINO}}
 & PFT & \multirow{2}{*}{100\%} & 0.6230 & 0.8487 & 0.4862 & 0.8038 & 14.05 \\
 & JT  &                        & 0.5126 & 0.5552 & 0.4224 & 0.4964 & 42.08 \\
 & PFT & \multirow{2}{*}{10\%}  & 0.6095 & 0.8144 & 0.458 & 0.7637 & 13.99 \\
 & JT  &                        & 0.5095 & 0.4638 & 0.4104 & 0.3613 & 17.33 \\
\midrule

\multirow{4}{*}{\textbf{MAE}}
 & PFT & \multirow{2}{*}{100\%} & 0.6183 & 0.8092 & 0.4782 & 0.7741 & 0.71 \\
 & JT  &                        & 0.6311 & 0.8315 & 0.5101 & 0.7940 & 0.73 \\
 & PFT & \multirow{2}{*}{10\%}  & 0.5888 & 0.7716 & 0.4170 & 0.7003 & 0.62 \\
 & JT  &                        & 0.6116 & 0.8129 & 0.5097 & 0.7837 & 0.70 \\
\midrule

\multirow{4}{*}{\textbf{MoCo}}
 & PFT & \multirow{2}{*}{100\%} & 0.5926 & 0.7912 & 0.3852 & 0.7144 & 0.88 \\
 & JT  &                        & 0.6076 & 0.7965 & 0.4573 & 0.7276 & 1.01 \\
 & PFT & \multirow{2}{*}{10\%}  & 0.5750 & 0.7735 & 0.3827 & 0.7354 & 0.84 \\
 & JT  &                        & 0.6252 & 0.8242 & 0.4472 & 0.7414 & 0.96 \\
\midrule

\multirow{4}{*}{\textbf{Rotation}}
 & PFT & \multirow{2}{*}{100\%} & 0.5986 & 0.8025 & 0.4879 & 0.7792 & 0.75 \\
 & JT  &                        & 0.6155 & 0.8371 & 0.4803 & 0.7626 & 0.77 \\
 & PFT & \multirow{2}{*}{10\%}  & 0.5350 & 0.7683 & 0.4414 & 0.7019 & 0.63 \\
 & JT  &                        & 0.6168 & 0.8504 & 0.4655 & 0.7621 & 0.67 \\
\midrule

\multirow{4}{*}{\textbf{SimCLR}}
 & PFT & \multirow{2}{*}{100\%} & 0.592 & 0.8318 & 0.4334 & 0.7683 & 0.98 \\
 & JT  &                        & 0.5727 & 0.7884 & 0.3975 & 0.7118 & 6.24 \\
 & PFT & \multirow{2}{*}{10\%}  & 0.5768 & 0.8125 & 0.414 & 0.7759 & 0.93 \\
 & JT  &                        & 0.5868 & 0.8014 & 0.4244 & 0.7542 & 1.02 \\

\bottomrule
\end{tabular}
\label{tab:earthscape_classification}
\end{table*}

\begin{table*}[t]
\centering
\caption{Semantic segmentation performance on EarthScape using DEM inputs under PFT and JT training regimes with full (100\%) and limited (10\%) supervision. Results are reported for both in-domain and cross-domain test sets using mIoU and Dice score, along with training time (hours).}
\begin{tabular}{l l c c c c c c}
\toprule
\multirow{2}{*}{\textbf{Framework}} & \multirow{2}{*}{\textbf{Config}} & \multirow{2}{*}{\textbf{Labels}} &
\multicolumn{2}{c}{\textbf{In-domain}} &
\multicolumn{2}{c}{\textbf{Cross-domain}} & \multirow{2}{2cm}{\textbf{\centering{Training Time (hrs)}}} \\
\cmidrule(lr){4-5} \cmidrule(lr){6-7}
& & & \textbf{mIoU} & \textbf{Dice} & \textbf{mIoU} & \textbf{Dice} \\
\midrule

\multirow{4}{*}{\textbf{Barlow Twins}}
 & PFT & \multirow{2}{*}{100\%} & 0.3691 & 0.4273 & 0.3217 & 0.3661 & 0.91 \\
 & JT  &                        & 0.3257 & 0.3857 & 0.3123 & 0.3531 & 1.27 \\
 & PFT & \multirow{2}{*}{10\%}  & 0.3249 & 0.3768 & 0.3084 & 0.3397 & 0.78 \\
 & JT  &                        & 0.3221 & 0.3741 & 0.2999 & 0.3394 & 1.21 \\
\midrule

\multirow{4}{*}{\textbf{BYOL}}
 & PFT & \multirow{2}{*}{100\%} & 0.3487 & 0.4008 & 0.3094 & 0.3412 & 0.93 \\
 & JT  &                        & 0.3820 & 0.4357 & 0.3091 & 0.3515 & 1.29 \\
 & PFT & \multirow{2}{*}{10\%}  & 0.3082 & 0.3541 & 0.2964 & 0.3308 & 0.78 \\
 & JT  &                        & 0.3984 & 0.4561 & 0.3693 & 0.4192 & 1.18 \\
\midrule

\multirow{4}{*}{\textbf{DINO}}
 & PFT & \multirow{2}{*}{100\%} & 0.4170 & 0.4803 & 0.3463 & 0.3959 & 1.28 \\
 & JT  &                        & 0.2813 & 0.3338 & 0.2601 & 0.3071 & 1.51 \\
 & PFT & \multirow{2}{*}{10\%}  & 0.3612 & 0.4152 & 0.3467 & 0.3900 & 1.03 \\
 & JT  &                        & 0.3357 & 0.3928 & 0.257 & 0.3010 & 1.47 \\
\midrule

\multirow{4}{*}{\textbf{MAE}}
 & PFT & \multirow{2}{*}{100\%} & 0.4217 & 0.4813 & 0.3625 & 0.4147 & 0.55 \\
 & JT  &                        & 0.4414 & 0.5005 & 0.3326 & 0.3832 & 0.95 \\
 & PFT & \multirow{2}{*}{10\%}  & 0.3916 & 0.4452 & 0.3599 & 0.4034 & 0.44 \\
 & JT  &                        & 0.4184 & 0.4767 & 0.3549 & 0.4060 & 0.86 \\
\midrule

\multirow{4}{*}{\textbf{MoCo}}
 & PFT & \multirow{2}{*}{100\%} & 0.3426 & 0.3939 & 0.3027 & 0.3332 & 1.00 \\
 & JT  &                        & 0.3718 & 0.4265 & 0.3319 & 0.3756 & 1.23 \\
 & PFT & \multirow{2}{*}{10\%}  & 0.3043 & 0.3468 & 0.3192 & 0.3470 & 0.75 \\
 & JT  &                        & 0.3905 & 0.4464 & 0.3074 & 0.3487 & 1.12 \\
\midrule

\multirow{4}{*}{\textbf{Rotation}}
 & PFT & \multirow{2}{*}{100\%} & 0.3675 & 0.4286 & 0.3347 & 0.3836 & 0.52 \\
 & JT  &                        & 0.4423 & 0.5032 & 0.3733 & 0.4266 & 0.44 \\
 & PFT & \multirow{2}{*}{10\%}  & 0.3187 & 0.3739 & 0.3007 & 0.3456 & 0.46 \\
 & JT  &                        & 0.4281 & 0.4896 & 0.3752 & 0.4281 & 0.76 \\
\midrule

\multirow{4}{*}{\textbf{SimCLR}}
 & PFT & \multirow{2}{*}{100\%} & 0.3702 & 0.4289 & 0.3252 & 0.3693 & 0.99 \\
 & JT  &                        & 0.3401 & 0.3966 & 0.3054 & 0.3400 & 1.27 \\
 & PFT & \multirow{2}{*}{10\%}  & 0.3211 & 0.3735 & 0.3209 & 0.3545 & 0.81 \\
 & JT  &                        & 0.3444 & 0.4039 & 0.3210 & 0.3564 & 1.20 \\

\bottomrule
\end{tabular}
\label{tab:earthscape_segmentation}
\end{table*}

Tables~\ref{tab:earthscape_classification} and~\ref{tab:earthscape_segmentation} show multilabel classification and semantic segmentation performance on EarthScape using a single-channel digital elevation model (DEM) input. The dataset exhibits substantial class imbalance and a long-tailed distribution across seven classes. Across both tasks, MAE, MoCo, BYOL, and Rotation consistently benefit from joint optimization, whereas Barlow Twins and DINO exhibit degraded performance with JT, while SimCLR shows mixed behavior depending on the supervision regime. DINO and MAE achieve the strongest overall performance under PFT, whereas Rotation and MAE emerge as the top-performing methods with JT. Notably, DINO transitions from one of the strongest performers under PFT to one of the weakest under JT, indicating a potential incompatibility between its self-distillation objective and the supervised loss. 

Under limited supervision, performance degrades across all methods, but the drop is consistently larger for PFT than JT (Fig.~\ref{fig:earthscape_delta}). Several methods maintain or slightly improve performance under reduced supervision with JT, suggesting that joint optimization can substantially mitigate the effects of limited supervision and, in some cases, approach or exceed fully supervised PFT performance. This effect is particularly pronounced for segmentation, where JT produces substantially larger gains than in classification, especially under limited supervision (Fig.~\ref{fig:earthscape_delta}). 

Geographic domain-shift retention under JT is substantially more variable than in-domain performance improvements (Fig.~\ref{fig:earthscape_shift}). While several methods exhibit strong in-domain gains under joint optimization, these improvements do not consistently translate to improved cross-domain retention, particularly for segmentation under limited supervision. This suggests that methods benefiting from JT may become more specialized to the training distribution, resulting in weaker geographic transferability despite improved in-domain accuracy.

These results highlight two key and distinct observations. First, the relative behavior of SSL methods under PFT and JT is consistent across both multilabel classification and segmentation, indicating that the benefits and failures of joint optimization are intrinsic to the SSL objectives themselves, rather than dependent on the downstream task (Fig.~\ref{fig:earthscape_delta}). Second, while these trends are stable, the absolute performance of individual methods depends on their alignment with the underlying data structure. In the case of DEM inputs, methods that preserve or exploit geometric structure perform more effectively, whereas those that rely on transformations that disrupt spatial relationships are less well suited. Together, these findings suggest that the effectiveness of JT is governed by both optimization dynamics and compatibility between the SSL objective and the data modality.

\clearpage

\section{Experimental Details}
This section outlines the complete training configurations used throughout our experiments, including optimization settings, data preprocessing procedures, and hardware specifics. These details offer the full experimental specification needed to reproduce all reported results.

\subsection{Hyperparameters}
We report the dataset-specific training hyperparameters used in all self-supervised pre-training, fine-tuning, and joint training experiments.

\noindent\textbf{CIFAR-10:}
For all CIFAR-10 experiments, we use a unified optimization setup across pre-training, linear evaluation, and joint training. All SSL frameworks use a ResNet-18 backbone, except MAE, which uses a ViT-Base encoder. Training is performed with SGD using momentum 0.9, a weight decay of $5\times10^{-4}$, cosine annealing learning-rate scheduling, and a fixed random seed of 1. Each method is pre-trained for 200 epochs with a batch size of 512 and a learning rate of $6\times10^{-2}$. During linear evaluation, the encoder is frozen, and only a linear classifier is trained for 100 epochs using the same batch size (512). Joint training follows the same schedule as pre-training, using 200 epochs, a batch size of 512, and a learning rate of $6\times10^{-2}$. This configuration applies consistently across SimCLR, BYOL, MoCo, DINO, and Barlow Twins. MAE uses the same schedule and hyperparameters, differing only in its ViT-Base backbone.

\noindent\textbf{COCO:} For the object detection experiments on COCO, all SSL models were pre-trained using SGD optimization with cosine annealing learning rate scheduling, a momentum of 0.9, and weight decay of 1$\times10^{-4}$. During both pre-training and fine-tuning, the models were trained for 100 epochs. In the pre-training phase, each SSL method was trained using a ResNet-18 backbone. During fine-tuning and joint training, a backbone of YOLOv12s was used. When YOLOv12s was used, the learning rate was $1\times10^{-3}$, and the batch size was reduced to 32 compared to the 256 batch size used for pre-training. The only deviations from the hyperparameters described previously were DINO, which used the AdamW optimizer, and MAE, which employed AdamW as well as a ViT backbone.

\noindent\textbf{KADID-10k:}
For all KADID-10k experiments, we follow a consistent setup across SimCLR pre-training, frozen-backbone regression PFT, and end-to-end/JT. All models use a ResNet-18 backbone, and during pre-training, we generate two augmented views of each image using a transform (random cropping, color jittering, and Gaussian blur). The encoder and projection head are trained for 200 epochs with SGD (momentum~0.9, weight decay $5 \times 10^{-4}$) using cosine-annealed learning rates and a starting learning rate of $6 \times 10^{-2}$, with a batch size of 128. For the fine-tuning stage, the pre-trained encoder is kept frozen, and we train a lightweight regression head (512~$\rightarrow$~1) for 200 epochs. We use AdamW with a learning rate of $2 \times 10^{-2}$, weight decay $5 \times 10^{-4}$, and cosine scheduling ($T_{\text{max}}{=}50$). The checkpoint with the lowest validation loss is selected for evaluation. In the JT setting, we optimize the backbone, models and regression head for 200 epochs. Training uses the same SGD setup as pre-training (momentum~0.9, weight decay $5 \times 10^{-4}$, cosine scheduling) with an initial learning rate of $2 \times 10^{-2}$. The total loss combines the NT-Xent contrastive objective with an MSE-based IQA regression loss. This configuration is used consistently for KADID-10k experiments when comparing six PFT and JT experiments.

\noindent\textbf{CrisisMMD:} 
For CrisisMMD, all self-supervised frameworks: SimCLR, BYOL, MoCo, Barlow Twins, DINO, and MAE share a unified training configuration across pre-training (PT), linear evaluation (FT), and joint training (JT). Each method uses SGD with momentum 0.9, a weight decay of $5\times10^{-4}$, cosine annealing learning-rate scheduling, and a fixed random seed of 1. All models are trained for 200 epochs during pre-training and joint training, with a batch size of 128 and a learning rate of $6\times10^{-2}$. During fine-tuning, the encoder is frozen, and only a linear classifier is optimized for 100 epochs using the same batch size (128) but with an increased learning rate of $3\times10^{1}$. MAE follows the same schedule but uses a ViT-Base backbone. This consistent setup ensures a fair comparison across SSL frameworks under both sequential and joint training paradigms.

\noindent\textbf{EarthScape:} 
For EarthScape, all SSL frameworks and the Rotation prediction pretext task are trained under a unified experimental configuration. Colorization was not evaluated because the input consists of a single-channel DEM, making the colorization objective unsuitable for the data modality. Data augmentations were selected to preserve geomorphic relationships. These included random resized crops, random horizontal and vertical flips, and random brightness (elevation offset) and contrast (elevation exaggerations) adjustments. MAE used random flips, random 90$^\circ$ rotations, and a mask ratio of 0.5 to better preserve continuous topographic structure during reconstruction. A ResNet-18 backbone was used across all methods except MAE, which employed a ViT-B/32 encoder. Batch sizes were set to 512 for multilabel classification and 256 for semantic segmentation. All models were optimized using AdamW with a weight decay of $1\times10^{-4}$, an initial learning rate of $3\times10^{-4}$, and a cosine annealing learning-rate schedule. In the PFT setting, models were first pre-trained for 200 epochs and subsequently fine-tuned on downstream tasks with the encoder frozen. In the JT setting, models were trained for a maximum of 200 epochs using early stopping with a patience of 10 epochs. In both paradigms, reported results correspond to the model achieving the lowest validation loss during fine-tuning (PFT) or joint optimization (JT).

\noindent\textbf{ISIC:}
For the ISIC 2016 segmentation task, all SSL frameworks follow the same three-stage training procedure consisting of pre-training (PT), segmentation fine-tuning (FT), and joint training (JT). All models use AdamW for both PT and JT, and the batch size is fixed at 64 across all settings. During PT, each method trains for 200 epochs with a learning rate of $6\times10^{-2}$. During FT, the SSL encoder is frozen, and only the U-Net decoder is optimized for 100 epochs using a learning rate of $1\times10^{-3}$. Joint training is then performed for 200 epochs, combining the SSL loss with the segmentation loss, using a reduced learning rate of $1\times10^{-4}$. This consistent design provides a controlled comparison of SSL frameworks in downstream medical image segmentation.

\noindent\textbf{JSRT:}
For the JSRT lung segmentation and nodule classification experiments, all SSL frameworks follow a unified experimental configuration. During pre-training, all methods use a ResNet-18 backbone and are trained on chest X-ray images resized to 224$\times$224. Pre-training is performed for 100 epochs using SGD optimization with momentum 0.9 and a weight decay of $1\times10^{-4}$. A batch size of 32 is used across all SSL methods. During downstream segmentation fine-tuning, a U-Net architecture with a ResNet-18 encoder initialized from the SSL backbone is used for multi-label lung structure segmentation with five output classes. All images and masks are resized to 224$\times$224. Fine-tuning is conducted for 100 epochs using Adam optimization with a learning rate of $1\times10^{-4}$ and a batch size of 8. In the JT setting, pre-training and supervised fine-tuning are performed simultaneously under the same overall configuration. For downstream classification fine-tuning, ResNet-18 was used with the previously mentioned settings.

\subsection{Hardware Configuration}
All experiments were performed on a machine with an Intel Xeon CPU, 128 GB of system memory, and two NVIDIA RTX A4000 GPUs with 16 GB of VRAM each.

\section{Limitations}
\label{sec:limitations}
Although our study spans multiple SSL frameworks, datasets, and downstream tasks, several limitations remain. First, our experiments are restricted to a finite set of architectures and training configurations, with most evaluations relying on relatively lightweight backbones. More advanced or larger-scale architectures may exhibit different dynamics between PFT and JT. Second, while our analysis is primarily image-based, we include a preliminary extension to multimodal learning using CLIP on CIFAR-10. However, this exploration remains limited to a single small-scale dataset, and broader multimodal settings involving video or other modalities remain unexplored. Finally, while we observe consistent trends across tasks and labeled data percentages, our work is primarily empirical in nature, and a deeper theoretical understanding of why certain SSL objectives benefit more from JT than others remains an important direction for future work. Nevertheless, we believe the breadth of the evaluation provides meaningful practical guidance for informed selection between PFT and JT across real-world vision applications.

\end{document}